\pdfoutput=1 % arXiv: compile with pdfLaTeX (the figures are PDF)
\documentclass[10pt,twocolumn]{article}

\usepackage[a4paper,top=18mm,bottom=18mm,left=16mm,right=16mm,columnsep=6mm,headsep=4mm,footskip=7mm]{geometry}
\usepackage[T1]{fontenc}
\usepackage{newtxtext,newtxmath}
\usepackage[final]{microtype}
\usepackage{amsmath}
\usepackage{bm}
\usepackage{graphicx}
\usepackage{booktabs,multirow,array}
\usepackage[table]{xcolor}
\usepackage{tikz}
\usetikzlibrary{arrows.meta,positioning,calc}
\usepackage[font=small,labelfont=bf,skip=3pt]{caption}
\usepackage[numbers,sort&compress]{natbib}
\usepackage{enumitem}
\usepackage{titlesec}
\usepackage{fancyhdr}
\usepackage[colorlinks=true,linkcolor=blue!60!black,citecolor=blue!60!black,urlcolor=blue!60!black]{hyperref}
\hypersetup{pdftitle={NovoCRF: Globally Normalized De Novo Peptide Sequencing on a Prefix-Mass Lattice},
  pdfauthor={Rui Qiao, Lei Xin, Baozhen Shan}}

\definecolor{crf}{HTML}{1F5FA8}
\definecolor{gold}{HTML}{D9730D}
\definecolor{ev}{HTML}{8A8A8A}
\definecolor{otf}{HTML}{2E7D5B}
\definecolor{sib}{HTML}{7B4FA6}

\titleformat{\section}{\normalfont\large\bfseries}{\thesection}{0.6em}{}
\titleformat{\subsection}{\normalfont\normalsize\bfseries}{\thesubsection}{0.5em}{}
\titlespacing*{\section}{0pt}{1.6ex plus .4ex minus .2ex}{0.8ex plus .2ex}
\titlespacing*{\subsection}{0pt}{1.1ex plus .3ex minus .2ex}{0.5ex plus .1ex}
\setlist{nosep,leftmargin=*}
\newcommand{\R}{\mathbb{R}}
\newcommand{\Da}{\ensuremath{\,\mathrm{Da}}}
\newcommand{\bx}{\bm{x}}
\newcommand{\bv}{\bm{v}}
\newcommand{\be}{\bm{e}}
\newcommand{\bg}{\bm{g}}
\newcommand{\bu}{\bm{u}}
\newcommand{\mE}{\mathbf{E}}
\newcommand{\mH}{m_{\mathrm{H}^+}}
\newcommand{\st}{\mathrm{st}}
\newcommand{\Cfive}{CasanovoV5}
\newcommand{\mz}{\textit{m/z}}
\newcommand{\cpuone}{342}   % spectra/s on the CPU alone, one pass, 12 test sets
\newcommand{\cputwo}{130}   % spectra/s on the CPU alone, two passes, 12 test sets
\newcommand{\cpux}{102}     % spectra/s on the CPU alone, two passes + CRSE, 8 sets: 1,971,695 spectra / 19,377.8 s

\makeatletter
\renewcommand{\maketitle}{\twocolumn[{%
  \begin{center}
  {\Large\bfseries NovoCRF: Globally Normalized \textit{De Novo} Peptide Sequencing\\[1pt] on a Prefix-Mass Lattice\par}
  \vspace{2.5mm}
  {\setlength{\tabcolsep}{14pt}\renewcommand{\arraystretch}{1.1}%
  \begin{tabular}{ccc}
  \textbf{Rui Qiao} & \textbf{Lei Xin} & \textbf{Baozhen Shan}\\
  {\small\href{mailto:rqiao@bioinfor.com}{\texttt{rqiao@bioinfor.com}}} &
  {\small\href{mailto:lxin@bioinfor.com}{\texttt{lxin@bioinfor.com}}} &
  {\small\href{mailto:bshan@bioinfor.com}{\texttt{bshan@bioinfor.com}}}
  \end{tabular}\par}
  \vspace{2pt}
  {\normalsize Bioinformatics Solutions Inc.\par}
  \end{center}
  \vspace{2mm}}]}
\makeatother

\begin{document}
\maketitle
\thispagestyle{fancy}

\begin{abstract}
\noindent In the last decade, \textit{de novo} peptide sequencing has moved from dynamic programming over hand-designed scores to
large transformers that learn the task end to end and are applied to every run unchanged. We present NovoCRF, to our
knowledge the first integrated \textit{de novo} sequencing workflow that adapts to the characteristics of each MS/MS
run, including but not limited to its mass shift, its fragmentation pattern and its peptide elution order. NovoCRF's core
is a conditional random field (CRF) whose states
are prefix masses, so that its paths are exactly the peptides within $\pm0.06\Da$ of the precursor mass. A small network scores each
edge from physics-derived ion-match features and is trained with its scores globally normalized over all
mass-feasible peptides. On each run, NovoCRF recalibrates the precursor and fragment masses, fine-tunes part of its
weights on the run's own confident answers, uses retention time as further evidence, and lets the sibling spectra of
one peptide, within the run and/or across replicate runs, support each other. None of this needs labels or a database. On
twelve held-out data-dependent acquisition datasets, NovoCRF has the highest peptide recall on eleven, by 5.5 to 46.5
points over the best of four transformer models, and on the twelfth, a
human tryptic digest, once it reads two replicate runs as well. On the peptides that no tool has seen in training, it
has the highest recall on all twelve. NovoCRF also searches modifications it was never trained on at their exact mass: on a public
spike-in of peptides with non-natural residues it finds 74.7\% of the labeled spectra, against 64.4\% for RNovA. The
model has 1.80M parameters and trains in about 21 GPU-hours. With
a GPU it sequences 313 spectra/s in its default two passes, and it also runs on the CPU alone, at \cputwo\ spectra/s.
\end{abstract}

% ================================================================================================
\section{Introduction}

\textit{De novo} sequencing reads a peptide from its tandem mass (MS/MS) spectrum without a protein database. It is the method
of choice when the sequence is not in a reference database: neoantigens, antibodies, RNA-edited proteins and organisms without a
curated proteome. The spectrum constrains the peptide through \emph{prefix masses}: a cleavage after residue $j$ yields
a b ion at prefix mass $p_j$ and a complementary y ion at $M-p_j$, where $M$ is the precursor mass.

Classical methods made this structure explicit. PEAKS~\cite{ma2003} and NovoHMM~\cite{fischer2005} use dynamic
programming (DP) along the prefix-mass axis to find the optimal path under a defined scoring function: hand-tuned ion
statistics in PEAKS, a generative hidden Markov model (HMM) trained by expectation--maximization in NovoHMM. Decoding is exact
and the precursor mass is satisfied by construction, but the scores are not learned discriminatively.

DeepNovo~\cite{tran2017} introduced deep learning: a convolutional neural network reads intensity windows at the theoretical ions of each
candidate next residue, a long short-term memory network (LSTM) adds sequence context, and beam search decodes a per-step softmax.
PointNovo~\cite{qiao2021} replaced the intensity grid with an order-invariant network over the peaks themselves, fed
with the \mz{} difference between each peak and each theoretical ion, which made it independent of instrument
resolution. It also started the LSTM from a sinusoidal embedding of the spectrum. Casanovo~\cite{yilmaz2022,yilmaz2024}
made the task fully end to end: each peak is encoded by sinusoids into a 512-d vector, and a 9-layer
transformer encoder and 9-layer decoder translate the peak set into residues. Later transformer models include
InstaNovo~\cite{eloff2025}, an autoregressive model whose companion InstaNovo+ refines a whole predicted sequence by
multinomial diffusion, and the non-autoregressive $\pi$-PrimeNovo~\cite{zhang2025}. GraphNovo~\cite{mao2023} combines a
spectrum graph with graph transformers. Like Casanovo, all of them lift each peak's \mz{} into a high-dimensional
vector. InstaNovo and $\pi$-PrimeNovo use sinusoidal embeddings. GraphNovo feeds each peak's scaled \mz{}, together with seven intensity
and rank features of the peak, through a learned network into its graph node's 2{,}048-d vector, so the
\mz{} information is used only implicitly, in a high-dimensional latent space. Building on GraphNovo,
RNovA~\cite{mao2026} first finds a path of masses through the spectrum graph, keeping gaps that no residue explains
as mass tags, and then fills the path with residues, giving each candidate step an action value.

Peak encoding is popular and proven to work, yet it is a counterintuitive use of capacity: one scalar \mz{} is encoded by
512 floating-point numbers, and not even losslessly. The network must then relearn arithmetic that is known exactly,
such as $\mathrm{b}^{+}+\mathrm{y}^{+}=M+2\mH$, with $\mH$ the proton mass. These models also score a peptide
locally: with a softmax per decoding step or, with connectionist temporal classification (CTC), per output position,
or, in RNovA, with an action value per candidate at each step. None of them normalizes over the set of peptides that
fit the precursor mass: the mass constraint enters through decoding rules, filters or the space the model decodes in.

These models share one more property: each is trained once and then applied to every run unchanged. Yet mass
spectrometry runs differ from one another. The mass calibration drifts, the collision energy and the
type of mass analyzer change how a peptide fragments, and the chromatography sets when a peptide would elute. Nor are the spectra
of a run independent of each other: a peptide is often fragmented several times in a run, and again in the other runs
of the same sample. Database search pipelines have used both facts for a long time (\S\ref{sec:related}).

We present NovoCRF, to the best of our knowledge the first integrated \textit{de novo} sequencing workflow that
adapts to the characteristics of each MS/MS run, including but not limited to its mass shift, its fragmentation
pattern and its peptide elution order. After a first pass over a run, NovoCRF recalibrates the run's precursor and
fragment masses, calibrates the collision energy that its fragment-intensity predictor reads, and fine-tunes part of
its sequencing model and intensity predictor on the run's own confident answers. It fits a per-run retention time (RT)
regression on the output of an indexed RT (iRT) predictor that is itself fine-tuned on the run, so the
liquid-chromatography dimension, which per-spectrum sequencing leaves unused, also informs the answer. It links the
sibling spectra of each peptide, within the run and optionally across the runs of one sample, so that a spectrum with
missing fragments is read together with spectra that have them. The run is then decoded again. None of these steps
needs labels or a database.

Such a workflow needs a sequencing model that is small enough to be fine-tuned on every run and exact enough to
search the same space in training, fine-tuning and decoding.
NovoCRF therefore returns to the classical formulation and makes it discriminative. Conceptually it is close to NovoHMM, a chain
along the prefix-mass axis decoded by DP, but it is a conditional random field (CRF)~\cite{lafferty2001}: in training, the
score of the true path is normalized over all mass-feasible paths rather than step by step. To the best of our knowledge, NovoCRF is the
first \textit{de novo} sequencing method trained with exact normalization over \emph{all} mass-feasible peptides. The closest
transformer model, $\pi$-PrimeNovo~\cite{zhang2025}, is also trained on the probability of a whole sequence, but its CTC loss
normalizes each output position separately, over sequences of any mass, and its precise mass control enforces the
precursor only at decoding, with a search that need not find the most probable sequence of the right mass
(\S\S\ref{sec:train}--\ref{sec:inference}). In spirit NovoCRF
is a successor to PointNovo: a \textit{de novo} tool using carefully engineered input features that encode domain knowledge
(theoretical fragment \mz{}, ppm-scale soft matching, isotope envelopes, neutral losses, immonium ions and internal
fragments), instead of a black box with tens of millions of weights. With 1.80M parameters in all, NovoCRF is small
enough to run on a computer without a GPU, using the CPU only.

In this paper we also discuss how to better benchmark \textit{de novo} tools. Currently, \textit{de novo} tools are usually scored by
peptide recall against database-search identifications. A database search may accept a peptide with long stretches
that no b or y ion supports, because a search engine looks for the sequence that best matches the spectrum within a
search space confined by the database, which already incorporates prior information from the genome. A \textit{de novo}
tool, on the other hand, must choose the residues in such a stretch with little direct evidence, from an open search
space. We therefore report recall by how much fragment evidence each label has, and by whether its peptide was in a
tool's training data. Both splits show how much of a tool's recall depends on having seen the peptide in
training (\S\S\ref{sec:evidence}--\ref{sec:seen}).
Our contributions are:
\begin{itemize}
\item The first integrated \textit{de novo} workflow that adapts to each run without labels or a database:
  calibration of the run's masses and collision energy, fine-tuning of the CRF and of the intensity and iRT
  predictors on the run's own confident answers, a retention-time term, and sibling spectra enhancement within a run
  and across runs (\S\S\ref{sec:otf}--\ref{sec:xrun}).
\item A CRF over a 0.002\Da\ prefix-mass lattice whose partition function is computed exactly in $\approx M/(57\Da)$
  sequential steps, with 18.5k parameters in the potentials, followed by a cascade of three bidirectional gated recurrent unit (BiGRU) rerankers
  (\S\ref{sec:method}).
\item A modular design in which precursor correction from MS1, exact search and sequence context are separate
  modules, and in which modifications never seen in training are searched at their exact mass (\S\ref{sec:userptm}).
\item An evaluation on twelve held-out datasets that splits recall by fragment evidence and by seen and unseen
  peptides. NovoCRF has the highest recall on eleven of them, and on all twelve on the peptides that no tool has
  seen in training (\S\ref{sec:results}).
\item A lightweight implementation: 1.80M parameters, about 21 GPU-hours of training, and 2.3 to 33 times the
  throughput of four transformer models with a GPU, while it also runs on the CPU alone
  (\S\ref{sec:speed}).
\end{itemize}

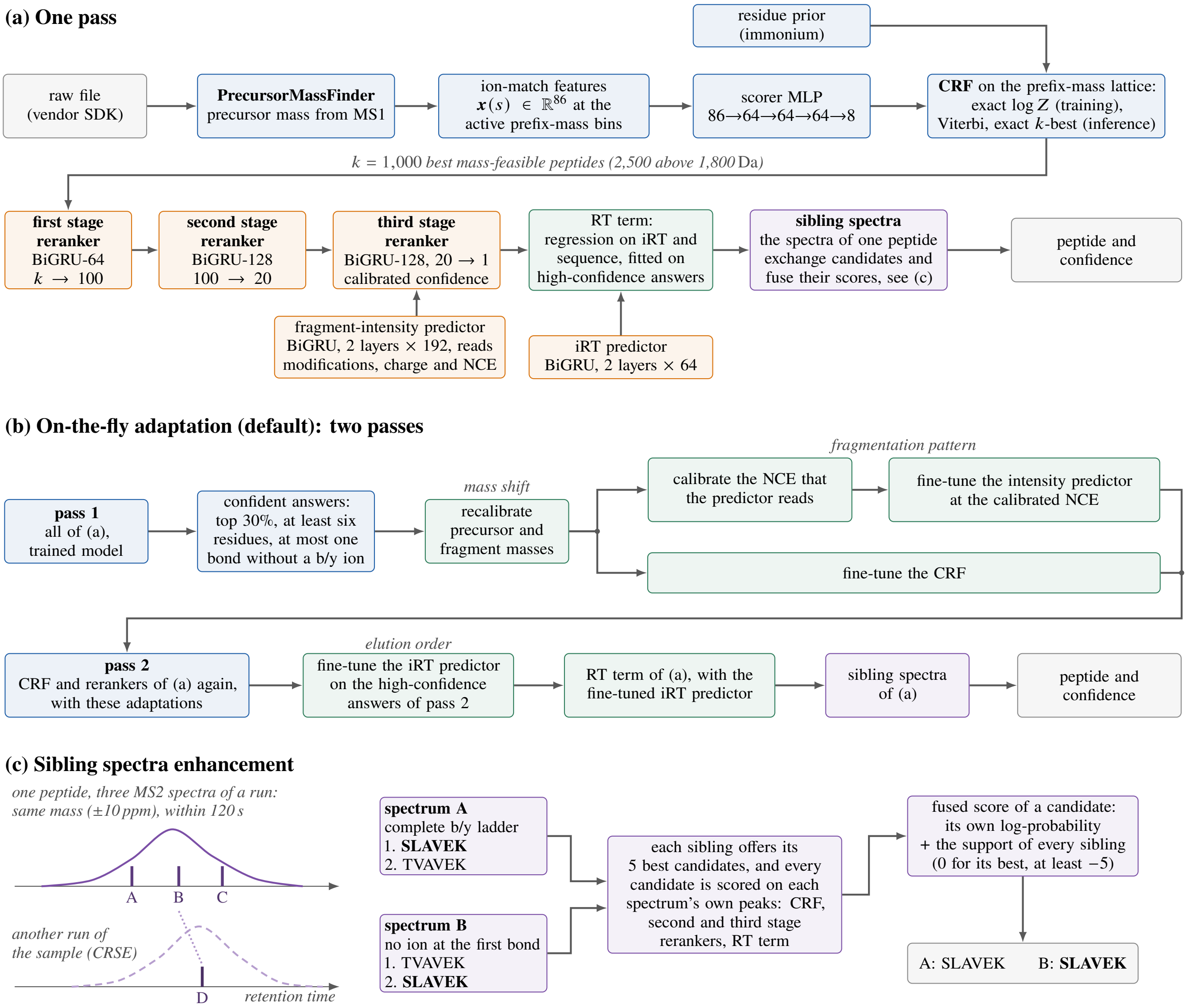
\begin{figure*}[t]
\centering
\begin{tikzpicture}[x=1cm,y=1cm,font=\scriptsize,
    box/.style={draw=#1,fill=#1!8,rounded corners=2pt,align=center,inner sep=2pt,minimum height=0.95cm},
    box/.default=crf,
    arr/.style={-{Latex[length=1.6mm]},thick,draw=black!65},
    lab/.style={font=\scriptsize\itshape,text=black!70,inner sep=1.5pt},
    panel/.style={font=\small\bfseries,anchor=west,inner sep=0pt}]
  % ------------------------------------------------------------------ (a) one pass
  \node[panel] at (0,1.3) {(a) One pass};
  \node[box=ev,text width=2.0cm] (raw) at (1.05,0) {raw file\\(vendor SDK)};
  \node[box,text width=2.8cm] (pmf) at (4.35,0) {\textbf{PrecursorMassFinder}\\precursor mass from MS1};
  \node[box,text width=3.0cm] (feat) at (8.05,0) {ion-match features\\$\bx(s)\in\R^{86}$ at the\\active prefix-mass bins};
  \node[box,text width=2.5cm] (mlp) at (11.6,0) {scorer MLP\\$86{\to}64{\to}64{\to}64{\to}8$};
  \node[box,text width=2.5cm,minimum height=0.55cm] (prior) at (11.6,1.2) {residue prior (immonium)};
  \node[box,text width=3.4cm] (crf) at (15.55,0) {\textbf{CRF} on the prefix-mass lattice:\\exact $\log Z$ (training),\\Viterbi, exact $k$-best (inference)};
  \node[box=gold,text width=1.75cm] (s1) at (0.95,-2.15) {\textbf{first stage}\\\textbf{reranker}\\BiGRU-64\\$k\to100$};
  \node[box=gold,text width=2.05cm] (s2) at (3.4,-2.15) {\textbf{second stage}\\\textbf{reranker}\\BiGRU-128\\$100\to20$};
  \node[box=gold,text width=2.35cm] (s3) at (6.15,-2.15) {\textbf{third stage}\\\textbf{reranker}\\BiGRU-128, $20\to1$\\calibrated confidence};
  \node[box=otf,text width=2.6cm] (rt) at (9.2,-2.15) {RT term:\\regression on iRT and\\sequence, fitted on\\high-confidence answers};
  \node[box=sib,text width=2.8cm] (sib) at (12.6,-2.15) {\textbf{sibling spectra}\\the spectra of one peptide\\exchange candidates and\\fuse their scores, see (c)};
  \node[box=ev,text width=2.4cm] (out) at (16.3,-2.15) {peptide and\\confidence};
  \node[box=gold,text width=3.3cm,minimum height=0.6cm] (ip) at (5.75,-3.6) {fragment-intensity predictor\\BiGRU, 2 layers $\times$ 192, reads\\modifications, charge and NCE};
  \node[box=gold,text width=2.6cm,minimum height=0.6cm] (irt) at (9.2,-3.75) {iRT predictor\\BiGRU, 2 layers $\times$ 64};
  \draw[arr] (raw) -- (pmf);
  \draw[arr] (pmf) -- (feat);
  \draw[arr] (feat) -- (mlp);
  \draw[arr] (mlp) -- (crf);
  \draw[arr] (prior.east) -| (crf.north);
  \draw[arr] (crf.south) -- ++(0,-0.55) -| node[lab,pos=0.25,above] {$k=1{,}000$ best mass-feasible peptides (2{,}500 above 1{,}800\Da)} (s1.north);
  \draw[arr] (s1) -- (s2);
  \draw[arr] (s2) -- (s3);
  \draw[arr] (ip.north -| s3.south) -- (s3.south);
  \draw[arr] (s3) -- (rt);
  \draw[arr] (irt) -- (rt);
  \draw[arr] (rt) -- (sib);
  \draw[arr] (sib) -- (out);
  % ------------------------------------------------------------------ (b) two passes
  % row 1: the first pass and what is learnt from it between the passes; row 2: the second pass. The iRT predictor
  % is fine-tuned inside the second pass, after its decoding and before its RT term (the order of the engine).
  \begin{scope}[yshift=-6.35cm]
  \node[panel] at (0,1.55) {(b) On-the-fly adaptation (default): two passes};
  \node[box,text width=2.0cm] (p1) at (1.07,0) {\textbf{pass 1}\\all of (a),\\trained model};
  \node[box,text width=2.5cm] (sel) at (4.2,0) {confident answers:\\top 30\%, at least six\\residues, at most one\\bond without a b/y ion};
  \node[box=otf,text width=2.0cm] (cal) at (7.35,0) {recalibrate\\precursor and\\fragment masses};
  \node[box=otf,text width=2.9cm,anchor=west] (nce) at (9.6,0.62) {calibrate the NCE that\\the predictor reads};
  \node[box=otf,text width=3.9cm,anchor=east] (ftp) at (17.25,0.62) {fine-tune the intensity predictor\\at the calibrated NCE};
  % the CRF box spans exactly from the left edge of the NCE box to the right edge of the predictor box
  \node[box=otf,text width=7.51cm,minimum height=0.6cm] (ftc) at ($(nce.west)!0.5!(ftp.east)+(0,-1.24)$) {fine-tune the CRF};
  \node[lab,anchor=south] at (cal.north) {mass shift};
  \node[lab,anchor=south] at ($(nce.north west)!0.5!(ftp.north east)$) {fragmentation pattern};
  \draw[arr] (p1) -- (sel);
  \draw[arr] (sel) -- (cal);
  \draw[arr] (cal.east) -- ++(0.42,0) |- (nce.west);
  \draw[arr] (cal.east) -- ++(0.42,0) |- (ftc.west);
  \draw[arr] (nce) -- (ftp);
  % row 2
  \node[box,text width=3.5cm,anchor=west] (p2) at (0,-2.3) {\textbf{pass 2}\\CRF and rerankers of (a) again,\\with these adaptations};
  \node[box=otf,text width=3.0cm,anchor=west] (ftr) at (4.45,-2.3) {fine-tune the iRT predictor\\on the high-confidence\\answers of pass 2};
  \node[box=otf,text width=3.0cm,anchor=west] (rt2) at (8.35,-2.3) {RT term of (a), with the\\fine-tuned iRT predictor};
  \node[box=sib,text width=2.0cm,anchor=west] (sb2) at (12.25,-2.3) {sibling spectra\\of (a)};
  \node[box=ev,text width=2.3cm,anchor=east] (fin) at ($(ftp.east)+(0.32,-2.92)$) {peptide and\\confidence};
  \node[lab,anchor=south] at (ftr.north) {elution order};
  % both branches end in one line at the right, which wraps to the second row as the arrow in (a) does
  \coordinate (bus) at ($(ftp.east)+(0.32,0)$);
  \draw[arr] (ftp.east) -- (bus) -- ++(0,-1.92) -| (p2.north);
  \draw[thick,draw=black!65] (ftc.east) -- (ftc.east -| bus);
  \fill[black!65] ($(cal.east)+(0.42,0)$) circle (1.1pt) (ftc.east -| bus) circle (1.1pt);
  \draw[arr] (p2) -- (ftr);
  \draw[arr] (ftr) -- (rt2);
  \draw[arr] (rt2) -- (sb2);
  \draw[arr] (sb2) -- (fin);
  \end{scope}
  % ------------------------------------------------------------------ (c) sibling spectra
  \begin{scope}[yshift=-11.4cm]
  \node[panel] at (0,1.55) {(c) Sibling spectra enhancement};
  % elution profile of one peptide in the run, with its MS2 spectra
  \draw[black!65,-{Latex[length=1.6mm]}] (0.15,-0.25) -- (5.0,-0.25);
  \draw[sib,thick] plot[smooth,tension=0.55] coordinates {(0.55,-0.25) (1.3,-0.16) (1.9,0.16) (2.5,0.6) (3.1,0.2) (3.75,-0.14) (4.45,-0.25)};
  \foreach \x/\n in {1.9/A,2.6/B,3.25/C}{
    \draw[sib!75!black,very thick] (\x,-0.25) -- (\x,0.05);
    \node[font=\scriptsize,text=sib!65!black,anchor=north,inner sep=1.5pt] at (\x,-0.27) {\n};}
  \node[lab,anchor=west,align=left] at (0.05,1.0) {one peptide, three MS2 spectra of a run:\\same mass ($\pm$10\,ppm), within 120\,s};
  % a companion run (CRSE)
  \draw[black!65,-{Latex[length=1.6mm]}] (0.15,-1.75) -- (5.0,-1.75) node[lab,anchor=north east,inner sep=1.5pt] {retention time};
  \draw[sib!55,thick,densely dashed] plot[smooth,tension=0.55] coordinates {(1.0,-1.75) (1.7,-1.66) (2.3,-1.32) (2.9,-0.85) (3.5,-1.3) (4.1,-1.68) (4.75,-1.75)};
  \draw[sib!60!black,very thick] (2.95,-1.75) -- (2.95,-1.45);
  \node[font=\scriptsize,text=sib!65!black,anchor=north,inner sep=1.5pt] at (2.95,-1.77) {D};
  \draw[sib!60,densely dotted,thick] (2.6,-0.62) -- (2.95,-1.42);
  \node[lab,anchor=west,align=left] at (0.05,-1.12) {another run of\\the sample (CRSE)};
  % the exchange
  \node[box=sib,text width=2.35cm,align=left] (sa) at (6.85,0.5) {\textbf{spectrum A}\\complete b/y ladder\\1.\ \textbf{SLAVEK}\\2.\ TVAVEK};
  \node[box=sib,text width=2.35cm,align=left] (sb) at (6.85,-1.25) {\textbf{spectrum B}\\no ion at the first bond\\1.\ TVAVEK\\2.\ \textbf{SLAVEK}};
  \node[box=sib,text width=3.35cm] (sc) at (10.75,-0.375) {each sibling offers its\\5 best candidates, and every\\candidate is scored on each\\spectrum's own peaks: CRF,\\second and third stage\\rerankers, RT term};
  \node[box=sib,text width=3.3cm] (fu) at (15.2,0.5) {fused score of a candidate:\\its own log-probability\\$+$ the support of every sibling\\(0 for its best, at least $-5$)};
  \node[box=ev,text width=3.3cm,minimum height=0.6cm] (ans) at (15.2,-1.4) {A: SLAVEK\qquad B: \textbf{SLAVEK}};
  \draw[arr] (sa.east) -- ++(0.35,0) |- ($(sc.west)+(0,0.2)$);
  \draw[arr] (sb.east) -- ++(0.35,0) |- ($(sc.west)+(0,-0.2)$);
  \draw[arr] (sc.east) -- ++(0.4,0) |- (fu.west);
  \draw[arr] (fu) -- (ans);
  \end{scope}
\end{tikzpicture}
\caption{NovoCRF. (a) One pass. PrecursorMassFinder supplies the precursor mass, the conditional random field (CRF)
finds the best-scoring mass-feasible peptides on the prefix-mass lattice, and three bidirectional gated recurrent unit
(BiGRU) rerankers rescore them, the third stage reranker
also comparing each candidate's observed fragment intensities with predicted ones. A retention-time (RT) regression,
fitted on the run's high-confidence answers, predicts each candidate's RT from its sequence features and the indexed
RT (iRT) that a predictor gives it, and the deviation from the observed RT adds a penalty term, the RT term,
to the scores of the third stage reranker's candidates. Sibling spectra then exchange their candidates. (b) The default two-pass mode adapts
to the run's mass shift, fragmentation pattern and elution order. Each pass runs all of (a) and so fits the RT
regression on its own high-confidence answers. The confident answers of the first pass recalibrate the run's precursor and
fragment masses. Two independent branches then use the recalibrated spectra: the normalized collision energy (NCE)
that the intensity predictor reads is calibrated before the predictor is fine-tuned at it, and the CRF is fine-tuned.
The second pass decodes the run again with these adaptations. It then fine-tunes the iRT predictor on its own
high-confidence answers, and its RT term uses the fine-tuned predictor. Each fine-tuned model is kept only if it
improves on held-out answers. (c) Sibling spectra enhancement. Spectra of one peptide offer each
other their best candidates, every candidate is scored on each spectrum's own peaks, and each spectrum takes the
candidate with the best fused score of Eq.~\eqref{eq:fuse}. In the example, spectrum B cannot tell SL from TV, which
weigh the same (Fig.~\ref{fig:lattice}), and follows spectrum A. With cross-run spectra enhancement (CRSE), the
siblings also come from other runs of the same sample.}
\label{fig:pipeline}
\end{figure*}

% ================================================================================================
\section{Related work}\label{sec:related}

\noindent\textbf{Spectrum graphs and dynamic programming.} Before deep learning, \textit{de novo} tools scored paths through a graph
of candidate prefix masses: PEAKS~\cite{ma2003} with hand-tuned ion statistics, PepNovo~\cite{frank2005} with a
probabilistic network of fragmentation, NovoHMM~\cite{fischer2005} with a generative HMM and Novor~\cite{ma2015}
with decision trees learned from a spectral library. All decode exactly, and none trains a discriminative model against
the set of all paths.

\noindent\textbf{Mass constraints in neural \textit{de novo} sequencing.} Neural sequencers generate residues step by step or position
by position and bring the precursor mass in at decoding time. DeepNovo filters beam-search extensions with a knapsack
over residue masses~\cite{tran2017}, and InstaNovo's knapsack beam search keeps only extensions that can still reach
the precursor mass~\cite{eloff2025}. Casanovo ranks predictions that miss the precursor below those that
fit~\cite{yilmaz2024}. $\pi$-PrimeNovo, a non-autoregressive transformer trained with CTC, adds a precise mass-control
unit that decodes again under the mass constraint when its first answer misses the precursor~\cite{zhang2025}
(\S\ref{sec:inference}). GraphNovo first searches for an optimal path in a spectrum graph, which then guides a
second-stage sequence predictor through the segments without fragmentation~\cite{mao2023}. RNovA keeps this two-stage
design and fills the segments under a knapsack-like mass constraint~\cite{mao2026}. None of them normalizes over the
peptides that fit the precursor mass: they score each step or output position separately. NovoCRF builds the mass constraint into the state space instead, so that the model's
normalizer ranges only over mass-feasible peptides.

\noindent\textbf{Global normalization.} Conditional random fields~\cite{lafferty2001} were introduced to avoid the label bias of
locally normalized models such as maximum-entropy Markov models (MEMMs), whose per-state normalization lets early decisions
ignore later evidence. Andor et al.~\cite{andor2016} showed that globally normalized neural models are strictly more
expressive than their locally normalized counterparts for transition-based structured prediction. Exact global
normalization is usually expensive because the partition function ranges over an exponential set. The prefix-mass
lattice makes it cheap, because its states are ordered by mass and every edge spans at least one glycine
(\S\ref{sec:train}).

\noindent\textbf{Adapting to each run.} Database-search pipelines routinely adapt to the run they analyze by searching
it more than once: MaxQuant recalibrates each run's masses from the peptides of a first search before its main
search~\cite{cox2008}, and DIA-NN refines its retention-time alignment and mass correction on the run's own peptides
over successive passes~\cite{demichev2020}. NovoCRF adapts without a database: it fine-tunes the sequencing model
itself on its own confident first-pass answers, and decodes the run again with recalibrated masses and a retention-time
term.

\noindent\textbf{Several spectra of one peptide.} A run usually holds more than one spectrum of a peptide, at other
charge states or from repeated selection, and replicate runs hold more. Database-search pipelines have long used this
redundancy: spectral clustering merges such spectra into a consensus spectrum before the search~\cite{frank2008},
spectral networks align the spectra of overlapping and modified peptides~\cite{bandeira2007}, and match between runs
transfers an identification to another run by its mass and its aligned retention time~\cite{tyanova2016}. For
\textit{de novo} sequencing, spectrum fusion combined the two spectra that two fragmentation methods give of one
precursor~\cite{datta2009}. The neural network based \textit{de novo} models above sequence every spectrum on its own. NovoCRF also searches every
spectrum on its own, and then lets the spectra of one peptide, within a run and across the runs of one sample,
exchange their candidates and support each other (\S\S\ref{sec:sibling}--\ref{sec:xrun}).

\noindent\textbf{Evaluating \textit{de novo} sequencing.} \textit{De novo} tools are compared on spectra labeled by database search, usually by
peptide-level recall or precision~\cite{yilmaz2024,zhang2025,mao2023,eloff2025,mao2026}.
NovoBench~\cite{zhou2024} reports accuracy against each spectrum's missing-fragmentation ratio. We split peptide recall
by the fragment evidence of each label, to see whether a tool predicts a peptide from the
spectrum's support or by remembering the sequence (\S\ref{sec:evidence}). We also split it by whether the
labeled sequence is in the tool's training set, to compare a tool's recall on seen and on unseen
peptides (\S\ref{sec:seen}). We argue that for \textit{de novo} tools, recall on unseen peptides matters far more
than recall on seen peptides.

% ================================================================================================
\section{Method}\label{sec:method}

\noindent\textbf{The Zen of NovoCRF.} NovoCRF consists of five modules, each with a single job (Fig.~\ref{fig:pipeline}):
\begin{enumerate}
\item PrecursorMassFinder reads the MS1 and MS2 scans through the vendor's software development kit (SDK) and finds the correct precursor mass of
  each MS2 scan from MS1 information (\S\ref{sec:rdr}).
\item The CRF trusts the precursor mass it is given. A CRF with a multilayer perceptron (MLP) scorer and the Viterbi algorithm efficiently find
  the sequences that match that mass and have good spectrum support (\S\S\ref{sec:lattice}--\ref{sec:inference}).
\item The rerankers read each candidate as a whole sequence, with the spectrum's evidence at every residue and
  cleavage, including internal fragments and predicted fragment intensities, and so order the candidates where the
  lattice cannot, such as inside a stretch without fragment ions (\S\S\ref{sec:rerank}--\ref{sec:ipred}).
\item On-the-fly adaptation fits the characteristics of each run from the run's own confident answers. It calibrates
  the precursor and fragment masses and the collision energy, fine-tunes the CRF and the intensity predictor to the
  run's fragmentation, and learns the run's elution order, through a per-run regression on the output of an iRT
  predictor that is itself fine-tuned on the run (\S\ref{sec:irt}, \S\ref{sec:otf}).
\item Sibling spectra enhancement lets the spectra of one peptide, within a run and optionally across the runs of one
  sample, exchange their candidates and support each other (\S\S\ref{sec:sibling}--\ref{sec:xrun}).
\end{enumerate}
Each module in NovoCRF has one clear responsibility, so it is easier to fine-tune and optimize.

\begin{figure}[t]
\centering
\resizebox{\columnwidth}{!}{%
\begin{tikzpicture}[x=1cm,y=1cm,font=\scriptsize]
  % x = prefix mass x 0.012290 cm/Da (627.359 Da of SLAVEK at 7.71 cm), so every edge is as long as its residue mass
  % evidence bars (active bins)
  \foreach \x/\h in {0.55/0.10,1.070/0.36,2.00/0.12,2.459/0.30,3.332/0.27,3.677/0.09,4.550/0.34,5.20/0.13,
                     6.136/0.29,6.837/0.08,7.20/0.11}
    \fill[ev!70] (\x-0.04,0) rectangle (\x+0.04,-\h);
  \draw[-{Latex[length=2mm]}] (-0.1,0) -- (8.25,0);
  \node[anchor=north east,inner sep=1pt] at (8.25,-0.4) {prefix mass (bins of $\Delta\!\approx\!0.002$\,Da)};
  \node[anchor=east] at (-0.1,0) {0};
  % terminal window
  \fill[gold!18] (7.55,-0.05) rectangle (7.87,1.12);
  \node[gold!80!black,anchor=south,inner sep=1pt] at (7.71,1.12) {$t^\ast\!\pm\!W$};
  % other mass-feasible edges: TV for SL, VA for AV, GA for K (ends 18 bins below t*)
  \foreach \a/\b/\r in {0/1.242/T,1.242/2.459/V,2.459/3.677/V,3.677/4.550/A,6.136/6.837/G,6.837/7.710/A}
    \draw[ev,thin,dashed] (\a,0) .. controls +(82:1.2) and +(98:1.2) .. node[above=-1pt,text=ev!60!black]{\r} (\b,0);
  \foreach \x in {1.242,3.677,6.837}
    \fill[ev] (\x,0) circle (1.2pt);
  % gold path S L A V E K
  \foreach \a/\b/\r in {0/1.070/S,1.070/2.459/L,2.459/3.332/A,3.332/4.550/V,4.550/6.136/E,6.136/7.710/K}
    \draw[gold,very thick] (\a,0) to[out=65,in=115,looseness=1.25] node[above=0.3pt,inner sep=0.6pt,fill=white,text=gold!80!black]{\textbf{\r}} (\b,0);
  \foreach \x in {0,1.070,2.459,3.332,4.550,6.136,7.710}
    \fill[gold!80!black] (\x,0) circle (1.5pt);
\end{tikzpicture}}
\caption{The prefix-mass lattice. The true peptide (SLAVEK, orange) is one path from bin 0 to the terminal window,
normalized against all mass-feasible paths (gray). Bars show the evidence at active bins.}
\label{fig:lattice}
\end{figure}
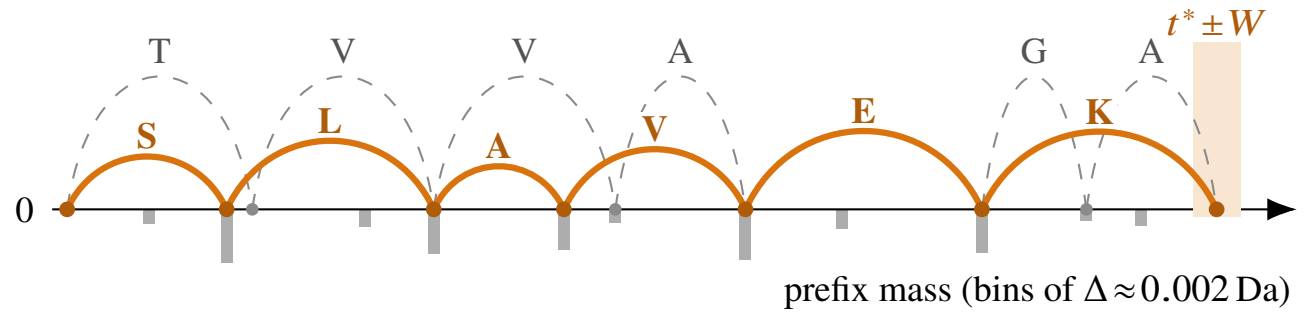

\noindent\textbf{Notation.} Italic letters are scalars, bold lowercase letters vectors (e.g.\ $\bv\in\R^8$), bold uppercase
letters matrices. The inner product is $\langle\cdot,\cdot\rangle$, and $\mH$ is the proton mass (masses in Da).

\subsection{PrecursorMassFinder: the precursor from MS1}\label{sec:rdr}
PrecursorMassFinder reads the MS1 and MS2 spectra with the vendor's file reader and tries to find the right precursor
mass for each MS2 scan from the signals in the corresponding MS1 scan. For each MS2 scan, the MS1 peak nearest
the isolation target (within 15\,ppm on Orbitrap data, 20\,ppm on timsTOF) is the anchor. The anchor may be a heavier isotope peak, so the candidate monoisotopic peaks are the anchor and the
positions 1 to $J$ isotope spacings ($1.00335/z$ for the vendor's charge $z$) lower in \mz\ ($J=2$--6, growing with
mass, and on timsTOF data at least three spacings below the vendor's monoisotopic \mz), each scored by how well the intensities $\bm{r}$ at its first four isotope positions match the averagine
abundances $\bm{a}$, penalizing a peak one spacing lighter still (intensity $r_{-1}$): $\langle\bm{a},\bm{r}\rangle^2/(\|\bm{r}\|^2+r_{-1}^2/4)$. The best candidate's centroid is reported.
When the runner-up scores at least 0.97 times as high, NovoCRF searches the spectrum at both masses and ranks the
candidates of both together, with a fixed penalty on the runner-up's.
PrecursorMassFinder is built into the NovoCRF engine, which reads Thermo and Bruker raw files directly, and is also a
separate program that writes Mascot generic format (MGF) files for other tools.

\subsection{Prefix-mass lattice}\label{sec:lattice}
A spectrum $X$ has peaks $\{(m_i,I_i)\}$, precursor neutral mass $M$ and charge $z$. States are bins $s=0,\dots,S$ of
width $\Delta=0.001995495\Da$, chosen near 0.002\Da\ to minimize the worst residue-mass rounding error (0.00066\Da). Bin
$s$ stands for the prefix mass $s\Delta$. Each of the $K=24$ edges adds one residue of mass $m_k$ and spans
$o_k=\mathrm{round}(m_k/\Delta)$ bins. The edges cover 19 residue masses (I/L merged, fixed carbamidomethyl C) and oxidized M,
as well as unmodified C and phosphorylated S, T and Y, which are searched only when requested. Deamidated N/Q have exactly the
masses of D/E and share their edges. Three more edges, for N-terminal modifications, may leave only bin 0, the
peptide's N-terminus. Acetylation and carbamylation edges add only the modification mass ($+42.011$ and
$+43.006\Da$), and the first residue follows as its own edge. Pyro-Glu makes an N-terminal Q 17.027\Da\ lighter. An
edge cannot have negative mass, so it is one edge for the modified Q (111.032\Da). A path starts at bin 0
and ends in the terminal window, within $W=30$ bins ($\pm0.06\Da$) of $t^\ast=\mathrm{round}((M-m_{\mathrm{H_2O}})/\Delta)$, so every path is a
mass-feasible peptide, and equal-mass compositions (GG vs.\ N) are distinct paths through shared states
(Fig.~\ref{fig:lattice}).

\subsection{Potentials}\label{sec:potentials}
Every edge is scored from the fragment evidence at the bins it touches. Evidence is computed only at a small set of \emph{active} bins, and all other bins share one learned vector.

\noindent\textbf{Active bins.} Reading a peak of \mz{} $m$ as a b or y ion of charge 1 or 2 implies one of four prefix masses,
$\hat{x}\in\{m-\mH,\ M+\mH-m,\ 2m-2\mH,\ M+2\mH-2m\}$. Over the most intense peaks (400 in training, 800 at inference), with intensities scaled to
the base peak, the active set $\mathcal{A}(X)$ contains every bin within $\pm20$\,ppm of such an $\hat{x}$. The window
is relative to the fragment \mz{} and at least one bin wide. With 400 peaks, a spectrum has about 6k active bins out of $\sim$10$^6$.

\noindent\textbf{Bin features.} Each active bin $s$, with prefix mass $x=s\Delta$, gets a vector $\bx(s)\in\R^{86}$ of
physics-derived features (Appendix~\ref{app:features}). For each of 23 ion types $j$ (b and y ions of charge 1--3, a
ions, and neutral losses of water, ammonia and phosphoric acid, among others), the peak nearest to the ion's theoretical \mz{}
$\mu_j(s)$ within 50\,ppm gives three numbers: its encoded intensity, its proximity $\exp(-|\epsilon|/10)$ for an error
of $\epsilon$\,ppm, and the signed error. The proximity is PointNovo's soft matching: a near miss counts partially
instead of being cut off by a hard tolerance. Eight isotope features compare the peaks one isotope spacing above and
below the four primary ions (b$^{+}$, y$^{+}$, b$^{2+}$, y$^{2+}$) with an averagine envelope, and nine global features
give the bin's relative position, the precursor mass, whether b$^{+}$ and y$^{+}$ fall inside the observed \mz{} range,
and the charge.

\noindent\textbf{Scorer.} An MLP $f_\theta$ with three hidden layers of 64 ReLU units maps the features to an evidence vector,
$\bv(s)=f_\theta(\bx(s))\in\R^{8}$ for $s\in\mathcal{A}(X)$, and every inactive bin gets the learned vector $\bv_0\in\R^{8}$.

\noindent\textbf{Edge potentials.} Edge $k$ enters bin $s$ from bin $s-o_k$ and scores
\begin{equation}
\psi_k(s)=\underbrace{\langle\bv(s),\be_k\rangle}_{\text{destination}}
+\underbrace{\langle\bv(s-o_k),\bg_k\rangle}_{\text{source}}
+\underbrace{c_k(X)}_{\text{prior}},
\label{eq:pot}
\end{equation}
with learned destination and source embeddings $\be_k,\bg_k\in\R^8$. We call the three parts the destination term, the
source term and the prior. Stacking the 24 vectors $\be_k$ as the rows of
$\mE\in\R^{24\times8}$ gives all destination terms of bin $s$ at once as $\mE\bv(s)$, a rank-8 factorization of the
bins-by-edges score matrix. Rank 8 lets the residue change how the evidence at its two flanking cleavages is read, e.g.\
the enhanced cleavage N-terminal to proline (the source bin of a P edge). An edge between two inactive bins has the same
score, $\bar\psi_k=c_k(X)+\langle\bv_0,\be_k+\bg_k\rangle$, everywhere in the spectrum.

\noindent\textbf{Residue prior.} The scalar $c_k(X)=\gamma_k+[h_\phi(\bu)]_k$ adds a spectrum-conditioned prior to a static bias
$\gamma_k$.
The input $\bu\in\R^{31}$ holds the encoded intensities at the 24 immonium-ion \mz{} values $m_k-m_{\mathrm{CO}}+\mH$
(within 20\,ppm) and 7 globals (precursor mass, log peak count, charge one-hot). Immonium ions tell which residues are
present but not where, so they enter as a prior rather than as per-bin evidence. The MLP $h_\phi$ ($31{\to}64{\to}24$, GELU)
has its last layer initialized to zero, so training starts from the static prior. The CRF's only sequence prior is
thus a per-residue frequency, with no notion of which residue follows which. The rerankers (\S\ref{sec:rerank}) add
sequence context, but only among the CRF's candidates.

\begin{table}[t]
\centering\footnotesize
\setlength{\tabcolsep}{5pt}
\caption{NovoCRF's learnable parameters, counted from the model's checkpoints.}
\label{tab:params}
\begin{tabular}{@{}llr@{}}
\toprule
Component & Shape & Parameters\\
\midrule
Scorer MLP $f_\theta$ & $86{\to}64{\to}64{\to}64{\to}8$ & 14{,}408\\
Residue prior $h_\phi$, biases $\gamma_k$ & $31{\to}64{\to}24$ and 24 & 3{,}632\\
Edge embeddings $\be_k,\bg_k$ & $2\times24\times8$ & 384\\
Start edges $\be^{\st}_j,\gamma^{\st}_j$, $\bv_0$, $\kappa$ & $3\times9$, 8, 1 & 36\\
User-modification offsets & $8+8+1$ & 17\\
\textbf{CRF total} & & \textbf{18{,}477}\\
\addlinespace[2pt]
First stage reranker & BiGRU-64 & 70{,}626\\
Second stage reranker & BiGRU-128 & 219{,}714\\
Third stage reranker, two calibrators & BiGRU-128, logistic & 222{,}350\\
Fragment-intensity predictor & BiGRU, $2\times192$ & 1{,}121{,}253\\
iRT predictor & BiGRU, $2\times64$ & 148{,}033\\
\midrule
\textbf{Total} & & \textbf{1{,}800{,}453}\\
\bottomrule
\end{tabular}
\end{table}

\noindent\textbf{Path score.} A start edge $j$ landing at bin $q_j$ scores
$\psi^{\st}_j=\langle\bv(q_j),\be^{\st}_j\rangle+\gamma^{\st}_j$, and the end bin $t$ pays a penalty for its distance from
the measured mass,
\begin{equation}
\tau(t)=-\kappa\left(\frac{M-m_{\mathrm{H_2O}}-t\Delta}{0.02\Da}\right)^{2},
\label{eq:tau}
\end{equation}
with a learned $\kappa>0$. A path $\pi$ that optionally begins with start edge $j$ and then takes edges
$k_1,\dots,k_L$ through bins $s_1<\dots<s_L$ scores $\Phi(\pi)=\psi^{\st}_j+\sum_{l=1}^{L}\psi_{k_l}(s_l)+\tau(s_L)$, and
\begin{equation}
P(\pi\mid X)=\frac{\exp\Phi(\pi)}{Z(X)},\qquad Z(X)=\sum_{\pi'\in\Pi(X)}\exp\Phi(\pi'),
\label{eq:crf}
\end{equation}
where $\Pi(X)$ is the set of all mass-feasible paths. The CRF layer itself has no parameters
(Table~\ref{tab:params}).

\subsection{Training: computing $\log Z$ exactly}\label{sec:train}
A labeled peptide determines exactly one path $\pi^\ast$, so the loss (the negative log-likelihood) is
$\ell=\log Z(X)-\Phi(\pi^\ast)$: one forward pass plus a gather of $L$ scores, with no alignment to marginalize, unlike
CTC. With $\alpha(0)=0$, the forward recursion is
\begin{equation}
\begin{aligned}
\alpha(s)&=\log\Big[\textstyle\sum_{k}e^{\alpha(s-o_k)+\psi_k(s)}+\sum_{j:\,q_j=s}e^{\psi^{\mathrm{st}}_j}\Big],\\
\log Z&=\log\textstyle\sum_{|t-t^\ast|\le W}e^{\alpha(t)+\tau(t)}.
\end{aligned}
\label{eq:fwd}
\end{equation}
The recursion looks sequential over $\sim$10$^6$ bins, but every edge spans at least $o_{\min}=28{,}575$ bins (glycine,
57.02\Da). All bins of a block of $o_{\min}$ consecutive bins thus depend only on earlier blocks, and the scan takes
$\lceil S/o_{\min}\rceil\approx35$ sequential steps for a precursor of 2\,kDa, each updating $\sim$28.5k bins $\times$ $K$
edges in parallel. Our Triton~\cite{tillet2019} kernels fuse the scan with Eq.~\eqref{eq:pot}, so the dense
$B\times K\times S$ tensor of edge scores never exists. A backward recursion over the same blocks gives the gradient of
$\log Z$ with respect to every edge score, which is that edge's posterior probability (Appendix~\ref{app:train}).

The closest transformer model, $\pi$-PrimeNovo~\cite{zhang2025}, is also trained on the probability of a whole sequence,
through a CTC loss, but that probability is normalized differently. CTC sums, over all alignments of the label to the
decoder's 40 output positions, the product of per-position softmax probabilities. Each softmax sums to one over the
vocabulary at its own position, whatever the other positions hold. The sequence probabilities that CTC derives from
them therefore sum to one over all sequences of up to 40 residues, whatever their mass. The
precursor mass is an input of the network but does not restrict that sum. NovoCRF's $\log Z$ sums over the mass-feasible peptides only, every edge is scored at the
prefix mass where it starts, and a label is one path, not a set of alignments.

\subsection{Inference: Viterbi and exact $k$-best}\label{sec:inference}
Replacing log-sum-exp by max in Eq.~\eqref{eq:fwd} gives the Viterbi algorithm~\cite{viterbi1967}: $\alpha(s)$ becomes
the best prefix score of bin $s$, and the arg-max path is the exact maximum a posteriori peptide, not a beam approximation. On the CPU,
the scan also proceeds in blocks of $o_{\min}$ consecutive bins (\S\ref{sec:train}). An edge $k$ between two inactive
bins has the same score $\bar\psi_k$ everywhere in the spectrum (\S\ref{sec:potentials}). For each of the $K$ edges,
one loop therefore updates all bins of a block, $\alpha(s)\leftarrow\max\big(\alpha(s),\,\alpha(s-o_k)+\bar\psi_k\big)$.
The loop has no branches and runs over consecutive bins, so the processor's vector instructions update many bins at
once. The active bins are then rescored exactly, and back-pointers are recomputed
along the best path rather than stored. To obtain the $k$ best paths ($k=1{,}000$, or 2{,}500 for precursors above
1{,}800\Da), a best-first (A$^\ast$)~\cite{hart1968} search runs backward from the terminal window with $\alpha(s)$ as an
exact heuristic, so complete paths leave the queue in descending score order, and $\log Z$ is not needed. The search
drops paths of more than 48 residues. Permutations of residues inside a stretch without b or y ions often score exactly the same, and such ties are broken
deterministically. Disabling a variable modification deletes its edges, which keeps search
and normalization exact over the allowed peptides, unlike masking the per-step softmax of an autoregressive decoder.

$\pi$-PrimeNovo enforces the precursor mass only at decoding\footnote{From its code as packaged in denovo\_benchmarks
(\url{https://github.com/bittremieuxlab/denovo_benchmarks}): \texttt{model.py} and \texttt{mass\_con.py}.}. It
first decodes without the mass constraint. If that answer misses the precursor by 0.1\Da\ or more, its precise mass
control (PMC) runs a knapsack dynamic program over output positions and 1\Da\ mass bins, which returns a sequence
within 0.1\Da\ of the precursor or, when it finds none, leaves the unconstrained answer in place. PMC keeps one path
per bin and position and checks the 0.1\Da\ tolerance only for paths that reach the precursor's bin, and it maximizes
the score of a single alignment rather than the CTC probability of the sequence, so its answer need not be the most
probable sequence within tolerance. NovoCRF decodes on the lattice it was trained on: every path fits the precursor
window by construction, and Viterbi returns the exact best one.

\subsection{Reranker cascade}\label{sec:rerank}
The CRF on the prefix-mass lattice cannot order residues inside a gap without cleavage evidence and has no sequence
context. As in
$N$-best rescoring for speech recognition~\cite{schwartz1990}, a cascade of three rerankers rescores the CRF's
candidates, each a larger network on a shorter list (Fig.~\ref{fig:pipeline}a): the first stage reranker scores all
$k$ of the CRF's candidates and passes on its top 100, the second stage reranker passes on its top 20, and the third
stage reranker picks the answer.

\noindent\textbf{Shared design.} Each residue of a candidate is represented by a 32-d token embedding concatenated with a 32-d
ReLU projection of the evidence at that residue, and a one-layer BiGRU reads the candidate, with 64, 128
and 128 units per direction in the three rerankers (Appendix~\ref{app:rerank}). Its mean- and max-pooled states and a
few candidate-level inputs feed an MLP with one hidden layer. The candidate's score is the MLP's output plus a learned
multiple of its CRF score gap to the list's best. Each reranker's score replaces the previous one. The rerankers are
trained one after another, by the softmax cross-entropy of the correct candidate over each list, skipping lists
without it. The first and second stage rerankers are trained on the CRF's candidate lists, and the third stage
reranker on the lists that the second stage reranker passes on. The third stage reranker is first trained with 384
units and then distilled into the 128-unit network (Appendix~\ref{app:training}).

\noindent\textbf{What each reranker reads.} The first stage reranker reads, for each residue, the 86 bin features of
\S\ref{sec:potentials} at the cleavage that follows it, and eight candidate inputs: the CRF score gap, the length, five
spectrum-fit statistics (explained b/y intensity, explained top-20 peaks, share of cleavages with b/y support, longest
uncovered run and explained internal-fragment intensity) and an input fixed at $-50$, which acts as an intercept. The
second stage reranker adds, for each residue, its
immonium ion and the most intense internal fragments of 1--4 residues that start or end at it. An internal fragment
is cut out of the middle of the peptide by two cleavages, so its mass is that of a run of consecutive residues. Inside
a stretch without b or y ions, the lattice sees only the stretch's total mass and scores every order of its residues
the same. Internal fragments tell these orders apart: if a stretch could be RAS or SRA, an internal ion that weighs
A\,+\,S supports RAS, and one that weighs S\,+\,R supports SRA. The third
stage reranker reads internal fragments of up to 6 residues and, at each cleavage, the predicted and observed b$^+$ and y$^+$
intensities (\S\ref{sec:ipred}). Its 27 candidate inputs add the second stage reranker's score and rank, its fit
statistics relative to the second stage reranker's top candidate, five list-contrast features (the peaks it explains
that this top candidate does not, and the reverse) and seven similarity features between its predicted and observed
intensities. The full input lists are in Appendix~\ref{app:rerank}.

\noindent\textbf{Confidence.} NovoCRF uses two logistic calibrators (Appendix~\ref{app:rerank}). The first has 15
weights. It maps the chosen candidate's log-softmax among the third stage reranker's scores, that reranker's top-2
margin, the same two of the second stage reranker, and candidate inputs to a confidence, which we call the calibrated
confidence. NovoCRF's own decisions, such as which answers count as confident (\S\ref{sec:otf}), read this value. The
second has 29 more weights and gives the reported confidence. It also reads the evidence of the RT term
(\S\ref{sec:otf}) and of the sibling fusion (\S\ref{sec:sibling}), for example the number of siblings that agree with
the answer.

\subsection{Fragment-intensity predictor}\label{sec:ipred}
Fragment intensities depend on the sequence, and so tell apart candidates that explain the same peaks. As in
Prosit~\cite{gessulat2019}, a two-layer BiGRU (192 units per direction) reads the residues of a peptide,
and an MLP maps the two states that flank each cleavage to the square-root intensities of five ions: b$^{+}$,
b$^{2+}$, y$^{+}$, y$^{2+}$ and a$^{+}$. A residue enters as its amino acid's embedding plus, if it is modified, an
embedding of the modification and a small network of the modification's mass shift. The predictor's
vocabulary includes 28 modifications alongside the base residues (Appendix~\ref{app:training}). Any other modification, such as a user-defined one (\S\ref{sec:userptm}), is read as an
unknown modification with its mass shift. The N-terminal modification is encoded the same way, and embeddings of the
precursor charge and of the normalized collision energy (NCE) are added at every position.

The predictor is trained to maximize the normalized spectral angle between the predicted and observed intensities, on
the higher-energy collisional dissociation (HCD) spectra of PROSPECT PTMs~\cite{gabriel2024,prospectms2}, whose NCE is
known, mixed with NovoCRF's own training spectra (\S\ref{sec:protocol}). It has 1.12M of NovoCRF's 1.80M parameters
(Table~\ref{tab:params}), and the third stage reranker is the only reranker that reads it. The engine takes each spectrum's NCE
from the raw file, converting timsTOF collision energies from eV. Since the
same nominal NCE fragments differently on different instruments, on-the-fly adaptation calibrates the collision energy that
the predictor reads and fine-tunes its two output layers on each run (\S\ref{sec:otf}).

\subsection{Retention-time predictor}\label{sec:irt}
A peptide's retention time depends on its sequence. On the iRT scale, which is defined by reference
peptides, it is nearly independent of the run, and each run maps iRT to minutes by its own
curve~\cite{escher2012}. NovoCRF's iRT predictor uses the intensity predictor's residue encoding, with 15 named
modifications, a two-layer BiGRU (64 units per direction), mean and max pooling over the residues, and an
MLP that outputs the iRT. It reads neither charge nor collision energy. It has 148k parameters and is trained with an
L1 loss on the PROSPECT PTMs retention-time dataset~\cite{gabriel2024,prospectirt}, on whose test split its mean
absolute error is 2.9 iRT units (Appendix~\ref{app:training}). On each run, the predictor is fine-tuned and a per-run
regression maps its output to the run's retention times (\S\ref{sec:otf}).

\subsection{Modifications never seen in training}\label{sec:userptm}
A modification the model was never trained on can still be searched exactly. A user-defined modification that is
unseen in the training set, such as C+119.004, adds one edge at the unmodified residue's mass plus the delta, so the lattice, Viterbi and $k$-best search stay exact.
The edge is scored with the trained parameters of its unmodified residue, the embeddings $\be_k,\bg_k$ and the prior
$c_k$, plus three offsets shared by all user modifications: two 8-d vectors added to the embeddings and a scalar added
to the prior, which an optional user offset can lower for fewer calls. The rerankers read the unmodified residue's
token, to which the third stage reranker adds a learned vector that marks a user modification. The offsets and that vector come from
synthetic-modification training: in a quarter of the training spectra, one residue of the label gets a random mass
shift, either a known modification's or one drawn uniformly from $-20$ to $+250\Da$. Every fragment peak that contains
it, and the precursor, move with it, and the lattice gets an edge for it scored as above (Appendix~\ref{app:training}).
The model thus learns how a modified residue looks in general rather than any one modification. The intensity and
iRT predictors read a user-defined modification by name if it is one of their named modifications, and otherwise as
an unknown modification with its mass shift (\S\S\ref{sec:ipred}--\ref{sec:irt}). A modification within one bin (0.002\Da)
of an existing residue is skipped.

\subsection{On-the-fly adaptation}\label{sec:otf}
Unless told otherwise, NovoCRF decodes each run twice (Fig.~\ref{fig:pipeline}b). The first pass is the single pass
of Fig.~\ref{fig:pipeline}a. Its confident answers serve as the run's own training data: the answers among the 30\% with the
highest calibrated confidence that have at least six residues and leave at most one bond without a b or y ion. With
fewer than 1{,}000 of them, the spectra are not decoded again, and the first pass's candidates only get the final
RT term and sibling fusion described below. Otherwise NovoCRF adapts the following to the run and decodes
every spectrum again from scratch:

\noindent\textbf{Mass calibration.} The precursor masses are corrected by the run's systematic error. It is the median of the
per-peptide errors in their densest 4\,ppm cluster, which must stand out from the errors around it. A cubic
polynomial in \mz{} replaces this value when it lowers the spread of the errors by at least 10\%. The fragment \mz{}
values are corrected by a polynomial of degree 4 in \mz{}, fitted to the b and y ions of the confident answers.

\noindent\textbf{Collision-energy calibration.} The intensity predictor reads each spectrum's NCE, but the NCE set in the
method is only nominally comparable between instruments. NovoCRF therefore calibrates the NCE
that the predictor is given. On up to 4{,}000 confident answers, it picks the setting under which the unadapted
predictor matches the observed intensities best: an offset of $-8$ to $+8$ added to the raw file's NCE or, for runs
without one, a single NCE for all spectra, or an unknown NCE.

\noindent\textbf{CRF fine-tuning.} The scorer, $\bv_0$ and the edge embeddings and priors (14{,}824 of the CRF's 18{,}477
parameters) are fine-tuned on the recalibrated spectra, with the offsets of user modifications when these are
searched. The loss is the training loss, with the exact $\log Z$ over the lattice that decoding uses. Training
uses Adam at a learning rate of $10^{-3}$ (0.3$\times$ for the scorer's lower layers), batches of 32, two epochs.

\noindent\textbf{Predictor fine-tuning.} The intensity predictor's two output layers (149k parameters) are fine-tuned
with its spectral-angle loss, at the calibrated collision energy and with its embeddings and recurrent layers frozen: Adam
at $10^{-3}$, batches of 64, two epochs.

Both fine-tunings also train on the answers outside the confident set that a sibling spectrum inside it confirms
(\S\ref{sec:sibling}). Each fine-tuned model is kept only if it improves on held-out confident answers: the CRF's
negative log-likelihood, the predictor's spectral angle. The rerankers are not fine-tuned.

\noindent\textbf{Retention time.} After each pass, NovoCRF fits an RT regression on the run's high-confidence answers,
those with a calibrated confidence of at least 0.9 (at least 500 are needed). Its 66 features are the iRT that the predictor of \S\ref{sec:irt}
gives each sequence, standardized over the run, with its square and cube, and 63 sequence features, namely
composition, hydrophobicity and charge indices, length, the N-terminal residue and modification counts
(Appendix~\ref{app:otf}). The cubic in iRT maps the predictor's scale to the run's gradient, and the sequence features
correct what the predictor misses, such as the retention shift of a user-defined modification that it reads as
unknown. The regression is a robust ridge regression, cross-fitted over five folds. After the last pass, the iRT
predictor itself is first fine-tuned on the run: its second recurrent layer and its output layers are trained on the
retention times of the same sequences, and the tuned copy is kept only if it lowers the spread of the held-out
residuals by at least 5\%. Each of the third stage reranker's 20 candidates then gets a penalty term for its deviation
$\delta=t_{\mathrm{obs}}-t_{\mathrm{pred}}$. The log-density of a Student-$t$ distribution fitted to the held-out
residuals, relative to its value at zero, is added to the candidate's final score as a penalty, the RT term:
\begin{equation}
\rho(\delta)=-\frac{\nu+1}{2}\,\ln\!\Big(1+\frac{\delta^{2}}{\nu\,\omega^{2}}\Big),
\label{eq:rt}
\end{equation}
with the scale $\omega$ and the degrees of freedom $\nu$ chosen by maximum likelihood. Retention time is thus soft
evidence, scaled by how well it is predicted in each run, and the heavy tails of the distribution keep a peptide with
a poorly predicted retention time in play. A run without retention times or with too few high-confidence answers
gets no RT term. None of these steps uses labels or a database.

\subsection{Sibling spectra enhancement}\label{sec:sibling}
A run often holds several spectra of one peptide: the precursor is selected again later in its elution, or at another
charge state. These \emph{sibling spectra} rarely miss the same fragments, so a bond that one of them leaves without
a b or y ion may be covered in another. Sequencing every spectrum on its own does not use this. NovoCRF links the
siblings and lets them agree on an answer (Fig.~\ref{fig:pipeline}c). We call this step the sibling fusion.

\noindent\textbf{Links.} Two spectra of a run are siblings when their neutral precursor masses agree within 10\,ppm,
whatever their charges, and they elute within 120\,s of each other. Each spectrum keeps at most its 12 nearest
siblings. The links are made from the measured masses after the first pass, and again, from the measured and the
recalibrated masses, after the second.

\noindent\textbf{Exchange of candidates.} The spectra are not merged. Every spectrum keeps its own search and its own
list of 20 candidates from the third stage reranker, and each of its siblings offers it its five best candidates. An offered
candidate that fits the spectrum's precursor is scored on the spectrum's own peaks, as if the search had found it
there: by the CRF on the spectrum's lattice, by the second and third stage rerankers, and by the RT term at the
spectrum's retention time.

\noindent\textbf{Fused score.} Let $a_i(c)$ be the third stage reranker's score of candidate $c$ on spectrum $i$, with the RT
term, $\mathcal{L}_i$ the extended list of spectrum $i$, and $\mathcal{S}(i)$ its siblings. The fused score is
\begin{equation}
\begin{aligned}
F_i(c)={}&\underbrace{a_i(c)-\log\!\sum_{c'\in\mathcal{L}_i}\!e^{a_i(c')}}_{\text{own spectrum}}\\
&+\sum_{j\in\mathcal{S}(i)}\underbrace{\max\!\Big(a_j(c)-\max_{c'\in\mathcal{L}_j}a_j(c'),\,-5\Big)}_{\text{support of sibling }j},
\end{aligned}
\label{eq:fuse}
\end{equation}
and the answer of spectrum $i$ is the candidate with the highest $F_i$. The support of a sibling is zero for the
sibling's own best candidate and falls to $-5$ for a candidate that the sibling ranks far lower or does not list. The
floor bounds what one sibling can do: it changes an answer only where the spectrum's own scores of the two candidates
differ by less than 5. A well-fragmented spectrum thus keeps its answer, and a poorly fragmented one follows its
siblings. All spectra are decided from the scores before any answer changes.

\noindent\textbf{Where it runs.} The fusion follows the RT term of each pass (Fig.~\ref{fig:pipeline}a). After the
first pass it also enlarges the run's training data: an answer outside the confident set joins the fine-tuning
examples of \S\ref{sec:otf} when a sibling inside the set has the same peptide. The reported confidence takes the
agreement of the siblings into account (\S\ref{sec:rerank}). On the CPU alone, each sibling offers only its best
candidate, which saves time.

\subsection{Cross-run spectra enhancement}\label{sec:xrun}
Replicate runs of one sample generate multiple spectra of the same peptides. When several
runs are given together and cross-run spectra enhancement (CRSE) is switched on, the sibling links extend across the
runs: a spectrum of one run and a spectrum of another support each other when they have the same precursor mass and
similar retention times on a common time scale.

\noindent\textbf{Aligning two runs.} Every run first goes through its own first pass. NovoCRF then aligns the
retention times of each pair of runs. It takes the peptides that both runs identified among their 30\% most confident
answers. Each of these peptides gives one pair of retention times, one from each run. A map from the first run's
retention time to the second run's is fitted to these pairs. The map is a running median of the paired times. It is
made non-decreasing and is interpolated linearly. Let $\sigma$ be the robust spread of the pairs around the map. A
spectrum of the first run and a spectrum of the second run can be linked when the mapped time of the first lies
within $3\sigma$ of the time of the second. This limit is kept between 1 and 8 iRT units. The iRT scale of the first
run converts these two bounds to seconds, so one rule serves every gradient length. A pair of runs with fewer than
50 such shared peptides gets no links. This keeps the runs of other samples apart.

\noindent\textbf{Cross-run siblings.} Two spectra of different runs are siblings when two conditions hold. Their
precursor masses agree within 10\,ppm, after each is corrected by its own run's systematic mass error. Their
retention times agree within the limit above. Each spectrum keeps its eight nearest cross-run siblings. Cross-run
siblings enter Eq.~\eqref{eq:fuse} like the siblings within the run. They take part in the fusion after the first
pass and in the final one. An answer that a cross-run sibling confirms also joins the fine-tuning examples of its
run. Every run keeps its own adaptation. A candidate offered across runs is scored on the receiving spectrum with
the receiving run's models and RT term. The support of a cross-run sibling is read from the scores of its own run's
models.

CRSE is off by default. By explicitly turning it on, the user gives NovoCRF the prior knowledge that the input runs
are replicate runs of the same sample.

\subsection{Implementation and model size}\label{sec:impl}
Training runs in PyTorch with the Triton kernels above. Inference is performed in a C++ engine. A GPU is optional: when one with CUDA 12 or later and at least 12\,GB of memory is present,
the engine computes the CRF potentials and the Viterbi scan, the three rerankers, the intensity and iRT predictors and
the fine-tunings on the GPU, while the $k$-best search, the rerankers' evidence features, the calibrations and the RT
regression stay on the CPU. NovoCRF has 1.80M learnable parameters, 18 to 53 times fewer than the transformer models we
compare with (\S\ref{sec:speed}).

% ================================================================================================
\section{Experiments}\label{sec:results}

\subsection{Data and protocol}\label{sec:protocol}

\noindent\textbf{Training.} The CRF and the three rerankers are trained from scratch on public data only: 10M labeled
spectra that keep every one of the 3.26M peptides of their sources, with about $\sqrt{n}$ of the $n$ spectra of each
peptide. The sources are MassIVE-KB~\cite{wang2018} (4.97M spectra), InstaNovo's multi-enzyme
set~\cite{eloff2025} (2.95M), CompOmics' reprocessed PRIDE projects (0.92M), synthetic ProteomeTools
peptides~\cite{zolg2017} (0.81M) and the phosphoproteomics projects of the InstaNovo-P set~\cite{lauridsen2026}
(0.35M). Common tryptic charge-2/3 spectra are kept with probability 0.4 to favor non-tryptic and charge-1 data. The
iRT predictor is trained on PROSPECT
PTMs~\cite{gabriel2024}, and the intensity predictor on PROSPECT PTMs together with these training spectra
(\S\S\ref{sec:ipred}--\ref{sec:irt}).

One routine trains the components in a fixed order (Appendix~\ref{app:training}): the CRF for 80k steps, the
first and second stage rerankers for 150k steps each, on the CRF's candidate lists of 1.7M and 1.2M training spectra,
and the third stage reranker twice for 150k steps on the lists of 4.6M training spectra, first as a larger teacher
and then as the student distilled from it. Training takes about 21 GPU-hours on one A100 in all: 5.7 for the CRF,
12.0 for the rerankers and their candidate lists, and 3.1 for the two predictors.

\noindent\textbf{Development and test sets.} Design choices were compared on 19 development datasets and on a
validation split of the training data. The development sets cover a wide variety of instruments and MS/MS
applications. They are three human leukocyte antigen (HLA) peptidomes (PXD079646~\cite{catalan2026},
PXD060958~\cite{garcia2026}, PXD073207~\cite{simhadri2026}), a native tear peptidome (PXD072578~\cite{chang2026}), tryptic
digests of zebrafish eye (PXD061443~\cite{maffioli2025}), of a human cell line (PXD070049) and of six non-mammalian
organisms (PXD052473, PXD070102, PXD066296, PXD059278, PXD056405, PXD084437), three antibody sets digested with
several proteases (PXD060500, PXD063526, PXD072803), a phospho-enriched set (PXD068908), and three sets with a
modification outside NovoCRF's vocabulary (PXD065497 and two runs of PXD061971). Fourteen were acquired on Thermo
instruments, from a Q Exactive to an Orbitrap Astral, and five on timsTOF instruments.
The ablations of \S\ref{sec:ablation} are on development sets. All other results below are on test datasets that
are held out from the training data and the development sets (Table~\ref{tab:data}): twelve main test sets, five sets rich in
hydroxyproline and the spike-in data of RNovA.

\begin{table}[t]
\centering\footnotesize
\setlength{\tabcolsep}{0.9pt}
\caption{The test datasets, one raw file each unless noted. Labels: PEAKS DB search, 1\% PSM FDR. QE: Q Exactive, prot.\ K: proteinase K.
$^a$PASEF precursors. $^b$Eight runs, one per protease. $^c$Phosphopeptide-enriched. $^d$Labels from the authors'
MaxQuant search. $^e$Four runs. The labels are the spectra of the group's eight synthetic peptides.}
\label{tab:data}
\begin{tabular}{@{}lllrr@{}}
\toprule
Dataset & Sample & Enzyme, instrument & MS2 & Labels\\
\midrule
\multicolumn{5}{@{}l}{\emph{Main test sets}}\\
PXD076585~\cite{pal2026} & human A549 & trypsin, Exploris & 131{,}789 & 31{,}275\\
PXD071561~\cite{moen2026} & squid optic lobe & Lys-C, Exploris & 59{,}389 & 24{,}701\\
PXD073929~\cite{sachslehner2026} & frog toe & Glu-C, Exploris & 103{,}345 & 2{,}276\\
PXD072115~\cite{bera2026} & chickpea peptides & none, timsTOF$^a$ & 32{,}981 & 19{,}267\\
PXD074750~\cite{rodrigues2026} & \emph{Candida auris} & prot.\ K, Exploris & 62{,}389 & 22{,}488\\
PXD055085~\cite{comtois2025} & rabbit antibody & eight$^b$, QE HF & 226{,}262 & 23{,}091\\
PXD071022~\cite{courcelles2026} & HLA-I, colorectal & none, Ascend & 85{,}732 & 21{,}887\\
PXD082505~\cite{hwang2026} & HLA-I, B*15:02 & none, Ascend & 208{,}196 & 4{,}643\\
PXD054498~\cite{pandey2025} & HLA-I, effusion & none, timsTOF$^a$ & 108{,}045 & 25{,}451\\
PXD064980~\cite{dierickx2026} & HLA-II (DR) & none, timsTOF$^a$ & 111{,}840 & 23{,}333\\
PXD062962~\cite{long2025} & mouse liver$^c$ & trypsin, Exploris & 31{,}723 & 16{,}664\\
PXD062859~\cite{lauridsen2026} & human T47D$^c$ & trypsin, Exploris & 38{,}015 & 11{,}963\\
Total & & & 1{,}199{,}706 & 227{,}039\\
\addlinespace[2pt]
\multicolumn{5}{@{}l}{\emph{Hydroxyproline-rich sets}}\\
PXD073112~\cite{rao2026} & Denisovan enamel & none, Lumos & 104{,}944 & 7{,}644\\
PXD073112~\cite{rao2026} & Denisovan dentine & trypsin, Lumos & 46{,}433 & 2{,}782\\
PXD075875~\cite{cordelier2026} & tortoise bone & trypsin, QE & 38{,}135 & 9{,}571\\
PXD065046~\cite{karjalainen2026} & mouse collagen & trypsin, QE HF & 6{,}048 & 1{,}802\\
PXD065297~\cite{sun2026} & lion bone$^d$ & elastase, Exploris & 24{,}784 & 3{,}674\\
\addlinespace[2pt]
\multicolumn{5}{@{}l}{\emph{RNovA's spike-in of peptides with non-natural residues}}\\
PXD076296~\cite{mao2026} & 293T + group 1$^e$ & trypsin, Ascend & 239{,}394 & 508\\
PXD076296~\cite{mao2026} & 293T + group 2$^e$ & trypsin, Ascend & 227{,}547 & 524\\
PXD076296~\cite{mao2026} & 293T + group 3$^e$ & trypsin, Ascend & 226{,}565 & 661\\
\bottomrule
\end{tabular}
\end{table}

No test dataset is part of NovoCRF's training data. We compared the accession numbers of the development and test
projects with the 18 CompOmics projects, the 29 projects of the InstaNovo-P set and the 82 projects of InstaNovo's
multi-enzyme set. MassIVE-KB was assembled in 2018, ProteomeTools and PROSPECT hold synthetic peptides, and every
test project was made public in 2025 or 2026. Peptides, unlike projects, do recur between training and test data, and
\S\ref{sec:seen} splits the results on the twelve main test sets by this overlap. PXD062859 holds the T47D FGFR2-signaling data that InstaNovo's
developers generated to validate InstaNovo-P experimentally~\cite{lauridsen2026}. It is not in the training or the
validation split of InstaNovo-P's fine-tuning data, which come from 29 older projects reprocessed for Scop3P.

\noindent\textbf{Labels.} The labels come from a PEAKS 13.5 database search~\cite{zhang2012} at 1\% peptide-spectrum
match (PSM) false discovery rate (FDR), with the submitters' modifications. The two phospho-enriched sets were searched with
phosphorylation of S, T and Y. Of their labels, 72\% and 91\% are phosphopeptides. \Cfive\ and $\pi$-PrimeNovo have
no phosphorylated residues, so they are not scored on these two sets.

\noindent\textbf{Companion runs for CRSE.} For cross-run spectra enhancement (\S\ref{sec:xrun}), the test run of a
dataset, the run that its labels come from, needs other runs of the same sample. We chose two companion runs for each test set by one rule,
from the project's description, its sample table and its file names, without reading a label or an identification
result. A companion comes from the test run's project. It is a replicate injection of the test run's sample or a
replicate sample of the same type and condition, prepared in the same way, and it was acquired on the same instrument
with the same gradient, which we checked on the acquisition length recorded in the raw files. A run does not qualify
if it is another fraction of the sample, another enzyme digest, another fragmentation method or instrument, another
tissue, species or preparation, a mixture with other samples, or a run that the project does not describe well enough
to show what it is. A set without two such runs gets no CRSE result.

Eight of the twelve main test sets have companion runs (Appendix~\ref{app:companions}). For six of them the
companions are replicates: a second injection and a second biological replicate of the same control sample (human),
biological replicates (squid, chickpea, mouse liver and T47D) and repeated loads of the same donor's peptides
(HLA-II). For two they are not replicates in the strict sense, and the tables say so. The colorectal HLA-I set has a
second injection of the same tumor and the tumor of another patient, who shares four of its six HLA-I alleles. The
B*15:02 set has two runs of the same cell line after a drug treatment, because the project holds no second run of
the untreated cells. NovoCRF reads the test run and its two companions in one call with CRSE switched on, with
no setting chosen per dataset, and only the test run's spectra are scored.

\begin{table*}[t]
\centering\footnotesize
\setlength{\tabcolsep}{2.4pt}
\caption{Top-1 peptide recall (\%) on the twelve main test sets, and throughput in spectra/s over the 1{,}199{,}706 MS2
spectra of their raw files on one desktop computer (32 CPU threads, one RTX 3090), where NovoCRF with a GPU uses
the CPU threads as well. The four other tools were timed with the GPU only. \Cfive\ and $\pi$-PrimeNovo have
no phosphorylated residues and are not scored on the two phospho-enriched sets. CRSE: cross-run spectra enhancement,
which also reads two companion runs of the test run's sample (\S\ref{sec:protocol}). A dash marks the four sets
without such runs, and for PXD071022 and PXD082505 the companions are not replicates in the strict sense. Best recall
per column among the methods that read the test run alone in bold. $^a$Over the 1.97M spectra of the three runs of
each of the eight sets.}
\label{tab:main}
\begin{tabular}{@{}l*{12}{r}rr@{}}
\toprule
 & \multicolumn{12}{c}{Peptide recall (\%) on PXD\ldots} & \multicolumn{2}{c}{Spectra/s}\\
\cmidrule(lr){2-13}\cmidrule(l){14-15}
 & 076585 & 071561 & 073929 & 072115 & 074750 & 055085 & 071022 & 082505 & 054498 & 064980 & 062962 & 062859 & with & CPU\\
Method & human & squid & frog & chickpea & \emph{C.\ auris} & antibody & HLA-I & HLA-I & HLA-I & HLA-II & phospho & phospho & GPU & only\\
\midrule
InstaNovo & \textbf{53.1} & 75.7 & 44.3 & 10.6 & 30.6 & 67.5 & 66.3 & 62.5 & 33.5 & 9.9 & 74.2 & 77.4 & 15.3 & --\\
RNovA & 39.0 & 73.3 & 38.4 & 7.6 & 29.3 & 57.1 & 54.9 & 52.9 & 28.9 & 4.3 & 67.6 & 64.0 & 9.35 & --\\
$\pi$-PrimeNovo & 41.0 & 75.2 & 40.5 & 11.9 & 32.1 & 64.6 & 54.0 & 56.8 & 23.7 & 7.5 & -- & -- & 136 & --\\
\Cfive & 35.0 & 65.4 & 40.9 & 8.7 & 24.1 & 53.9 & 50.7 & 49.4 & 16.8 & 3.9 & -- & -- & 28.2 & --\\
\addlinespace[2pt]
NovoCRF, one pass & 48.2 & 80.5 & 45.7 & 43.7 & 44.3 & 69.3 & 81.4 & 84.0 & 72.0 & 34.0 & 80.2 & 82.2 & 676 & \cpuone\\
\textbf{NovoCRF}, two passes & 51.5 & \textbf{82.7} & \textbf{51.0} & \textbf{50.1} & \textbf{52.7} & \textbf{73.4} & \textbf{83.2} & \textbf{85.7} & \textbf{80.0} & \textbf{43.7} & \textbf{82.2} & \textbf{82.9} & 313 & \cputwo\\
\addlinespace[2pt]
NovoCRF, two passes\,+\,CRSE & 59.7 & 83.8 & -- & 51.0 & -- & -- & 84.2 & 88.0 & -- & 47.4 & 83.5 & 84.0 & 162$^a$ & \cpux$^a$\\
\bottomrule
\end{tabular}
\end{table*}
\begin{figure*}[t]
\centering
\includegraphics[width=\textwidth]{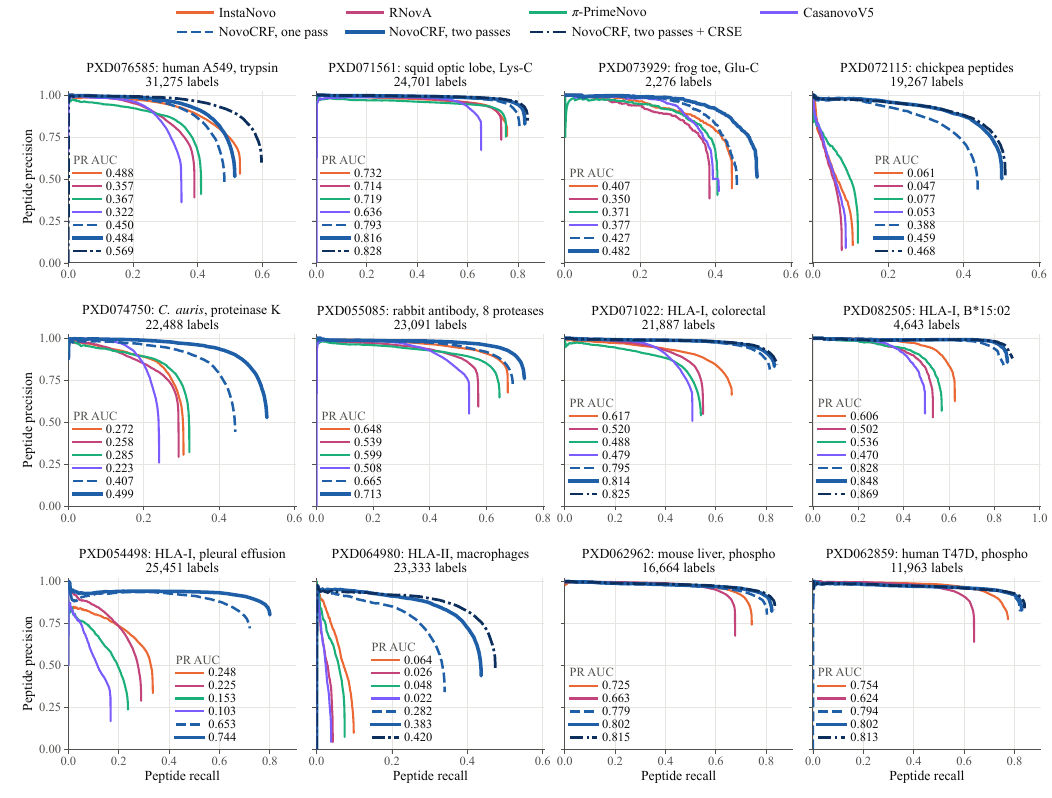}
\caption{Peptide-level precision--recall curves on the twelve main test sets. Each curve ranks a method's answers on
the labeled spectra by the method's own score. Precision is the share of correct answers among the answers given
so far, and recall is their share of all labeled spectra. Each panel lists the area under every curve (PR AUC), in the
order of the legend, and its recall axis ends just above the highest recall. \Cfive\ and $\pi$-PrimeNovo are not scored
on the two phospho-enriched sets, and NovoCRF with CRSE is shown for the eight sets with companion runs.}
\label{fig:pr}
\end{figure*}
\begin{table*}[t]
\centering\footnotesize
\setlength{\tabcolsep}{2.9pt}
\renewcommand{\arraystretch}{0.95}
\caption{Peptide recall (\%) by the number of the label's peptide bonds without a b or y ion, for each main test set.
Labels: labeled spectra in the bin. CRSE: with two companion runs, where a set has them. Best per column among the
methods that read the test run alone in bold. For PXD055085, 21 labels whose spectrum is not in the peak lists
count in the last column only.}
\label{tab:missing}
\begin{tabular}{@{}l*{7}{r}@{\hspace{12pt}}*{7}{r}@{}}
\toprule
 & \multicolumn{7}{c}{PXD076585: human A549, trypsin} & \multicolumn{7}{c}{PXD071561: squid optic lobe, Lys-C}\\
\cmidrule(lr){2-8}\cmidrule(l){9-15}
Missing bonds & 0 & 1 & 2 & 3 & 4--5 & $\ge$6 & All & 0 & 1 & 2 & 3 & 4--5 & $\ge$6 & All\\
Labels & 3{,}492 & 5{,}231 & 4{,}291 & 3{,}667 & 5{,}219 & 9{,}375 & 31{,}275 & 9{,}126 & 5{,}813 & 4{,}408 & 1{,}604 & 1{,}394 & 2{,}356 & 24{,}701\\
\cmidrule(r){1-1}\cmidrule(lr){2-8}\cmidrule(l){9-15}
InstaNovo & 94.6 & 86.6 & 70.2 & 51.3 & \textbf{35.1} & \textbf{21.8} & \textbf{53.1} & 94.5 & 88.1 & 80.3 & 54.1 & 29.1 & 5.5 & 75.7\\
RNovA & 92.7 & 80.3 & 57.6 & 34.5 & 15.7 & 2.1 & 39.0 & 93.4 & 86.3 & 79.1 & 48.9 & 20.2 & 1.1 & 73.3\\
$\pi$-PrimeNovo & 89.6 & 78.2 & 58.7 & 38.6 & 20.9 & 6.3 & 41.0 & 91.8 & 86.1 & 82.1 & 58.7 & 35.5 & 5.3 & 75.2\\
\Cfive & 84.0 & 70.7 & 47.1 & 29.8 & 16.1 & 3.8 & 35.0 & 85.9 & 76.7 & 66.9 & 36.9 & 17.4 & 2.6 & 65.4\\
NovoCRF, one pass & 94.4 & 84.9 & 69.8 & 48.9 & 30.2 & 10.5 & 48.2 & 95.9 & 92.6 & 85.9 & 65.2 & 45.0 & 12.9 & 80.5\\
\textbf{NovoCRF}, two passes & \textbf{95.7} & \textbf{88.3} & \textbf{74.3} & \textbf{55.5} & 34.8 & 11.6 & 51.5 & \textbf{96.9} & \textbf{94.1} & \textbf{88.6} & \textbf{70.6} & \textbf{51.6} & \textbf{14.9} & \textbf{82.7}\\
NovoCRF, two passes\,+\,CRSE & 96.4 & 91.6 & 82.8 & 67.4 & 50.3 & 20.1 & 59.7 & 97.2 & 94.6 & 89.9 & 74.9 & 55.7 & 16.5 & 83.8\\
\addlinespace[5pt]
 & \multicolumn{7}{c}{PXD073929: frog toe, Glu-C} & \multicolumn{7}{c}{PXD072115: chickpea peptides}\\
\cmidrule(lr){2-8}\cmidrule(l){9-15}
Missing bonds & 0 & 1 & 2 & 3 & 4--5 & $\ge$6 & All & 0 & 1 & 2 & 3 & 4--5 & $\ge$6 & All\\
Labels & 259 & 344 & 319 & 292 & 413 & 649 & 2{,}276 & 909 & 2{,}170 & 2{,}818 & 2{,}617 & 3{,}930 & 6{,}823 & 19{,}267\\
\cmidrule(r){1-1}\cmidrule(lr){2-8}\cmidrule(l){9-15}
InstaNovo & 96.5 & 86.0 & 65.5 & 43.2 & 23.7 & 4.6 & 44.3 & 41.3 & 30.9 & 20.2 & 10.0 & 3.6 & 0.2 & 10.6\\
RNovA & 93.8 & 84.9 & 57.1 & 33.2 & 12.8 & 1.1 & 38.4 & 32.9 & 25.5 & 13.9 & 6.2 & 1.6 & 0.1 & 7.6\\
$\pi$-PrimeNovo & 95.8 & 82.3 & 60.5 & 35.3 & 17.9 & 3.2 & 40.5 & 42.6 & 34.7 & 22.7 & 13.7 & 3.7 & 0.2 & 11.9\\
\Cfive & 95.4 & 81.4 & 61.4 & 39.4 & 17.7 & 2.9 & 40.9 & 34.7 & 27.6 & 16.2 & 7.6 & 2.5 & 0.2 & 8.7\\
NovoCRF, one pass & 97.3 & 89.0 & 68.0 & 45.5 & 23.0 & 5.7 & 45.7 & 84.9 & 77.1 & 68.3 & 56.1 & 39.0 & 15.4 & 43.7\\
\textbf{NovoCRF}, two passes & \textbf{98.8} & \textbf{92.2} & \textbf{74.6} & \textbf{56.5} & \textbf{28.8} & \textbf{10.2} & \textbf{51.0} & \textbf{89.5} & \textbf{83.6} & \textbf{76.8} & \textbf{64.7} & \textbf{46.2} & \textbf{19.7} & \textbf{50.1}\\
NovoCRF, two passes\,+\,CRSE & -- & -- & -- & -- & -- & -- & -- & 89.2 & 83.0 & 77.0 & 64.3 & 48.4 & 21.4 & 51.0\\
\addlinespace[5pt]
 & \multicolumn{7}{c}{PXD074750: \emph{C.\ auris}, proteinase K} & \multicolumn{7}{c}{PXD055085: rabbit antibody, eight proteases}\\
\cmidrule(lr){2-8}\cmidrule(l){9-15}
Missing bonds & 0 & 1 & 2 & 3 & 4--5 & $\ge$6 & All & 0 & 1 & 2 & 3 & 4--5 & $\ge$6 & All\\
Labels & 2{,}629 & 3{,}938 & 3{,}891 & 2{,}782 & 3{,}548 & 5{,}700 & 22{,}488 & 6{,}007 & 6{,}433 & 3{,}484 & 2{,}310 & 2{,}083 & 2{,}753 & 23{,}091\\
\cmidrule(r){1-1}\cmidrule(lr){2-8}\cmidrule(l){9-15}
InstaNovo & 76.4 & 59.4 & 41.3 & 20.5 & 8.3 & 0.9 & 30.6 & 81.6 & 82.2 & 70.6 & 64.4 & 45.7 & 18.4 & 67.5\\
RNovA & 77.1 & 58.4 & 39.0 & 18.1 & 5.9 & 0.4 & 29.3 & 78.6 & 74.3 & 59.0 & 47.6 & 21.1 & 3.5 & 57.1\\
$\pi$-PrimeNovo & 75.0 & 61.2 & 45.6 & 24.1 & 9.8 & 0.8 & 32.1 & 78.4 & 78.5 & 69.3 & 63.0 & 41.9 & 15.3 & 64.6\\
\Cfive & 65.3 & 48.0 & 31.6 & 13.4 & 5.2 & 0.4 & 24.1 & 69.2 & 68.2 & 54.8 & 49.8 & 29.3 & 8.1 & 53.9\\
NovoCRF, one pass & 90.0 & 74.6 & 57.6 & 36.4 & 24.6 & 9.3 & 44.3 & 84.4 & 83.7 & 72.7 & 64.4 & 48.1 & 19.2 & 69.3\\
\textbf{NovoCRF}, two passes & \textbf{93.3} & \textbf{81.9} & \textbf{68.4} & \textbf{51.1} & \textbf{35.9} & \textbf{14.3} & \textbf{52.7} & \textbf{85.8} & \textbf{86.4} & \textbf{78.5} & \textbf{71.5} & \textbf{56.9} & \textbf{23.9} & \textbf{73.4}\\
NovoCRF, two passes\,+\,CRSE & -- & -- & -- & -- & -- & -- & -- & -- & -- & -- & -- & -- & -- & --\\
\addlinespace[5pt]
 & \multicolumn{7}{c}{PXD071022: HLA-I, colorectal} & \multicolumn{7}{c}{PXD082505: HLA-I, B*15:02}\\
\cmidrule(lr){2-8}\cmidrule(l){9-15}
Missing bonds & 0 & 1 & 2 & 3 & 4--5 & $\ge$6 & All & 0 & 1 & 2 & 3 & 4--5 & $\ge$6 & All\\
Labels & 5{,}155 & 5{,}750 & 4{,}105 & 3{,}036 & 2{,}994 & 847 & 21{,}887 & 374 & 1{,}012 & 953 & 754 & 1{,}041 & 509 & 4{,}643\\
\cmidrule(r){1-1}\cmidrule(lr){2-8}\cmidrule(l){9-15}
InstaNovo & 94.3 & 84.9 & 66.9 & 45.2 & 19.4 & 8.3 & 66.3 & 98.1 & 94.9 & 81.4 & 61.7 & 30.1 & 4.5 & 62.5\\
RNovA & 89.8 & 75.8 & 50.0 & 24.1 & 7.7 & 1.5 & 54.9 & 96.5 & 89.1 & 70.7 & 45.9 & 15.5 & 2.2 & 52.9\\
$\pi$-PrimeNovo & 86.8 & 72.5 & 51.3 & 26.5 & 8.5 & 2.5 & 54.0 & 95.2 & 89.7 & 76.5 & 53.3 & 21.6 & 3.3 & 56.8\\
\Cfive & 82.9 & 68.2 & 46.0 & 23.9 & 8.4 & 3.7 & 50.7 & 93.9 & 83.2 & 64.7 & 44.4 & 13.8 & 0.8 & 49.4\\
NovoCRF, one pass & 97.1 & 91.9 & 83.7 & 70.7 & 54.4 & 36.8 & 81.4 & 99.7 & 97.3 & 93.6 & 83.7 & 73.5 & 49.5 & 84.0\\
\textbf{NovoCRF}, two passes & \textbf{97.6} & \textbf{93.1} & \textbf{85.9} & \textbf{74.0} & \textbf{57.4} & \textbf{39.8} & \textbf{83.2} & \textbf{100.0} & \textbf{97.6} & \textbf{93.9} & \textbf{87.1} & \textbf{76.8} & \textbf{52.5} & \textbf{85.7}\\
NovoCRF, two passes\,+\,CRSE & 97.6 & 93.8 & 87.0 & 75.9 & 59.5 & 42.0 & 84.2 & 100.0 & 98.4 & 95.8 & 90.2 & 81.4 & 54.4 & 88.0\\
\addlinespace[5pt]
 & \multicolumn{7}{c}{PXD054498: HLA-I, pleural effusion} & \multicolumn{7}{c}{PXD064980: HLA-II, macrophages}\\
\cmidrule(lr){2-8}\cmidrule(l){9-15}
Missing bonds & 0 & 1 & 2 & 3 & 4--5 & $\ge$6 & All & 0 & 1 & 2 & 3 & 4--5 & $\ge$6 & All\\
Labels & 8{,}844 & 9{,}148 & 4{,}454 & 1{,}780 & 1{,}017 & 208 & 25{,}451 & 906 & 1{,}716 & 2{,}811 & 3{,}040 & 4{,}987 & 9{,}873 & 23{,}333\\
\cmidrule(r){1-1}\cmidrule(lr){2-8}\cmidrule(l){9-15}
InstaNovo & 50.5 & 34.5 & 17.1 & 7.0 & 2.7 & 1.4 & 33.5 & 40.1 & 27.1 & 19.1 & 13.5 & 7.0 & 1.8 & 9.9\\
RNovA & 50.5 & 26.5 & 9.4 & 2.7 & 0.5 & 0.0 & 28.9 & 27.9 & 14.0 & 9.8 & 5.7 & 1.1 & 0.0 & 4.3\\
$\pi$-PrimeNovo & 38.5 & 23.1 & 9.8 & 3.4 & 1.2 & 0.0 & 23.7 & 40.6 & 25.4 & 17.0 & 10.1 & 2.8 & 0.1 & 7.5\\
\Cfive & 31.1 & 13.1 & 6.0 & 2.8 & 1.4 & 0.0 & 16.8 & 22.5 & 14.6 & 7.9 & 5.0 & 1.3 & 0.1 & 3.9\\
NovoCRF, one pass & 85.0 & 74.3 & 59.0 & 51.0 & 43.3 & 21.2 & 72.0 & 76.2 & 68.7 & 56.2 & 46.9 & 33.7 & 13.9 & 34.0\\
\textbf{NovoCRF}, two passes & \textbf{91.6} & \textbf{82.9} & \textbf{68.2} & \textbf{59.7} & \textbf{51.4} & \textbf{28.8} & \textbf{80.0} & \textbf{84.4} & \textbf{78.7} & \textbf{70.0} & \textbf{61.1} & \textbf{45.1} & \textbf{20.2} & \textbf{43.7}\\
NovoCRF, two passes\,+\,CRSE & -- & -- & -- & -- & -- & -- & -- & 85.5 & 81.8 & 73.4 & 65.8 & 51.1 & 23.0 & 47.4\\
\addlinespace[5pt]
 & \multicolumn{7}{c}{PXD062962: mouse liver, phospho} & \multicolumn{7}{c}{PXD062859: human T47D, phospho}\\
\cmidrule(lr){2-8}\cmidrule(l){9-15}
Missing bonds & 0 & 1 & 2 & 3 & 4--5 & $\ge$6 & All & 0 & 1 & 2 & 3 & 4--5 & $\ge$6 & All\\
Labels & 5{,}022 & 6{,}841 & 1{,}958 & 912 & 813 & 1{,}118 & 16{,}664 & 4{,}103 & 4{,}572 & 1{,}424 & 637 & 701 & 526 & 11{,}963\\
\cmidrule(r){1-1}\cmidrule(lr){2-8}\cmidrule(l){9-15}
InstaNovo & 90.3 & 83.8 & 63.6 & 46.6 & 29.8 & 17.5 & 74.2 & 92.1 & 85.9 & 65.4 & 52.0 & 31.4 & 13.9 & 77.4\\
RNovA & 88.9 & 79.8 & 49.4 & 28.8 & 12.2 & 1.5 & 67.6 & 80.6 & 75.9 & 45.1 & 27.3 & 9.0 & 0.6 & 64.0\\
$\pi$-PrimeNovo & -- & -- & -- & -- & -- & -- & -- & -- & -- & -- & -- & -- & -- & --\\
\Cfive & -- & -- & -- & -- & -- & -- & -- & -- & -- & -- & -- & -- & -- & --\\
NovoCRF, one pass & 94.9 & 87.4 & 75.1 & 59.3 & 47.4 & 20.7 & 80.2 & 95.1 & 90.6 & 73.9 & 58.6 & 39.5 & 16.3 & 82.2\\
\textbf{NovoCRF}, two passes & \textbf{96.3} & \textbf{89.3} & \textbf{76.6} & \textbf{63.0} & \textbf{50.1} & \textbf{24.1} & \textbf{82.2} & \textbf{95.8} & \textbf{91.0} & \textbf{74.5} & \textbf{59.3} & \textbf{42.7} & \textbf{16.7} & \textbf{82.9}\\
NovoCRF, two passes\,+\,CRSE & 96.6 & 90.5 & 78.9 & 66.0 & 53.5 & 26.0 & 83.5 & 96.0 & 92.1 & 76.3 & 65.0 & 44.8 & 16.9 & 84.0\\
\bottomrule
\end{tabular}
\end{table*}
\begin{table*}[t]
\centering\footnotesize
\setlength{\tabcolsep}{3.3pt}
\renewcommand{\arraystretch}{0.87}
\caption{Peptide recall (\%) on labels whose peptide is among a tool's own training peptides (S, seen) and on its other
labels (U, unseen), by the number of missing bonds, for each main test set. Peptides are compared as residue sequences
without modifications, with I\,=\,L, and a deamidated N or Q is also read as D or E. The first row of each
dataset gives the share of seen labels. +\,CRSE: NovoCRF in two passes with cross-run spectra enhancement. A dash marks fewer
than 30 labels, a set a tool is not scored on, or a set without companion runs for CRSE. RNovA and $\pi$-PrimeNovo
have the same training peptides. InstaNovo's are those documented for its previous version (\S\ref{sec:seen}). The
table continues on the next page.}
\label{tab:seen}
\begin{tabular}{@{}cl*{7}{rr}@{}}
\toprule
 & & \multicolumn{2}{c}{} & \multicolumn{2}{c}{} & \multicolumn{2}{c}{} & \multicolumn{2}{c}{} &
 \multicolumn{6}{c}{NovoCRF}\\
\cmidrule(l){11-16}
 & Missing & \multicolumn{2}{c}{InstaNovo} & \multicolumn{2}{c}{RNovA} & \multicolumn{2}{c}{$\pi$-PrimeNovo} &
 \multicolumn{2}{c}{CasanovoV5} & \multicolumn{2}{c}{one pass} & \multicolumn{2}{c}{two passes} &
 \multicolumn{2}{c}{+\,CRSE}\\
\cmidrule(lr){3-4}\cmidrule(lr){5-6}\cmidrule(lr){7-8}\cmidrule(lr){9-10}\cmidrule(lr){11-12}\cmidrule(lr){13-14}\cmidrule(l){15-16}
 & bonds & S & U & S & U & S & U & S & U & S & U & S & U & S & U\\
\midrule
\multirow{8}{*}{\rotatebox[origin=c]{90}{\shortstack{PXD076585\\{\scriptsize human, trypsin}}}} & Seen (\%) & \multicolumn{2}{c}{97.9} & \multicolumn{2}{c}{94.9} & \multicolumn{2}{c}{94.9} & \multicolumn{2}{c}{88.3} & \multicolumn{2}{c}{98.6} & \multicolumn{2}{c}{98.6} & \multicolumn{2}{c}{98.6}\\
 & 0 & 94.9 & 83.1 & 93.3 & 86.8 & 90.7 & 79.4 & 84.9 & 80.0 & 94.5 & -- & 95.8 & -- & 96.5 & --\\
 & 1 & 87.0 & 65.7 & 81.0 & 69.8 & 79.3 & 60.8 & 72.4 & 61.2 & 85.1 & 54.3 & 88.5 & 62.9 & 91.8 & 65.7\\
 & 2 & 70.4 & 56.5 & 58.0 & 48.4 & 59.5 & 40.6 & 48.0 & 40.1 & 70.0 & 47.1 & 74.4 & 58.8 & 83.0 & 55.9\\
 & 3 & 51.6 & 27.3 & 35.0 & 21.5 & 39.2 & 22.8 & 30.7 & 22.1 & 49.1 & 25.0 & 55.7 & 36.1 & 67.5 & 50.0\\
 & 4--5 & 35.1 & 33.3 & 15.8 & 11.6 & 21.0 & 18.4 & 16.2 & 14.3 & 30.3 & 19.0 & 35.0 & 16.7 & 50.5 & 26.2\\
 & $\ge$6 & 21.6 & 28.2 & 2.1 & 1.1 & 6.5 & 2.9 & 4.1 & 1.3 & 10.7 & 4.3 & 11.9 & 4.7 & 20.4 & 8.0\\
 & All & 53.3 & 44.6 & 38.8 & 41.9 & 41.2 & 38.8 & 34.6 & 37.9 & 48.7 & 19.5 & 51.9 & 22.2 & 60.2 & 26.2\\
\addlinespace[1pt]\cmidrule(l){2-16}\addlinespace[1pt]
\multirow{8}{*}{\rotatebox[origin=c]{90}{\shortstack{PXD071561\\{\scriptsize squid, Lys-C}}}} & Seen (\%) & \multicolumn{2}{c}{7.9} & \multicolumn{2}{c}{7.3} & \multicolumn{2}{c}{7.3} & \multicolumn{2}{c}{6.9} & \multicolumn{2}{c}{8.9} & \multicolumn{2}{c}{8.9} & \multicolumn{2}{c}{8.9}\\
 & 0 & 96.6 & 94.3 & 96.6 & 93.1 & 95.7 & 91.5 & 88.3 & 85.7 & 97.4 & 95.7 & 98.2 & 96.8 & 98.2 & 97.1\\
 & 1 & 93.0 & 87.7 & 90.5 & 86.0 & 88.9 & 85.9 & 83.1 & 76.3 & 93.1 & 92.5 & 96.4 & 93.9 & 96.4 & 94.5\\
 & 2 & 88.4 & 79.5 & 81.7 & 78.9 & 85.8 & 81.8 & 75.3 & 66.1 & 90.0 & 85.4 & 93.8 & 88.0 & 95.9 & 89.3\\
 & 3 & 75.7 & 52.6 & 60.2 & 48.2 & 76.7 & 57.5 & 57.3 & 35.5 & 86.2 & 63.6 & 87.9 & 69.2 & 92.2 & 73.5\\
 & 4--5 & 50.0 & 27.7 & 27.0 & 19.8 & 53.9 & 34.3 & 21.6 & 17.2 & 67.4 & 43.3 & 68.4 & 50.4 & 69.5 & 54.7\\
 & $\ge$6 & 32.4 & 3.3 & 1.7 & 1.1 & 20.2 & 4.1 & 11.4 & 2.0 & 31.7 & 11.4 & 37.2 & 13.1 & 37.7 & 14.7\\
 & All & 85.2 & 74.9 & 77.6 & 73.0 & 81.7 & 74.7 & 71.9 & 64.9 & 87.7 & 79.8 & 90.1 & 82.0 & 90.8 & 83.1\\
\addlinespace[1pt]\cmidrule(l){2-16}\addlinespace[1pt]
\multirow{8}{*}{\rotatebox[origin=c]{90}{\shortstack{PXD073929\\{\scriptsize frog, Glu-C}}}} & Seen (\%) & \multicolumn{2}{c}{0.5} & \multicolumn{2}{c}{0.3} & \multicolumn{2}{c}{0.3} & \multicolumn{2}{c}{5.8} & \multicolumn{2}{c}{9.8} & \multicolumn{2}{c}{9.8} & \multicolumn{2}{c}{--}\\
 & 0 & -- & 96.5 & -- & 93.8 & -- & 95.7 & -- & 95.4 & 95.3 & 97.7 & 100.0 & 98.6 & -- & --\\
 & 1 & -- & 85.9 & -- & 84.8 & -- & 82.1 & -- & 81.9 & 86.8 & 89.3 & 92.5 & 92.1 & -- & --\\
 & 2 & -- & 65.5 & -- & 56.9 & -- & 60.4 & -- & 61.8 & -- & 68.8 & -- & 75.2 & -- & --\\
 & 3 & -- & 43.2 & -- & 33.2 & -- & 35.3 & -- & 39.9 & -- & 46.4 & -- & 59.2 & -- & --\\
 & 4--5 & -- & 23.5 & -- & 12.8 & -- & 17.9 & -- & 17.9 & 27.3 & 22.6 & 39.4 & 27.9 & -- & --\\
 & $\ge$6 & -- & 4.5 & -- & 1.1 & -- & 3.2 & 3.3 & 2.9 & 2.2 & 6.0 & 8.9 & 10.3 & -- & --\\
 & All & -- & 44.2 & -- & 38.3 & -- & 40.4 & 40.6 & 40.9 & 53.6 & 44.8 & 59.0 & 50.1 & -- & --\\
\addlinespace[1pt]\cmidrule(l){2-16}\addlinespace[1pt]
\multirow{8}{*}{\rotatebox[origin=c]{90}{\shortstack{PXD072115\\{\scriptsize chickpea}}}} & Seen (\%) & \multicolumn{2}{c}{0.6} & \multicolumn{2}{c}{0.4} & \multicolumn{2}{c}{0.4} & \multicolumn{2}{c}{0.5} & \multicolumn{2}{c}{1.1} & \multicolumn{2}{c}{1.1} & \multicolumn{2}{c}{1.1}\\
 & 0 & -- & 40.8 & -- & 32.4 & -- & 42.2 & -- & 34.3 & -- & 84.8 & -- & 89.5 & -- & 89.0\\
 & 1 & 59.4 & 30.5 & -- & 25.1 & -- & 34.5 & -- & 27.3 & 83.3 & 77.0 & 90.0 & 83.5 & 90.0 & 82.8\\
 & 2 & -- & 20.0 & -- & 13.8 & -- & 22.6 & -- & 16.1 & 76.9 & 68.2 & 79.5 & 76.7 & 76.9 & 77.0\\
 & 3 & -- & 9.9 & -- & 6.2 & -- & 13.7 & -- & 7.5 & -- & 56.0 & -- & 64.8 & -- & 64.3\\
 & 4--5 & -- & 3.5 & -- & 1.6 & -- & 3.7 & -- & 2.5 & 36.7 & 39.0 & 38.8 & 46.3 & 38.8 & 48.5\\
 & $\ge$6 & -- & 0.2 & -- & 0.1 & -- & 0.2 & -- & 0.2 & 6.5 & 15.5 & 9.7 & 19.8 & 9.7 & 21.5\\
 & All & 43.5 & 10.4 & 34.7 & 7.5 & 38.9 & 11.8 & 26.5 & 8.6 & 59.5 & 43.5 & 62.3 & 49.9 & 63.2 & 50.9\\
\addlinespace[1pt]\cmidrule(l){2-16}\addlinespace[1pt]
\multirow{8}{*}{\rotatebox[origin=c]{90}{\shortstack{PXD074750\\{\scriptsize \emph{C.\ auris}}}}} & Seen (\%) & \multicolumn{2}{c}{0.3} & \multicolumn{2}{c}{0.2} & \multicolumn{2}{c}{0.2} & \multicolumn{2}{c}{0.3} & \multicolumn{2}{c}{0.6} & \multicolumn{2}{c}{0.6} & \multicolumn{2}{c}{--}\\
 & 0 & -- & 76.4 & -- & 77.0 & -- & 74.9 & -- & 65.3 & 96.8 & 89.9 & 96.8 & 93.3 & -- & --\\
 & 1 & -- & 59.4 & -- & 58.3 & -- & 61.2 & -- & 47.9 & -- & 74.5 & -- & 81.9 & -- & --\\
 & 2 & -- & 41.2 & -- & 39.0 & -- & 45.6 & -- & 31.5 & -- & 57.5 & -- & 68.3 & -- & --\\
 & 3 & -- & 20.4 & -- & 18.1 & -- & 24.1 & -- & 13.4 & -- & 36.4 & -- & 51.1 & -- & --\\
 & 4--5 & -- & 8.1 & -- & 5.8 & -- & 9.7 & -- & 5.1 & -- & 24.6 & -- & 35.8 & -- & --\\
 & $\ge$6 & -- & 0.9 & -- & 0.4 & -- & 0.8 & -- & 0.4 & -- & 9.3 & -- & 14.3 & -- & --\\
 & All & 56.2 & 30.5 & 52.5 & 29.2 & 52.5 & 32.1 & 45.5 & 24.0 & 64.0 & 44.2 & 71.2 & 52.6 & -- & --\\
\addlinespace[1pt]\cmidrule(l){2-16}\addlinespace[1pt]
\multirow{8}{*}{\rotatebox[origin=c]{90}{\shortstack{PXD055085\\{\scriptsize rabbit antibody}}}} & Seen (\%) & \multicolumn{2}{c}{9.5} & \multicolumn{2}{c}{8.4} & \multicolumn{2}{c}{8.4} & \multicolumn{2}{c}{8.8} & \multicolumn{2}{c}{10.4} & \multicolumn{2}{c}{10.4} & \multicolumn{2}{c}{--}\\
 & 0 & 80.0 & 81.8 & 77.4 & 78.7 & 78.4 & 78.4 & 77.9 & 68.1 & 80.0 & 85.2 & 80.7 & 86.7 & -- & --\\
 & 1 & 83.5 & 82.0 & 77.9 & 73.8 & 81.7 & 78.1 & 79.6 & 66.8 & 80.4 & 84.2 & 82.1 & 87.1 & -- & --\\
 & 2 & 81.1 & 69.8 & 65.6 & 58.5 & 79.6 & 68.6 & 72.3 & 53.4 & 73.3 & 72.7 & 77.3 & 78.6 & -- & --\\
 & 3 & 76.8 & 63.7 & 47.6 & 47.6 & 72.6 & 62.6 & 62.4 & 49.2 & 62.0 & 64.5 & 62.0 & 72.0 & -- & --\\
 & 4--5 & 69.4 & 44.4 & 30.9 & 20.6 & 64.9 & 40.8 & 46.3 & 28.4 & 43.1 & 48.4 & 45.7 & 57.6 & -- & --\\
 & $\ge$6 & 56.1 & 16.5 & 4.1 & 3.5 & 44.6 & 13.9 & 30.8 & 6.8 & 11.3 & 19.6 & 17.0 & 24.2 & -- & --\\
 & All & 79.2 & 66.3 & 68.0 & 56.1 & 76.6 & 63.5 & 71.8 & 52.1 & 72.5 & 68.9 & 74.3 & 73.3 & -- & --\\
\bottomrule
\end{tabular}
\end{table*}

\begin{table*}[t]
\ContinuedFloat
\centering\footnotesize
\setlength{\tabcolsep}{3.3pt}
\renewcommand{\arraystretch}{0.87}
\caption{Continued: the six other main test sets.}
\begin{tabular}{@{}cl*{7}{rr}@{}}
\toprule
 & & \multicolumn{2}{c}{} & \multicolumn{2}{c}{} & \multicolumn{2}{c}{} & \multicolumn{2}{c}{} &
 \multicolumn{6}{c}{NovoCRF}\\
\cmidrule(l){11-16}
 & Missing & \multicolumn{2}{c}{InstaNovo} & \multicolumn{2}{c}{RNovA} & \multicolumn{2}{c}{$\pi$-PrimeNovo} &
 \multicolumn{2}{c}{CasanovoV5} & \multicolumn{2}{c}{one pass} & \multicolumn{2}{c}{two passes} &
 \multicolumn{2}{c}{+\,CRSE}\\
\cmidrule(lr){3-4}\cmidrule(lr){5-6}\cmidrule(lr){7-8}\cmidrule(lr){9-10}\cmidrule(lr){11-12}\cmidrule(lr){13-14}\cmidrule(l){15-16}
 & bonds & S & U & S & U & S & U & S & U & S & U & S & U & S & U\\
\midrule
\multirow{8}{*}{\rotatebox[origin=c]{90}{\shortstack{PXD071022\\{\scriptsize HLA-I, colorectal}}}} & Seen (\%) & \multicolumn{2}{c}{59.1} & \multicolumn{2}{c}{9.8} & \multicolumn{2}{c}{9.8} & \multicolumn{2}{c}{6.6} & \multicolumn{2}{c}{82.3} & \multicolumn{2}{c}{82.3} & \multicolumn{2}{c}{82.3}\\
 & 0 & 95.1 & 92.9 & 95.6 & 89.0 & 94.9 & 85.6 & 93.8 & 81.8 & 97.6 & 93.5 & 98.0 & 94.8 & 98.2 & 93.5\\
 & 1 & 86.9 & 82.0 & 85.5 & 74.7 & 84.8 & 71.2 & 79.7 & 67.4 & 93.4 & 84.9 & 94.3 & 87.5 & 95.0 & 88.3\\
 & 2 & 68.7 & 64.4 & 55.5 & 49.4 & 63.1 & 50.1 & 57.4 & 45.3 & 86.4 & 72.6 & 88.4 & 75.6 & 89.7 & 76.1\\
 & 3 & 47.6 & 41.9 & 29.7 & 23.6 & 40.7 & 25.3 & 32.9 & 23.5 & 74.0 & 57.0 & 76.9 & 61.7 & 79.3 & 61.5\\
 & 4--5 & 20.6 & 18.0 & 13.5 & 7.3 & 19.7 & 7.6 & 20.6 & 7.8 & 57.3 & 42.9 & 59.9 & 47.8 & 62.1 & 49.0\\
 & $\ge$6 & 8.4 & 8.1 & 2.9 & 1.3 & 1.9 & 2.6 & 7.2 & 3.3 & 41.7 & 21.5 & 44.5 & 24.9 & 48.0 & 23.4\\
 & All & 68.5 & 63.1 & 66.3 & 53.7 & 69.0 & 52.4 & 65.9 & 49.6 & 83.9 & 69.6 & 85.5 & 72.8 & 86.7 & 73.0\\
\addlinespace[1pt]\cmidrule(l){2-16}\addlinespace[1pt]
\multirow{8}{*}{\rotatebox[origin=c]{90}{\shortstack{PXD082505\\{\scriptsize HLA-I, B*15:02}}}} & Seen (\%) & \multicolumn{2}{c}{42.1} & \multicolumn{2}{c}{1.5} & \multicolumn{2}{c}{1.5} & \multicolumn{2}{c}{1.7} & \multicolumn{2}{c}{66.6} & \multicolumn{2}{c}{66.6} & \multicolumn{2}{c}{66.6}\\
 & 0 & 99.1 & 96.9 & -- & 96.4 & -- & 95.3 & -- & 93.5 & 99.7 & 100.0 & 100.0 & 100.0 & 100.0 & 100.0\\
 & 1 & 96.7 & 93.4 & -- & 89.2 & -- & 89.6 & -- & 83.2 & 97.5 & 96.8 & 97.8 & 97.1 & 98.9 & 97.1\\
 & 2 & 86.8 & 77.9 & -- & 70.6 & -- & 76.6 & -- & 65.0 & 95.2 & 90.5 & 95.5 & 90.8 & 97.9 & 91.7\\
 & 3 & 69.8 & 56.1 & -- & 46.1 & -- & 53.4 & -- & 44.1 & 85.5 & 80.3 & 88.0 & 85.6 & 93.1 & 84.8\\
 & 4--5 & 32.6 & 28.3 & -- & 15.7 & -- & 21.9 & -- & 13.7 & 78.1 & 64.5 & 80.8 & 69.3 & 85.6 & 73.2\\
 & $\ge$6 & 2.2 & 5.8 & -- & 2.2 & -- & 3.4 & -- & 0.8 & 66.9 & 31.3 & 69.6 & 34.5 & 70.4 & 37.8\\
 & All & 68.2 & 58.5 & 52.2 & 52.9 & 55.1 & 56.8 & 61.7 & 49.2 & 88.5 & 74.9 & 89.9 & 77.5 & 92.5 & 79.0\\
\addlinespace[1pt]\cmidrule(l){2-16}\addlinespace[1pt]
\multirow{8}{*}{\rotatebox[origin=c]{90}{\shortstack{PXD054498\\{\scriptsize HLA-I, effusion}}}} & Seen (\%) & \multicolumn{2}{c}{43.5} & \multicolumn{2}{c}{3.3} & \multicolumn{2}{c}{3.3} & \multicolumn{2}{c}{2.8} & \multicolumn{2}{c}{63.6} & \multicolumn{2}{c}{63.6} & \multicolumn{2}{c}{--}\\
 & 0 & 53.6 & 47.4 & 53.9 & 50.4 & 47.6 & 38.1 & 41.8 & 30.8 & 87.3 & 79.2 & 93.7 & 86.0 & -- & --\\
 & 1 & 36.6 & 33.0 & 29.3 & 26.4 & 33.2 & 22.9 & 22.0 & 12.9 & 77.7 & 69.0 & 87.1 & 76.3 & -- & --\\
 & 2 & 20.2 & 15.2 & 19.7 & 9.1 & 21.3 & 9.5 & 7.2 & 6.0 & 65.0 & 50.6 & 76.1 & 57.3 & -- & --\\
 & 3 & 8.8 & 6.0 & 7.8 & 2.5 & 19.6 & 2.9 & 16.1 & 2.3 & 60.1 & 39.5 & 70.1 & 46.5 & -- & --\\
 & 4--5 & 4.0 & 1.8 & 0.0 & 0.5 & 4.3 & 1.0 & 8.2 & 1.0 & 50.1 & 34.2 & 59.4 & 40.9 & -- & --\\
 & $\ge$6 & 3.2 & 0.0 & 0.0 & 0.0 & 0.0 & 0.0 & 0.0 & 0.0 & 26.4 & 12.7 & 34.9 & 19.0 & -- & --\\
 & All & 37.7 & 30.4 & 33.5 & 28.8 & 33.2 & 23.4 & 24.4 & 16.6 & 76.9 & 63.4 & 85.5 & 70.4 & -- & --\\
\addlinespace[1pt]\cmidrule(l){2-16}\addlinespace[1pt]
\multirow{8}{*}{\rotatebox[origin=c]{90}{\shortstack{PXD064980\\{\scriptsize HLA-II, macrophage}}}} & Seen (\%) & \multicolumn{2}{c}{25.5} & \multicolumn{2}{c}{11.6} & \multicolumn{2}{c}{11.6} & \multicolumn{2}{c}{10.0} & \multicolumn{2}{c}{33.9} & \multicolumn{2}{c}{33.9} & \multicolumn{2}{c}{33.9}\\
 & 0 & 55.5 & 28.0 & 49.3 & 21.8 & 61.7 & 34.6 & 33.7 & 20.1 & 84.0 & 65.8 & 90.5 & 76.5 & 89.3 & 80.6\\
 & 1 & 41.1 & 18.4 & 29.7 & 11.3 & 47.5 & 21.5 & 32.1 & 11.9 & 76.6 & 60.5 & 84.8 & 72.4 & 86.7 & 76.7\\
 & 2 & 34.3 & 11.7 & 24.9 & 7.7 & 33.1 & 14.8 & 18.1 & 6.6 & 67.6 & 47.1 & 78.4 & 63.2 & 81.2 & 67.0\\
 & 3 & 23.2 & 9.2 & 11.3 & 4.9 & 20.4 & 8.5 & 11.0 & 4.3 & 56.7 & 39.9 & 68.6 & 55.7 & 74.3 & 59.6\\
 & 4--5 & 15.5 & 4.0 & 2.6 & 0.9 & 6.2 & 2.4 & 2.3 & 1.2 & 47.4 & 26.5 & 60.5 & 36.9 & 65.5 & 43.5\\
 & $\ge$6 & 6.2 & 0.9 & 0.1 & 0.0 & 0.4 & 0.1 & 0.4 & 0.0 & 28.6 & 9.6 & 38.1 & 15.0 & 41.9 & 17.4\\
 & All & 22.4 & 5.6 & 12.0 & 3.3 & 17.9 & 6.1 & 10.1 & 3.2 & 52.4 & 24.6 & 62.9 & 33.8 & 66.6 & 37.6\\
\addlinespace[1pt]\cmidrule(l){2-16}\addlinespace[1pt]
\multirow{8}{*}{\rotatebox[origin=c]{90}{\shortstack{PXD062962\\{\scriptsize mouse, phospho}}}} & Seen (\%) & \multicolumn{2}{c}{63.0} & \multicolumn{2}{c}{28.6} & \multicolumn{2}{c}{--} & \multicolumn{2}{c}{--} & \multicolumn{2}{c}{65.7} & \multicolumn{2}{c}{65.7} & \multicolumn{2}{c}{65.7}\\
 & 0 & 94.9 & 79.2 & 92.9 & 86.7 & -- & -- & -- & -- & 96.1 & 91.6 & 97.0 & 94.5 & 97.4 & 94.5\\
 & 1 & 91.3 & 68.0 & 86.3 & 77.0 & -- & -- & -- & -- & 89.8 & 81.8 & 91.4 & 84.5 & 92.3 & 86.1\\
 & 2 & 77.5 & 47.8 & 55.2 & 47.8 & -- & -- & -- & -- & 78.9 & 70.1 & 80.4 & 71.6 & 82.4 & 74.3\\
 & 3 & 60.3 & 34.8 & 32.4 & 28.0 & -- & -- & -- & -- & 63.3 & 55.6 & 66.4 & 59.9 & 69.2 & 63.1\\
 & 4--5 & 47.4 & 18.2 & 12.9 & 12.0 & -- & -- & -- & -- & 54.2 & 42.3 & 55.9 & 45.7 & 60.6 & 48.3\\
 & $\ge$6 & 31.5 & 4.5 & 4.3 & 0.9 & -- & -- & -- & -- & 25.5 & 15.2 & 28.9 & 18.6 & 31.4 & 20.0\\
 & All & 85.5 & 55.1 & 78.6 & 63.2 & -- & -- & -- & -- & 85.2 & 70.8 & 86.6 & 73.7 & 87.8 & 75.2\\
\addlinespace[1pt]\cmidrule(l){2-16}\addlinespace[1pt]
\multirow{8}{*}{\rotatebox[origin=c]{90}{\shortstack{PXD062859\\{\scriptsize human, phospho}}}} & Seen (\%) & \multicolumn{2}{c}{89.5} & \multicolumn{2}{c}{75.8} & \multicolumn{2}{c}{--} & \multicolumn{2}{c}{--} & \multicolumn{2}{c}{91.3} & \multicolumn{2}{c}{91.3} & \multicolumn{2}{c}{91.3}\\
 & 0 & 92.9 & 80.4 & 80.8 & 79.7 & -- & -- & -- & -- & 95.3 & 90.1 & 95.9 & 94.2 & 96.0 & 94.8\\
 & 1 & 87.8 & 56.8 & 79.1 & 61.3 & -- & -- & -- & -- & 91.3 & 76.2 & 91.6 & 78.0 & 92.4 & 86.9\\
 & 2 & 69.8 & 33.9 & 49.6 & 34.9 & -- & -- & -- & -- & 76.1 & 55.3 & 75.9 & 62.5 & 77.7 & 65.1\\
 & 3 & 57.8 & 26.3 & 30.5 & 22.4 & -- & -- & -- & -- & 62.0 & 42.5 & 63.0 & 42.5 & 68.3 & 49.6\\
 & 4--5 & 37.6 & 13.7 & 10.2 & 7.8 & -- & -- & -- & -- & 42.3 & 31.2 & 46.1 & 32.4 & 47.6 & 36.4\\
 & $\ge$6 & 24.0 & 0.9 & 1.4 & 0.0 & -- & -- & -- & -- & 20.4 & 10.1 & 22.0 & 8.7 & 23.0 & 7.7\\
 & All & 81.9 & 39.5 & 69.9 & 45.3 & -- & -- & -- & -- & 85.1 & 50.8 & 85.7 & 52.9 & 86.6 & 56.4\\
\bottomrule
\end{tabular}
\end{table*}
\begin{table*}[t]
\centering\footnotesize
\setlength{\tabcolsep}{2.9pt}
\caption{Peptide recall (\%) on the labels whose peptide is in no tool's training peptides, so that every method is
scored on the same labels and none has seen the answer. Top: all such labels. Bottom: those with at most one missing
bond. CRSE: with two companion runs, where a set has them. Best per column among the methods that read the test
run alone in bold.}
\label{tab:novel}
\begin{tabular}{@{}l*{12}{r}@{}}
\toprule
 & 076585 & 071561 & 073929 & 072115 & 074750 & 055085 & 071022 & 082505 & 054498 & 064980 & 062962 & 062859\\
Method & human & squid & frog & chickpea & \emph{C.\ auris} & antibody & HLA-I & HLA-I & HLA-I & HLA-II & phospho & phospho\\
\midrule
\multicolumn{13}{@{}l}{\emph{All such labels}}\\
Labels & 170 & 22{,}473 & 2{,}040 & 19{,}025 & 22{,}349 & 20{,}442 & 3{,}835 & 1{,}539 & 9{,}248 & 14{,}957 & 5{,}565 & 631\\
InstaNovo & 18.8 & 74.7 & 42.8 & 10.3 & 30.4 & 66.0 & 56.5 & 53.2 & 28.7 & 4.1 & 53.6 & 21.4\\
RNovA & 18.8 & 72.7 & 37.2 & 7.4 & 29.2 & 55.7 & 47.2 & 47.4 & 25.2 & 2.3 & 51.2 & 20.3\\
$\pi$-PrimeNovo & 14.1 & 74.5 & 39.9 & 11.8 & 32.0 & 63.3 & 44.4 & 49.5 & 18.6 & 4.3 & -- & --\\
\Cfive & 15.9 & 64.6 & 39.8 & 8.5 & 24.0 & 51.7 & 42.4 & 43.8 & 9.7 & 1.9 & -- & --\\
NovoCRF, one pass & 21.2 & 79.8 & 44.9 & 43.6 & 44.2 & 69.1 & 69.6 & 74.8 & 63.5 & 24.8 & 70.7 & 41.0\\
\textbf{NovoCRF}, two passes & \textbf{24.1} & \textbf{82.0} & \textbf{50.2} & \textbf{50.0} & \textbf{52.6} & \textbf{73.3} & \textbf{72.9} & \textbf{77.5} & \textbf{70.4} & \textbf{34.0} & \textbf{73.6} & \textbf{41.7}\\
NovoCRF, two passes\,+\,CRSE & 25.3 & 83.1 & -- & 50.9 & -- & -- & 73.1 & 78.9 & -- & 37.7 & 75.1 & 44.4\\
\addlinespace[3pt]
\multicolumn{13}{@{}l}{\emph{With at most one missing bond}}\\
Labels & 36 & 13{,}536 & 505 & 3{,}001 & 6{,}505 & 10{,}585 & 1{,}651 & 354 & 6{,}082 & 1{,}195 & 3{,}288 & 155\\
InstaNovo & 69.4 & 91.7 & 91.1 & 33.3 & 66.1 & 81.8 & 82.0 & 92.1 & 38.3 & 17.0 & 71.2 & 46.5\\
RNovA & 66.7 & 90.3 & 88.3 & 27.1 & 65.8 & 76.0 & 76.5 & 89.3 & 35.0 & 11.7 & 71.5 & 49.7\\
$\pi$-PrimeNovo & 55.6 & 89.4 & 88.7 & 36.7 & 66.7 & 78.0 & 69.2 & 89.8 & 25.5 & 21.7 & -- & --\\
\Cfive & 55.6 & 81.9 & 87.9 & 29.2 & 54.9 & 67.1 & 67.1 & 85.9 & 13.3 & 10.3 & -- & --\\
NovoCRF, one pass & \textbf{72.2} & 94.5 & 92.9 & 79.3 & 80.7 & 84.7 & 88.0 & 97.5 & 73.2 & 62.0 & 85.6 & 73.5\\
\textbf{NovoCRF}, two passes & \textbf{72.2} & \textbf{95.6} & \textbf{94.9} & \textbf{85.2} & \textbf{86.5} & \textbf{86.9} & \textbf{90.1} & \textbf{97.7} & \textbf{80.3} & \textbf{73.4} & \textbf{88.4} & \textbf{74.2}\\
NovoCRF, two passes\,+\,CRSE & 72.2 & 96.0 & -- & 84.7 & -- & -- & 90.1 & 97.7 & -- & 77.7 & 89.4 & 82.6\\
\bottomrule
\end{tabular}
\end{table*}

\noindent\textbf{Baselines.} We compare with four transformer models, each with its released weights:
InstaNovo~\cite{eloff2025} 1.2.2 (model v1.2.0, five beams, without the diffusion refinement), RNovA~\cite{mao2026}
(its sequencing model, SeqFiller, on every spectrum), $\pi$-PrimeNovo~\cite{zhang2025} and Casanovo
5.0.0~\cite{straub2025} with five beams (\Cfive). All four were run with the containers and recipes of the
denovo\_benchmarks project\footnote{\url{https://github.com/bittremieuxlab/denovo_benchmarks}, with its Apptainer
recipes translated to Docker.}, which provides a container and a run script for each tool. Each tool
received the modification setting closest to what a set's labels carry: residues outside it were switched off, and on
sets without alkylation the tool's Cys residue got the unmodified mass. The precursor tolerance was that of the database search
that produced the labels. Beyond these settings we changed one thing: RNovA's recipe drops sequences below a fixed score, a cutoff that
RNovA's own code does not have, so we kept every sequence. \Cfive\ does not sequence precursors of charge 5 or more, or spectra with
fewer than 20 peaks, which leaves 0--14\% of a set's labels without an answer.

The other tools read MGF files written by PrecursorMassFinder, so every tool gets the
corrected precursors (Appendix~\ref{app:precursor}). NovoCRF reads the raw files directly, because its intensity
predictor uses the collision energy that the raw files record and MGF files do not carry. It sees the same spectra, and
its precursor masses come from the same PrecursorMassFinder, which is built into the engine (\S\ref{sec:rdr}). NovoCRF results are reported in three configurations: one pass (Fig.~\ref{fig:pipeline}a,
its answer after the first pass and before any on-the-fly adaptation), its default two passes, and two passes
with CRSE on the sets that have companion runs. All timings are from one desktop computer (Intel i9-13900K with 32 threads, one RTX
3090), one tool at a time with no other job on the GPU. The four baselines ran on the GPU, and NovoCRF ran with the GPU,
where it uses the CPU threads as well, or on the CPU alone. A baseline's time is the inference time that its log reports, and NovoCRF's
is its whole run, reading the raw file included.

\noindent\textbf{Metrics.} A prediction is correct if it matches the label residue by residue, Casanovo style: each
residue within 0.1\Da\ and the running prefix or suffix mass within 0.5\Da, with I\,=\,L. Peptide recall is correct
predictions over labeled spectra, so a labeled spectrum without an answer counts as a miss. The precision--recall (PR)
curve ranks a tool's answers by its own score, and PR AUC is the area under it.

\subsection{Peptide recall and precision--recall curves}\label{sec:recall}
Table~\ref{tab:main} and Fig.~\ref{fig:pr} give the results. Among the methods that read the test run alone,
NovoCRF in its default two passes has the highest peptide recall and the highest PR AUC on eleven of the twelve
datasets. Its lead over the best of the four other tools ranges from 5.5 points on the human phospho-enriched set to
46.5 on the pleural-effusion HLA-I set, and it exceeds 20 points on five datasets: chickpea, \emph{C.\ auris} and three
of the four HLA sets. The exception is the human tryptic digest, where InstaNovo finds 53.1\% of the labels and
NovoCRF 51.5\%. NovoCRF is ahead of RNovA, $\pi$-PrimeNovo and \Cfive\ on every dataset. Its first pass alone is
ahead of all four tools on the same eleven datasets.

The second pass helps on every dataset, by 0.7 points (human phospho-enriched) to 9.7 (HLA-II). Over all twelve sets
it gains 14{,}887 labels and loses 3{,}988. CRSE helps on each of the eight sets with companion runs, by 0.9 points
(chickpea) to 8.2 (human trypsin), and over the eight sets it gains 7{,}345 labels and loses 2{,}748. On the human
tryptic digest, the two replicate runs lift NovoCRF from 51.5\% to 59.7\%, above every tool that reads the test
run alone. The two sets whose companions are not replicates in the strict sense gain 1.0 and 2.3 points. The next
sections look at where these differences come from.

\subsection{Accuracy by fragment evidence}\label{sec:evidence}
We classify each label by its fragment evidence in the vendor-centroided peak list. A peptide bond is \emph{observed}
if a peak lies within the fragment tolerance of its b or y ion, and \emph{missing} otherwise. The tolerance is that of
the database search that produced the labels (0.02\Da, or 0.05\Da\ for timsTOF).
Fragments of charge 2 count for precursors of charge $\ge3$. An \emph{unsupported stretch} is a run of consecutive
residues with no observed bond between them, so the spectrum constrains only its summed mass. Of the 227{,}039 labels,
20.6\% have a complete ladder, 59.3\% miss at most two bonds, and 37.7\% contain an unsupported stretch of three or
more residues.

Table~\ref{tab:missing} splits peptide recall by the number of missing bonds, for each dataset. With up to three
missing bonds, NovoCRF in two passes is ahead of the four other tools on every dataset. Pooled over the datasets on which both tools are scored, among the labels with at most one missing bond that only one of two tools gets
right, NovoCRF has 17{,}665 against InstaNovo's 2{,}419, 21{,}833 against RNovA's 1{,}906, 20{,}202 against
$\pi$-PrimeNovo's 1{,}436 and 26{,}097 against \Cfive's 1{,}002.

It turns out that large transformer models with peak encoding and end-to-end training have not discovered a
fragmentation pattern that engineered features miss. On labeled spectra with at most one missing bond, NovoCRF, with
1.80M parameters and features that encode known fragmentation chemistry, outperforms all four large models on every
dataset.

A spectrum that lacks the b and y ions of six or more bonds does not give the order of many of its residues, so a
correct answer needs another source. Large transformer models have the capacity to learn and remember the sequence
patterns of the training peptides that they have seen, and they generate sequences that resemble the reference
proteome when fragment support is missing. For NovoCRF, we did try adding peptide language model components during
development. But we found that adding a sequence prior boosts performance on seen peptides at the cost of recall on
peptides unseen in training, which we consider unacceptable for a \textit{de novo} tool. Because of that, we chose
the sibling spectra enhancement method (\S\ref{sec:sibling}) to recover the sequences of poorly fragmented spectra.

With six or more missing bonds, NovoCRF still finds 10.2\% to 52.5\% of a dataset's labels, and InstaNovo, the
strongest of the four other tools on such labels, 0.2\% to 21.8\%. On chickpea, where almost no label is a training
peptide of any tool, NovoCRF finds 19.7\% of these labels and no other tool more than 0.2\%. Over all twelve
datasets, among the labels with six or more missing bonds that only one of the two gets right, NovoCRF has 6{,}428
and InstaNovo 2{,}399. The one dataset where a transformer model leads on such labels is the human tryptic digest:
with six or more missing bonds, InstaNovo finds 21.8\% of its labels, NovoCRF 11.6\%, and NovoCRF with CRSE 20.1\%.
The next section shows how much the recall of the transformer models on poorly fragmented spectra depends on having
seen the peptide in training.

\subsection{Peptides seen and unseen in training}\label{sec:seen}
A label is \emph{seen} by a tool if its peptide is one of the tool's training peptides. We compare residue sequences
without modifications, with I\,=\,L. A deamidated N or Q is a D or E, so a label with one also counts as seen if the
sequence with D or E in its place is a training peptide. We take the training peptides of each tool from its own
data:
\begin{itemize}
\item \textbf{NovoCRF:} the 10M training set, and the PROSPECT PTMs splits that the two predictors were trained or
  selected on (3.18M such sequences).
\item \textbf{InstaNovo:} the training data of model v1.2.0 are not documented. We take the sources documented for
  v1.1.0: the published training splits of ProteomeTools, the CompOmics projects and InstaNovo-P, and all of
  MassIVE-KB, whose split was not published (1.71M).
\item \textbf{RNovA and $\pi$-PrimeNovo:} MassIVE-KB, all splits, because their splits are not known (1.08M).
\item \textbf{CasanovoV5:} the 2.9M training spectra of Casanovo v4.2~\cite{melendez2024}, on which v5.0 was
  trained as well\footnote{\url{https://casanovo.readthedocs.io/en/latest/faq.html}, data at
  \url{https://zenodo.org/records/12587317}.}: 2.0M tryptic spectra sampled from MassIVE-KB and the 0.9M training
  spectra of a multi-enzyme set (0.89M sequences).
\end{itemize}
Table~\ref{tab:seen} gives the recall on seen and unseen labels by missing bonds. The share of seen labels differs
widely: 88--99\% on human trypsin and about 10\% or less on squid, frog, chickpea, \emph{C.\ auris} and the antibody.
The four HLA sets are 34--82\% seen for NovoCRF, whose training data hold many HLA peptides, and 26--59\% for
InstaNovo. RNovA, $\pi$-PrimeNovo and \Cfive\ have seen at most 12\% of the labels of these four sets.

All five tools have a higher recall on seen peptides than on unseen ones, which is inevitable for a model that
learns from data. Among the cells of Table~\ref{tab:seen} with at least 30 labels on both sides, each a dataset and a
bin of missing bonds, the seen labels have the higher recall in 52 of 55 for InstaNovo, 44 of 48 for RNovA, 33 of 36
for $\pi$-PrimeNovo, 36 of 37 for \Cfive\ and 52 of 62 for NovoCRF with two passes.

On poorly fragmented spectra, the large transformer models fill in a sequence from their sequence prior. The problem
is that such a prior is right mostly for peptides that the model has seen in training. With six or more missing
bonds, InstaNovo recalls 32.4\% of its seen squid labels and 3.3\% of its unseen ones, 56.1\% against 16.5\% on the
antibody, and 31.5\% against 4.5\% on mouse liver. $\pi$-PrimeNovo has 20.2\% against 4.1\% on squid and 44.6\%
against 13.9\% on the antibody, and \Cfive\ 30.8\% against 6.8\% on the antibody.

Each tool's unseen labels are a different set of labels. Table~\ref{tab:novel} therefore scores all methods on the
labels that are in no tool's training peptides. NovoCRF is ahead of the four other tools on all twelve datasets, by
5.3 points (human trypsin, on 170 such labels) to 41.7 (pleural effusion), and also on every dataset among the labels
with at most one missing bond. NovoCRF's improvement in recall is therefore genuine: it holds on every dataset for
peptides that no tool has seen in training.

NovoCRF recovers poorly fragmented spectra from the sibling spectra of the same peptide instead of a sequence prior
(\S\ref{sec:sibling}), and this also works for peptides unseen in training. On the spectra with peptides that no tool has seen,
NovoCRF correctly sequences 17.0\% of those with six or more missing bonds and 43.9\% of those with four or five,
where the best of the four other tools reaches 2.5\% and 12.9\%. With sibling spectra enhancement switched off
(Appendix~\ref{app:nosib}, Table~\ref{tab:nosib}), NovoCRF reaches 1.0\% and 12.4\%, so this recall comes from
the sibling spectra. Table~\ref{tab:nosib} also shows that the enhancement raises the recall on both seen and unseen
peptides. Over all labels of a dataset, the recall of NovoCRF's two passes rises on each of the twelve datasets, by
3.1 to 37.0 points on its seen labels and by 4.7 to 25.7 points on its unseen labels. Without the enhancement, on the spectra with peptides that
no tool has seen, NovoCRF is still ahead of the four other tools on eleven of the twelve datasets, and on
all twelve among those with at most one missing bond.

We firmly believe that the primary goal of a \textit{de novo} tool is to find novel peptides with good spectrum
support that are not in the current reference proteome, not to rediscover peptides that database-search engines can
already identify. Recall on unseen peptides therefore matters far more than recall on seen ones. We advocate that
future \textit{de novo} tools report their metrics split by seen and unseen peptides and by missing bonds, so that
users and reviewers know where an improvement comes from.

\begin{table*}[t]
\centering\footnotesize
\setlength{\tabcolsep}{3.6pt}
\caption{Modifications never seen in CRF training, given to NovoCRF as user modifications. (a) Peptide recall (\%) on the
hydroxyproline-rich sets, with residues matched within 0.01\Da, so that a hydroxyproline (Hyp) site must be called as
P+15.995 and not as L. RNovA was run with the same modified residues added to its residue table. $^a$Labels from the authors' MaxQuant
search. (b) RNovA's spike-in of synthetic peptides with
four non-natural residues: recall (\%) of the labeled spectra of each peptide group, and the number of labels that
NovoCRF in two passes gets right and RNovA does not, and the reverse. RNovA: the sequences its authors deposited.
Labels from PEAKS. (c) The same runs with labels made without PEAKS: a spectrum is labeled with a peptide of its run's
group if its precursor lies within 10\,ppm of the peptide or of one isotope peak below or above it (37 of the
1{,}547 labels), at least six of the peptide's b and y ions (also 2+ for
precursors of charge 3 or more) match peaks within 0.02\Da, covering at least 30\% of its singly charged b and y
ions, at least three of the matched ions contain a non-natural residue, and every other peptide of the group matches
at least two ions fewer. In bold: the best of RNovA and NovoCRF's one and two passes, in (a) within each group of
labels.}
\label{tab:ptm}
\begin{tabular}{@{}lrrrrrrrrrrrr@{}}
\multicolumn{13}{@{}l}{(a) Hydroxyproline}\\
\toprule
 & \multicolumn{4}{c}{All labels} & \multicolumn{4}{c}{Hyp labels} & \multicolumn{4}{c}{Hyp labels, $\le$2 missing bonds}\\
\cmidrule(lr){2-5}\cmidrule(lr){6-9}\cmidrule(l){10-13}
 & & & \multicolumn{2}{c}{NovoCRF} & & & \multicolumn{2}{c}{NovoCRF} & & & \multicolumn{2}{c}{NovoCRF}\\
\cmidrule(lr){4-5}\cmidrule(lr){8-9}\cmidrule(l){12-13}
 & & & one & two & & & one & two & & & one & two\\
Dataset & Labels & RNovA & pass & passes & Labels & RNovA & pass & passes & Labels & RNovA & pass & passes\\
\midrule
PXD073112, enamel & 7{,}644 & 41.9 & 72.4 & \textbf{76.2} & 1{,}064 & 13.9 & 33.3 & \textbf{36.5} & 486 & 24.5 & 43.6 & \textbf{47.1}\\
PXD073112, dentine & 2{,}782 & 31.1 & 43.7 & \textbf{48.1} & 1{,}627 & 16.2 & 26.9 & \textbf{31.3} & 325 & 61.2 & 70.2 & \textbf{73.2}\\
PXD075875, tortoise bone & 9{,}571 & 24.4 & 36.8 & \textbf{42.8} & 8{,}409 & 20.7 & 34.0 & \textbf{39.9} & 3{,}523 & 44.8 & 63.9 & \textbf{71.0}\\
PXD065046, mouse collagen & 1{,}802 & 27.9 & 43.1 & \textbf{43.3} & 1{,}624 & 24.8 & 40.8 & \textbf{41.1} & 340 & 62.9 & \textbf{77.9} & 77.6\\
\addlinespace[2pt]
PXD065297, lion bone$^a$ & 3{,}674 & 49.6 & 64.7 & \textbf{75.9} & 2{,}766 & 43.8 & 59.8 & \textbf{72.6} & 2{,}254 & 50.7 & 65.9 & \textbf{77.4}\\
\bottomrule
\end{tabular}\par\vspace{6pt}
\begin{tabular}{@{}lrrrrrr@{}}
\multicolumn{7}{@{}l}{(b) RNovA's spike-in, PXD076296}\\
\toprule
 & & & \multicolumn{2}{c}{NovoCRF} &
\multicolumn{2}{c}{\begin{tabular}[b]{@{}c@{}}Labels that only\\one tool gets right\end{tabular}}\\
\cmidrule(lr){4-5}\cmidrule(l){6-7}
Peptide & & & one & two & NovoCRF, & \\
group & Labels & RNovA & pass & passes & two passes & RNovA\\
\midrule
Group 1 & 508 & 69.7 & 83.5 & \textbf{85.4} & 80 & 0\\
Group 2 & 524 & 63.5 & 78.2 & \textbf{78.4} & 79 & 1\\
Group 3 & 661 & 61.1 & \textbf{66.3} & 63.5 & 42 & 26\\
\addlinespace[2pt]
All three & 1{,}693 & 64.4 & \textbf{75.1} & 74.7 & 201 & 27\\
\bottomrule
\end{tabular}\hspace{18pt}
\begin{tabular}{@{}lrrrrrr@{}}
\multicolumn{7}{@{}l}{(c) The same runs, labels without PEAKS}\\
\toprule
 & & & \multicolumn{2}{c}{NovoCRF} &
\multicolumn{2}{c}{\begin{tabular}[b]{@{}c@{}}Labels that only\\one tool gets right\end{tabular}}\\
\cmidrule(lr){4-5}\cmidrule(l){6-7}
Peptide & & & one & two & NovoCRF, & \\
group & Labels & RNovA & pass & passes & two passes & RNovA\\
\midrule
Group 1 & 461 & 73.3 & 88.3 & \textbf{90.2} & 80 & 2\\
Group 2 & 471 & 63.7 & \textbf{79.4} & 79.2 & 74 & 1\\
Group 3 & 615 & 63.3 & \textbf{67.5} & 65.2 & 36 & 24\\
\addlinespace[2pt]
All three & 1{,}547 & 66.4 & \textbf{77.3} & 76.9 & 190 & 27\\
\bottomrule
\end{tabular}
\end{table*}

\subsection{Recall on modifications never seen in CRF training}\label{sec:ptm}
A user modification is an edge of the lattice at its exact mass (\S\ref{sec:userptm}). The CRF and the rerankers were
never trained on spectra of the modifications of this section: they learn user modifications from synthetic ones, made
by shifting peaks of ordinary training spectra (Appendix~\ref{app:training}). For the iRT and fragment-intensity predictors, training on such modifications
cannot be avoided entirely, because PROSPECT PTMs covers most modifications one can expect in an experiment. Of the
modifications here, hydroxyproline (Hyp) is one of the 28 modifications that the fragment-intensity predictor includes in its vocabulary,
learned from synthetic Hyp peptides of PROSPECT's hold-out split: the split has 151 Hyp peptide forms with HCD
spectra, and the predictor was trained on 85\% of its peptides (Appendix~\ref{app:training}). The iRT predictor was trained without that split and reads
Hyp as an unknown modification of its mass. Hydroxylysine and the four non-natural residues of RNovA's spike-in occur
in the training data of neither predictor. No result of this section uses CRSE.

\noindent\textbf{Hydroxyproline.} Collagen is rich in Hyp, whose residue mass (113.048\Da) lies 0.036\Da\ from
leucine. On the five hydroxyproline-rich sets, NovoCRF was given P+15.995 as a user modification, and
K+15.995 as well on the mouse collagen set. For the lion bone set, the labels are the search results that its authors
provided, from MaxQuant with Hyp as a variable modification~\cite{sun2026}. \Cfive, InstaNovo and $\pi$-PrimeNovo have no Hyp residue. RNovA accepts
user residues, so we added the same two residues to its residue table and ran it on the five sets as in the main
benchmark. Table~\ref{tab:ptm}a gives the recall of both tools, with residues matched within 0.01\Da.

On the four PEAKS-labeled sets, NovoCRF's recall on Hyp labels is 31--41\%, and 47--78\% on Hyp labels with at most
two missing bonds: these peptides are long, and only 20--46\% of the Hyp labels have that much evidence. RNovA's
recall is 14--25\% and 24--63\%. NovoCRF is ahead of RNovA in every cell of Table~\ref{tab:ptm}a, with one pass as
well as with two. On all labels, its default two passes lead by 34.3 points on enamel, 26.3 on lion bone, 18.4 on
tortoise bone, 17.0 on dentine and 15.4 on mouse collagen. Of the labels that only one of the two tools gets right,
NovoCRF has 79\% (mouse collagen, 377 labels against 99) to 94\% (tortoise bone, 1{,}878 against 118), and the sign
test gives $p<10^{-38}$ on every set. The second pass adds 3.8 points on enamel, 4.4 on dentine, 6.0 on tortoise bone
and 11.2 on lion bone. The mouse collagen run is too small to adapt to: with fewer than 1{,}000 confident answers it
is not decoded again, and only the final RT term and sibling fusion change its answers (\S\ref{sec:otf}).

\noindent\textbf{RNovA's spike-in.} RNovA's authors spiked 24 synthetic peptides, in three groups of eight, into a
293T digest~\cite{mao2026}. The peptides carry four non-natural residues, of 86.048, 100.064, 143.095 and 85.053\Da,
that no training set contains. NovoCRF was given the four as user modifications, and RNovA's deposited sequences name
the same four residues. The labels are the PEAKS identifications of the synthetic peptides, from a search of the
human proteome plus the 24 sequences with the four residues as variable modifications. Table~\ref{tab:ptm}b compares
the two tools. NovoCRF finds 74.7\% of the 1{,}693 labeled spectra in its default two passes and 75.1\% with one
pass, and RNovA 64.4\%. Of the labels that only one tool gets right,
NovoCRF with two passes has 201 and RNovA 27 ($p=5\times10^{-34}$). NovoCRF leads on every group, by 15.7 and 14.9
points on groups 1 and 2. On group 3 its lead of 2.4 points is within chance (42 labels against 26, $p=0.07$), and
the second pass costs 2.8 points there. The picture is the same with labels made without PEAKS, from the precursor mass and the b and y
ions of the known sequences (Table~\ref{tab:ptm}c): NovoCRF finds 76.9\% of the 1{,}547 labeled spectra and RNovA
66.4\%, and of the labels that only one tool gets right, NovoCRF has 190 and RNovA 27.

\begin{table*}[t]
\centering\footnotesize
\setlength{\tabcolsep}{5.5pt}
\caption{What NovoCRF fitted on each test run, from the run's own confident answers
(\S\ref{sec:otf}). Precursor: the systematic error of the precursor masses. Fragment: the error of the fragment masses
at \mz{} 500. NCE: the offset added to the collision energy that the raw file records, or unknown when the predictor
fits best without it. Intensity predictor: the normalized spectral angle between predicted and observed fragment
intensities on held-out confident answers, before and after fine-tuning. CRF: the negative log-likelihood per residue
on held-out confident answers. RT: the spread $\sigma$ of the held-out retention-time residuals, before and after
fine-tuning the iRT predictor. Siblings: share of the run's spectra with at least one sibling spectrum. $^a$Medians
over the eight runs. $^b$Corrected by a curve in \mz{}.}
\label{tab:adapt}
\begin{tabular}{@{}lrrrrrrrrrr@{}}
\toprule
 & \multicolumn{2}{c}{Mass error (ppm)} & & \multicolumn{2}{c}{Intensity predictor} & \multicolumn{2}{c}{CRF} &
 \multicolumn{2}{c}{RT $\sigma$ (s)} & Siblings\\
\cmidrule(lr){2-3}\cmidrule(lr){5-6}\cmidrule(lr){7-8}\cmidrule(lr){9-10}
Dataset & precursor & fragment & NCE & before & after & before & after & before & after & (\%)\\
\midrule
PXD076585, human & $-$0.90 & +0.9 & unknown & 0.740 & 0.781 & 0.379 & 0.329 & 170.2 & 109.7 & 52.4 \\
PXD071561, squid & +0.97 & +0.5 & +5 & 0.788 & 0.851 & 0.162 & 0.131 & 179.9 & 80.4 & 63.8 \\
PXD073929, frog & +2.68 & +2.9 & +4 & 0.691 & 0.725 & 0.445 & 0.382 & 152.7 & 97.2 & 43.3 \\
PXD072115, chickpea & $-$0.40$^b$ & $-$1.5 & $-$3 & 0.566 & 0.713 & 0.566 & 0.483 & 73.1 & 25.8 & 73.6 \\
PXD074750, \emph{C.\ auris} & +2.21 & +0.1 & +4 & 0.722 & 0.764 & 0.433 & 0.381 & 90.3 & 66.2 & 43.9 \\
PXD055085, antibody$^a$ & +0.79 & +1.0 & +1.5 & 0.680 & 0.771 & 0.419 & 0.291 & 248.5 & 89.9 & 50.9 \\
PXD071022, HLA-I & +1.53 & +1.2 & +3 & 0.771 & 0.796 & 0.420 & 0.342 & 154.1 & 94.6 & 71.0 \\
PXD082505, HLA-I & +2.89 & +3.0 & +2 & 0.747 & 0.789 & 0.500 & 0.378 & 209.5 & 143.6 & 92.8 \\
PXD054498, HLA-I & $-$7.57$^b$ & $-$10.4 & $-$3 & 0.601 & 0.727 & 0.583 & 0.473 & 71.0 & 55.4 & 80.4 \\
PXD064980, HLA-II & $-$2.89 & $-$3.0 & $-$1 & 0.575 & 0.683 & 0.734 & 0.620 & 130.8 & 87.5 & 77.6 \\
PXD062962, phospho & +0.27$^b$ & $-$1.5 & unknown & 0.842 & 0.881 & 0.135 & 0.095 & 198.6 & 89.0 & 66.7 \\
PXD062859, phospho & $-$0.16 & $-$1.0 & +2 & 0.826 & 0.874 & 0.153 & 0.104 & 173.9 & 80.8 & 70.7 \\
\bottomrule
\end{tabular}
\end{table*}

\subsection{What NovoCRF adapts in each run}\label{sec:adapt}
Table~\ref{tab:adapt} lists what NovoCRF fitted on each test run, taken from the logs of the runs behind
Table~\ref{tab:main}. Every value comes from the run's own confident answers, without labels.

\noindent\textbf{Mass shift.} The precursor masses of the twelve datasets are off by $-7.6$ to $+2.9$\,ppm, and the
fragment masses at \mz{} 500 by $-10.4$ to $+3.0$\,ppm. The pleural-effusion run has the largest shift. Its precursors
are 7.6\,ppm low, with a trend along \mz{} that a curve corrects better than one fixed value, and its fragments are
10.4\,ppm low at \mz{} 500 and 17.9\,ppm low at \mz{} 200.

\noindent\textbf{Fragmentation pattern.} On 17 of the 19 runs, the intensity predictor fits best with an offset of
$-3$ to $+5$ on the recorded NCE, and on two with the NCE left unknown. Before fine-tuning, its spectral angle on
held-out confident answers is 0.57 to 0.84, and it is lowest on the three timsTOF sets. Fine-tuning raises it on every
run, by 0.03 to 0.15, to 0.68 to 0.88. The fine-tuned CRF lowers the held-out negative log-likelihood per residue by
12\% to 32\%.

\noindent\textbf{Elution order.} Fine-tuning the iRT predictor on a run lowers the spread of the held-out
retention-time residuals by 22\% to 69\% over the 19 runs, for example from 180\,s to 80\,s on squid.

Each fine-tuned model passed its held-out check on all 19 runs, and the adaptation between the two passes took 1.3
to 25.2\,s per run.

\noindent\textbf{Sibling spectra.} Between 43\% and 93\% of a run's spectra have at least one sibling, and the fusion
after the first pass changes the answer of 8\% to 46\% of a run's spectra. With sibling spectra enhancement switched
off (Appendix~\ref{app:nosib}), the recall of the two passes is 5.9 to 27.5 points lower on the twelve sets. Over the
twelve sets, the enhancement gains 35{,}008 labels and loses 2{,}759. It adds most on poorly fragmented spectra. Pooled over the sets, it raises the
recall from 91.7\% to 93.8\% on labels without a missing bond, from 60.3\% to 77.0\% with two, from 37.3\% to 63.5\%
with three, from 14.9\% to 46.0\% with four or five and from 1.3\% to 18.0\% with six or more.

\begin{table}[t]
\centering\footnotesize
\setlength{\tabcolsep}{4pt}
\caption{Ablations: mean peptide recall (\%) over development sets, on all labels and on labels with at most two
missing bonds. The three other CRFs were trained like the released one, on the same data with the same seed, each
with one change. With CRSE, NovoCRF also reads two replicate runs of each set's run. Best of each block in bold.}
\label{tab:ablation}
\begin{tabular}{@{}lrr@{}}
\toprule
 & & $\le$2 missing\\
Model & All & bonds\\
\midrule
\multicolumn{3}{@{}l}{\emph{CRF alone (best path of the lattice), nine sets}}\\
Global CRF with source term (released) & & \textbf{45.31}\\
Locally normalized (MEMM), N$\to$C & & 34.55\\
Locally normalized (MEMM), C$\to$N & & 39.40\\
Global CRF without source term & & 42.84\\
\addlinespace[2pt]
\multicolumn{3}{@{}l}{\emph{Released model, all 19 sets}}\\
One pass & 63.81 & 83.83\\
Two passes (default) & \textbf{69.33} & \textbf{88.32}\\
\addlinespace[2pt]
\multicolumn{3}{@{}l}{\emph{Released model, the 11 sets with replicate runs}}\\
Two passes (default) & 65.32 & 87.17\\
Two passes\,+\,CRSE & \textbf{70.50} & \textbf{89.97}\\
\bottomrule
\end{tabular}
\end{table}

\subsection{Ablations}\label{sec:ablation}
Table~\ref{tab:ablation} reports the ablations as mean peptide recall over development sets.

\noindent\textbf{Global normalization and source term.} We trained the CRF part three more times like the released
CRF, on the same data with the same seed, each time with one change, and scored the best path of the lattice without
the rerankers on nine development sets, on labels with at most two missing bonds. A softmax over the edges that leave each bin, or over those that enter it, in place of the global
normalizer gives a maximum-entropy Markov model on the same lattice, features and scorer. Its recall is 34.6\% from N
to C and 39.4\% from C to N, against 45.3\% for the CRF. It is lower than the CRF's on all nine sets from N to C and
on eight from C to N. Without the source term, the CRF's recall is 42.8\%, 2.5 points lower, with a loss on each of
the nine sets.

\noindent\textbf{Passes and runs.} On the 19 development sets, the released model reaches a mean recall of 63.8\% in
one pass and 69.3\% in its default two passes, and 83.8\% and 88.3\% on labels with at most two missing bonds. On the
test sets the second pass adds 0.7--9.7 points on all labels (Table~\ref{tab:main}). CRSE was measured on the
15 development sets that have two companion runs by the rule of \S\ref{sec:protocol}. On the 11 sets whose companions
are replicates, it raises the mean recall from 65.3\% to 70.5\%, with a gain on each set, and on the four others by
2.0 points.

\noindent\textbf{Sibling spectra.} With sibling spectra enhancement switched off, the recall of the released model's
two passes falls by 5.9 to 27.5 points on the twelve test sets (\S\ref{sec:adapt}, Appendix~\ref{app:nosib}).

\noindent\textbf{Rerankers and predictors.} Two inputs of the rerankers were worth the most during development,
measured in the same way on earlier versions of the model, on labels with at most two missing bonds. The similarity between predicted and observed intensities
over the 12 most intense predicted ions, instead of over all ions, gained 0.9 points. An intensity predictor that
reads the precursor charge, the collision energy and the modifications gained 1.6 points over one that reads only the
sequence. Distilling the third stage reranker from 384 to 128 units did not cost reranking recall, the share of the
validation lists in which a reranker ranks the correct candidate first: the student reaches 0.680 and its teacher
0.679 (Appendix~\ref{app:training}).

\subsection{Speed, model size and training cost}\label{sec:speed}
NovoCRF is lightweight: its 1.80M parameters are 18 to 53 times fewer than those of the transformer models
(Table~\ref{tab:speed}), and it trains in about 21 GPU-hours. With a GPU, its default two passes sequence the twelve
main test sets at 313 spectra/s, which is 2.3 times the throughput of $\pi$-PrimeNovo, the fastest transformer model, and
11, 20 and 33 times that of \Cfive, InstaNovo and RNovA. One pass runs at 676 spectra/s. On the CPU alone, NovoCRF
runs at \cputwo\ spectra/s in two passes and \cpuone\ in one pass. There each sibling offers one candidate instead of
five (\S\ref{sec:sibling}). On the twelve main test sets, the recall of the two passes on the CPU alone lies between
0.9 points below that with a GPU (HLA-II) and 0.8 points above it (the B*15:02 HLA-I set).
These throughput rates include reading the raw files. With CRSE, the three runs of a set are decoded together, at 162
spectra/s over the eight sets, against 297 spectra/s for the eight test runs alone. On the CPU alone, CRSE
runs at \cpux\ spectra/s. On these eight sets, its recall lies between 1.0 points below that of CRSE with a GPU
(HLA-II) and 0.2 points above it (mouse liver and the B*15:02 HLA-I set).

\begin{table}[t]
\centering\footnotesize
\setlength{\tabcolsep}{1.7pt}
\caption{Model size and throughput. Parameters: learnable parameters, counted by loading each tool's released
checkpoint into its own model class (fixed sinusoidal encodings are not counted). Transformers: encoder\,+\,decoder
layers and model width $d$. Ratio: parameters relative to NovoCRF. Spectra/s: over the twelve main test sets, as in
Table~\ref{tab:main}, NovoCRF in its default two passes with the GPU and on the CPU alone. $^a$The model that the
benchmark runs. RNovA's path-search model is not included.}
\label{tab:speed}
\begin{tabular}{@{}llrrr@{}}
\toprule
Model & Network & Parameters & Ratio & Spectra/s\\
\midrule
\textbf{NovoCRF} & CRF and five BiGRUs & \textbf{1.80M} & 1 & \\
\quad with GPU & & & & \textbf{313}\\
\quad CPU only & & & & \cputwo\\
\addlinespace[2pt]
$\pi$-PrimeNovo & transformer, 9\,+\,9, $d=400$ & 32.5M & 18.0 & 136\\
CasanovoV5 & transformer, 9\,+\,9, $d=512$ & 47.9M & 26.6 & 28.2\\
RNovA SeqFiller$^a$ & transformer, 6\,+\,6, $d=768$ & 61.3M & 34.1 & 9.35\\
InstaNovo v1.2.0 & transformer, 9\,+\,9, $d=768$ & 94.6M & 52.6 & 15.3\\
\bottomrule
\end{tabular}
\end{table}

NovoCRF's workload fits a modern system on a chip: its networks are small enough for both the GPU and the
neural processing unit of a laptop, while its dynamic programming and search keep the CPU cores busy, so all three units work
at once. On a laptop with an Apple M2 Max (12 CPU cores, 30-core GPU, 16-core Neural Engine), the colorectal HLA-I
set PXD071022 (85{,}732 spectra, two passes) takes 131.5\,s (the mean of two runs), or 652 spectra/s. The Neural Engine runs the CRF scorer, the
intensity predictor and the third stage reranker. The GPU runs the first stage reranker, the iRT predictor and the
lattice recursions of the CRF fine-tuning, and the CPU the rest.
Over the run, NovoCRF keeps 9.3 of the 12 CPU cores busy on average, the GPU has work 76\% of the time, and the
Neural Engine has a prediction in flight about 60\% of the time. Without the
Neural Engine the run takes 155.5\,s, and on the CPU cores alone 285.8\,s. The desktop computer of Table~\ref{tab:main} takes
94\,s for the same file with its RTX 3090.

% ================================================================================================
\section{Discussion}

\noindent\textbf{Adapting to the run.} A model that is trained once meets every run with the same weights, and the
runs differ. On our twelve test sets, the precursor masses are off by $-7.6$ to $+2.9$\,ppm, the intensity predictor
fits best at collision energies up to 5 units away from the recorded one, and the pretrained iRT predictor leaves
retention-time residuals with a spread of about 70 to 250\,s before it is tuned to the run (Table~\ref{tab:adapt}). NovoCRF
measures each of these on the run's own confident answers, without labels, and adapts to the run. The calibration and
fine-tuning between the two passes cost 1 to 25\,s of computation per run. The
second pass that follows adds 0.7 to 9.7 points of recall. The three timsTOF sets, on which the unadapted intensity
predictor fits worst, are also the sets with the three largest margins over the other tools.

\noindent\textbf{What peptide recall rewards.} Peptide recall against database-search labels mixes two abilities:
reading a peptide from its fragment ions, and supplying what the fragment ions of one spectrum do not tell. A database search chooses
among candidates that already have genome-level support, so it can accept a PSM with long unsupported stretches (three
or more residues with no b or y ion between them, found in 38\% of our labels). A \textit{de novo} tool that gets such
a label right has filled the stretch from another source. One source is a sequence prior, which works best on
peptides the model has seen: with six or more missing bonds, the transformer models' recall is up to 40 points higher
on their own training peptides (\S\ref{sec:seen}). The other source is the experiment itself, the sibling spectra
of the same peptide, which NovoCRF uses. It serves peptides that no tool has seen during training as well: NovoCRF correctly sequences 17.0\% of
their spectra with six or more missing bonds, and the four transformer models at most 2.5\%. Every tool, NovoCRF
included, has a higher recall on the peptides of its own training data, which is one more reason to report recall on
seen and unseen peptides separately.

\noindent\textbf{One job per module.} A large end-to-end model can partly compensate for a wrong precursor mass
(Appendix~\ref{app:precursor}). The CRF cannot correct a wrong mass and does not try to: it searches only peptides that
fit the mass it is given, so when that mass is wrong it almost never finds the right peptide. We fix the input instead
of teaching the search to distrust it. NovoCRF uses a modular design, and each module can be checked and improved on
its own: PrecursorMassFinder,
the rerankers and the predictors were each developed and measured separately
(\S\ref{sec:ablation}, Appendix~\ref{app:precursor}).

\noindent\textbf{Novel peptides.} \textit{De novo} sequencing matters most for sequences that are not in a reference: a
neoantigen's mutation, an antibody's somatic hypermutation, an unexpected modification. NovoCRF's lattice search
returns the sequences best supported by the spectrum. NovoCRF has the highest recall in all twelve datasets on the
spectra with peptides that no tool has seen (Table~\ref{tab:novel}), and it searches modifications unseen in CRF
training at almost no extra cost (\S\ref{sec:ptm}).

\noindent\textbf{Limitations.} The comparison of seen and unseen labels rests on each tool's documented training data,
which for InstaNovo is that of an earlier version, and on exact peptide identity. The evidence analysis counts only b
and y ions and trusts the label inside unsupported stretches. Sibling spectra are linked by mass and retention time
alone, so peptides of equal mass that elute together, such as positional isomers of a modification, can be linked,
and the answer of one can replace the answer of the other.

\section{Conclusion}
NovoCRF is an integrated \textit{de novo} sequencing workflow that adapts to each MS/MS run. It combines a globally
normalized CRF over engineered physics features with small rerankers and two small predictors. The run's masses are
recalibrated, and the CRF and the predictors are fine-tuned on the run's own confident answers. The run's
retention times and the sibling spectra of each peptide serve as further evidence. The model has 1.80M parameters in all and trains in about 21 GPU-hours. On twelve held-out datasets NovoCRF has
the highest peptide recall on eleven, and on the twelfth as well once it reads two replicate runs. On labels with at most one
missing bond it outperforms InstaNovo, RNovA, $\pi$-PrimeNovo and \Cfive\ on every dataset, and it also does so on the
labels whose peptide is in no tool's training data. For labeled spectra with poor fragment ion support, the transformer models
are right mostly on peptides of their training data, and NovoCRF draws on the sibling spectra of the run, which raise
its recall on seen and on unseen peptides. NovoCRF sequences
313 spectra/s with a GPU and also runs on the CPU alone, at \cputwo\ spectra/s. Our detailed analysis suggests
that reporting recall by fragment evidence and by seen and unseen peptides would make
\textit{de novo} benchmarks more informative than the headline recall alone.

\noindent\textbf{Availability.} The compiled binaries of NovoCRF for Windows and Linux x86\_64 are provided at
\url{https://github.com/volpato30/NovoCRF-release}. The results in this paper were generated with NovoCRF v0.1.0.
Test data: PRIDE (Table~\ref{tab:data}).

% ------------------------------------------------------------------------------------------------
{\footnotesize\frenchspacing
\setlength{\bibsep}{0.5pt plus 0.3pt}

}

% ================================================================================================
\clearpage
\appendix
\section{Bin features of the CRF}\label{app:features}
For an active bin $s$ with prefix mass $x=s\Delta$, ion type $j$ (series b or y, charge $z_j$, mass shift $\delta_j$)
has the theoretical \mz{}
\begin{equation}
\mu_j(s)=\big(n_j(x)+\delta_j+z_j\mH\big)/z_j,
\label{eq:ion}
\end{equation}
where the fragment's neutral mass is $n_j(x)=x$ for the b series and $n_j(x)=M-x$ for the y series.

\noindent\textbf{Ion-match features ($23\times3$).} The 23 ion types, with their shifts $\delta_j$, are b and y of charge
1--3 ($\delta_j=0$), a$^{+}$ and a$^{2+}$ ($-$CO, $-27.995\Da$), b and y of charge 1 and 2 minus H$_2$O ($-18.011\Da$)
or NH$_3$ ($-17.027\Da$), b$^{+}$ and y$^{+}$ minus H$_3$PO$_4$ ($-97.977\Da$), which is the neutral loss of
phosphopeptides, and five more singly charged ions (a$-$NH$_3$, z$+$1, b$+$H$_2$O, b$-$H$_2$O$-$NH$_3$,
y$-$H$_2$O$-$NH$_3$) whose shifts follow from their formulas. The peak nearest to $\mu_j(s)$ within 50\,ppm, with
relative intensity $I$ and signed error $\epsilon$ in ppm, gives three scalars: the encoded intensity
$\min(1,\max(0,1+\tfrac{1}{3}\log_{10}I))$, the proximity $\exp(-|\epsilon|/10)$ and the error $\epsilon/50$. All three
are 0 if no peak lies within 50\,ppm.

\noindent\textbf{Isotope features ($4\times2$).} For the primary ions b$^+$, y$^+$, b$^{2+}$ and y$^{2+}$, the peaks at
$\mu_j\pm1.00335/z_j$ (intensities $I_{+1}$, $I_{-1}$) are compared with the averagine expectation that a fragment of
neutral mass $n$ has a first isotope peak $\rho(n)\approx5\times10^{-4}\,n$ times as intense as its monoisotopic peak:
$\log\!\big(I_{+1}/(\rho(n)\,I)+0.05\big)$ is near 0 for a consistent isotope envelope, and
$\log\!\big(\rho(n-1.00335)\,I_{-1}/I+0.05\big)$ is large when the matched peak is itself the $+1$ isotope of a lighter
fragment. Both are clipped to $[-3,3]$ and divided by 3.

\noindent\textbf{Global features (9).} Each bin also carries its relative position $x/M$, the precursor mass $M/3000$, two
flags for whether b$^+$ and y$^+$ fall inside the observed \mz{} range, and the charge one-hot (1, 2, 3, 4, $\ge$5). In
total, $23\times3$ ion-match, $4\times2$ isotope and 9 global features form $\bx(s)\in\R^{86}$.

\section{Exact training: the backward recursion}\label{app:train}
For a fixed edge, the reads $\alpha(s-o_k)$ of Eq.~\eqref{eq:fwd} over consecutive $s$ are a contiguous stream at a
fixed offset, not a gather, and $\psi_k(s)$ is recomputed on the fly from the 8-d $\bv(s)$ inside the kernel. The
gradient of $\log Z$ needs the mirror image of $\alpha$: $\alpha(s)$ sums the scores of all ways to reach bin $s$, and
$\beta(s)$ those of all ways to finish a path from $s$ (the remaining edges up to an end bin $t$, plus $\tau(t)$). The
backward recursion
\begin{equation}
\begin{aligned}
\beta(s)&=\log\textstyle\sum_{k}e^{\psi_k(s+o_k)+\beta(s+o_k)},\\
\beta(t)&=\tau(t)\quad\text{for }|t-t^\ast|\le W,
\end{aligned}
\label{eq:bwd}
\end{equation}
runs from the terminal window down to bin 0 over the same blocks in reverse order. Every path through edge $k$ into
bin $s$ is a prefix to $s-o_k$, the edge and a suffix from $s$, so the gradient is that edge's posterior probability,
$\partial\log Z/\partial\psi_k(s)=\exp(\alpha(s-o_k)+\psi_k(s)+\beta(s)-\log Z)$. Fused kernels accumulate it into the
gradients of $\bv(s)$, $\be_k$, $\bg_k$ and $c_k$.

\section{Reranker inputs and calibrators}\label{app:rerank}
\noindent\textbf{Network.} Residue $i$ of a candidate enters as $[\,\mathrm{tok}(a_i);\ \mathrm{ReLU}(\mathbf{W}\bx_i+\bm{b})\,]$,
a 32-d token embedding and a 32-d projection of its features $\bx_i$, with the embedding of the candidate's N-terminal
modification added at the first position. A one-layer BiGRU with $H$ units per direction reads the candidate. The mean
and the maximum of its states over the residues ($4H$ values) and the candidate inputs feed a layer of $H$ ReLU units
and a linear output $u$. The score is $u+w\,(\Phi-\Phi_{\max})$, with a learned $w$ and $\Phi_{\max}$ the best CRF score in the
list. $H$ is 64, 128 and 128 in the three rerankers.

\noindent\textbf{Per-residue features.} Fragment intensities below are encoded as $(\log_{10}I+4)/4$ for a relative
intensity $I$ matched within 20\,ppm, and 0 when no peak matches.
\begin{itemize}
\item First stage reranker (86): the bin features $\bx(s)$ at the cleavage after the residue, and zeros for the last
  residue.
\item Second stage reranker (93): the 86 plus seven intensities, namely the residue's immonium ion and its immonium
  ion minus NH$_3$, the single-residue internal b ion, the most intense internal b ions of 2--4 residues, touching
  neither terminus, that start and that end at the residue, and the most intense internal a ions of 2--4 residues
  that start and that end at the residue.
\item Third stage reranker (97): the second stage reranker's 93 with internal b ions of up to 6 residues, plus the
  predicted and the observed b$^+$ and y$^+$ intensities at the cleavage after the residue (square roots, each divided
  by the candidate's largest predicted or observed value).
\end{itemize}

\noindent\textbf{Candidate inputs.} The first and second stage rerankers read eight, the last five of which are the
candidate's fit statistics:
\begin{itemize}
\item the CRF score gap to the list's best,
\item the CRF log-probability $\Phi-\log Z$, clipped to $[-50,0]$ (inference does not compute $\log Z$,
  \S\ref{sec:inference}, so this input is fixed at $-50$ and acts as an intercept),
\item the length divided by 20,
\item the share of the total intensity explained by b and y ions of charge 1 and 2 within 0.02\Da,
\item the share of the 20 most intense peaks explained,
\item the share of cleavages with b$^+$ or y$^+$ evidence,
\item the longest run of cleavages without evidence, divided by 10,
\item the intensity share of internal b fragments of 2--4 residues.
\end{itemize}
The third stage reranker reads 27:
\begin{itemize}
\item these eight, with the gap taken within its 20 candidates,
\item the second stage reranker's score relative to its best, clipped to $[-20,0]$ and divided by 5, and the rank in
  the second stage reranker's order, divided by 10,
\item the candidate's five fit statistics minus those of the second stage reranker's top candidate, so that they are
  zero for that candidate,
\item five list-contrast features, namely the share of the intensity that any ion of the candidate explains, and the
  number (divided by 10) and intensity share of the peaks that the candidate explains and the second stage reranker's
  top candidate does not, and of those that the top candidate explains and the candidate does not, counting b, a and y
  ions (and their losses of water and ammonia), immonium ions and internal b and a ions of 1--4 residues,
\item seven similarity features between the predicted and observed intensities of the 12 ions with the highest
  predicted intensity: the normalized spectral angle, absolute and relative to the list's best and to the second
  stage reranker's top candidate, the spectral entropy similarity~\cite{li2021}, absolute and relative to the list's
  best, and the Spearman rank correlation, absolute and relative to the list's best.
\end{itemize}

\noindent\textbf{Calibrators.} NovoCRF uses two logistic calibrators. The first gives the calibrated confidence that
NovoCRF's own decisions read, $\sigma(\bm{w}^{\top}\bm{f})$ for 15 features $\bm{f}$ of the chosen candidate: its log-softmax
among the third stage reranker's scores and that reranker's top-1 minus top-2 score (clipped to $[0,20]$ and divided
by 10), the same two for the second stage reranker's scores, its clipped CRF log-probability divided by 10 (fixed at
$-5$, as above), its CRF score gap divided by 10, its first four fit statistics (all but the internal-fragment
share), the charge one-hot ($\le$2, 3, $\ge$4), its length divided by 20, and an intercept. The weights are fitted by
$L_2$-regularized logistic regression ($\lambda=10^{-3}$) to exact peptide matches on a validation set of 9{,}894
spectra, decoded in one pass without the RT term, the sibling fusion and the mass calibration.

The second calibrator gives the reported confidence. Its logit is the logit of the first plus 29 weighted inputs,
each zero when its evidence is absent:
\begin{itemize}
\item six on the RT term, among them its value for the answer and whether it changed the answer,
\item nine on the sibling fusion, among them the number of siblings, the number that agree with the answer, and
  whether the fusion changed the answer,
\item six on the cross-run siblings, among them their number and the number that agree with the answer,
\item eight for the answers that took part in a fusion, among them the answer's log-softmax under the fused score
  and under the spectrum's own score.
\end{itemize}
Without an RT term and without fusion, the two confidences are equal. The 29 weights are fitted by logistic
regression to labeled answers of the development sets.

\section{Training of each component}\label{app:training}
All components are trained with AdamW (weight decay $10^{-4}$), a one-cycle learning-rate schedule and gradient
clipping at norm 1, on single GPUs. The CRF and the rerankers are trained on the data of \S\ref{sec:protocol} by one
routine, which trains the components below in a fixed order. The iRT predictor is trained on PROSPECT PTMs~\cite{gabriel2024},
and the intensity predictor on PROSPECT PTMs mixed with NovoCRF's own training spectra.

\noindent\textbf{CRF.} The CRF is trained for 80k steps from scratch, with a warm-up over 2\% of the steps to a
learning rate of $2\times10^{-3}$. Batches hold a variable number of spectra, capped at $8\times10^{7}$ lattice bins
and 256 spectra, 84 on average. Training uses the 400 most intense peaks of each
spectrum, the phosphorylated residues with the bin features of b$^{+}$ and y$^{+}$ minus H$_3$PO$_4$, shared
parameters for unmodified and carbamidomethylated C, and synthetic modifications.

\noindent\textbf{Synthetic modifications.} Each batch draws four modifications. Each has a base residue, drawn uniformly
from the 19 residues, and a mass shift: with probability 0.5 one of 41 known modification masses, otherwise uniform on
$[-20,250]\Da$, with shifts below 0.5\Da\ in magnitude rejected with probability 0.7. A modified residue must weigh at
least 57\Da\ and lie more than 0.03\Da\ from every built-in edge, start edge and other modification of the batch. A
quarter of the spectra get one of them at a random occurrence of its base residue. Every peak that matches only
fragments containing it (b, a and y ions with losses of water, ammonia or CO, charges 1 to $z-1$, the first two
$^{13}$C isotopes, within 20\,ppm) moves by the shift divided by its charge, and the precursor moves by the shift. The
lattice of that spectrum gets an edge for the modified residue, scored as in \S\ref{sec:userptm}.

\noindent\textbf{Rerankers.} The inference engine itself makes the candidate lists, with every step of
\S\S\ref{sec:otf}--\ref{sec:sibling} switched off. The first stage reranker is trained for 150k steps, at a learning
rate of $10^{-3}$ on batches of 64 lists, on the CRF's top-50 lists of 1.7M training spectra. The second stage
reranker is trained for 150k steps, at $10^{-3}$ on batches of 32 lists, on the CRF's top-100 lists of 1.2M of these
spectra. The third stage reranker reads the 20 candidates that the second stage reranker ranks highest among the
CRF's 200 best: 4.62M lists of ordinary training spectra,
480k lists of spectra with a synthetic modification and 144k lists of unmodified spectra searched with a synthetic
modification offered. The intensity predictor stays frozen during this
training and reads every spectrum's collision energy as unknown. With probability 0.5, a sampled spectrum with at
least 10 peaks has its intensities raised to a power drawn uniformly from $[0.5,1.5]$ and about 10\% of its peaks
other than the base peak dropped before the third stage reranker computes its features. Each reranker keeps the step
with the best validation reranking recall.

\noindent\textbf{Distillation of the third stage reranker.} The third stage reranker is trained twice, each time for
150k steps at
$5\times10^{-4}$ on batches of 64 lists. A teacher with 384 units per direction is trained first. The 128-unit student
that NovoCRF uses is then trained on the same lists with the loss
$\tfrac12\,\mathrm{CE}+\tfrac12\,T^{2}\,\mathrm{KL}\big(\mathrm{softmax}(\bm{t}/T)\,\|\,\mathrm{softmax}(\bm{s}/T)\big)$,
where $\bm{s}$ and $\bm{t}$ are the student's and the frozen teacher's scores of a list, CE is the cross-entropy of
the correct candidate under the student, KL is the Kullback--Leibler divergence and $T=2$. The second term also covers the lists without the correct
candidate. The student's validation reranking recall is 0.680 and the teacher's 0.679.

\noindent\textbf{Cost.} Training takes about 21 GPU-hours on one A100 in all: 5.70 for the CRF, 9.64 for the four
reranker trainings, 2.34 for their candidate lists, 2.45 for the intensity predictor and 0.61 for the iRT predictor.

\noindent\textbf{Residue encoding of the predictors.} A residue (I read as L) enters as its amino acid's embedding
plus, if it is modified, the embedding of its modification and an MLP ($3\to32\to64$) of the mass shift $\Delta m$,
whose inputs are $\Delta m/100$, $(\Delta m/100)^2$ and $\mathrm{sign}(\Delta m)\ln(1+|\Delta m|)/5$. A named
(residue, modification) pair, matched within 0.01\Da, has its own embedding, and any other modification shares an
unknown embedding. During training, each named modification is read as unknown with probability 0.25, so that the
mass-shift network learns to stand in for the name. N-terminal modifications have their own table (none, unknown,
acetyl, TMT) and mass network, added to the first residue. The intensity predictor names 28 pairs, the iRT predictor
the first 15 of them:
\begin{itemize}
\item carbamidomethyl C, oxidized M, TMT6plex K, GlyGly K, phosphorylated S, T and Y, acetyl K, methyl R and K,
  pyro-Glu from Q and from E, deamidated R (citrulline), and HexNAc on S and T,
\item PROSPECT's 13 hold-out pairs: dimethyl R and K, trimethyl K, malonyl, butyryl, glutaryl, formyl,
  hydroxyisobutyryl, crotonyl, succinyl and propionyl K, hydroxylated P, and nitrated Y.
\end{itemize}

\noindent\textbf{Intensity predictor.} The charge (1--6) and the NCE, through an MLP ($1\to16\to64$) of
$(\mathrm{NCE}/100-0.30)/0.05$ or a learned vector when the NCE is unknown, add embeddings at every position. The
engine rounds NCEs to whole numbers between 10 and 60. The loss is $1-\mathrm{SA}$, with the normalized spectral angle
$\mathrm{SA}=1-2\arccos(\cos\theta)/\pi$ between the square roots of the predicted and observed intensities over the
unmasked ions. The predictor was first trained with the 15 pairs of the iRT predictor, for 100k steps at a learning rate of
$10^{-3}$, and then extended to all 28, starting the 13 new embeddings from the unknown one, for 50k steps at
$5\times10^{-4}$, both with a 3\% warm-up. Each step mixes 768 HCD spectra of peptides of up to 30 residues from the PROSPECT PTMs MS2 training
split~\cite{prospectms2} and 256 of NovoCRF's training spectra of 2--40 residues with at least 5 peaks, whose NCE is unknown.
In the second phase, each step also holds 128 spectra from 85\% of PROSPECT's hold-out split (split by peptide), which carries
the 13 added modifications.
PROSPECT gives the intensities of b and y ions of charge 1 and 2, and the a ion is masked. For NovoCRF's spectra, the
observed intensity of an ion is the largest relative intensity within 20\,ppm among the 400 most intense peaks, and
ions of charge 2 from singly charged precursors, ions outside 50--2{,}500 \mz{} and ions within 2 \mz{} of the
precursor are masked. The kept model has a mean validation SA of 0.899 on 20k PROSPECT spectra and 0.774 on NovoCRF's
9{,}894 validation spectra.

\noindent\textbf{iRT predictor.} Mean and max pooling of the BiGRU's states over the residues feed an MLP
($256\to64\to1$), whose output is scaled by 40 and shifted by 55 to the iRT scale. It was trained for 60k steps of
1{,}024 rows drawn uniformly from the training split of the PROSPECT PTMs retention-time dataset~\cite{prospectirt}
(1.04M rows before peptides with residues or modifications outside its tables or with more than 30 residues are
removed), with an L1 loss and a learning rate of $10^{-3}$ with a 3\% warm-up. The kept step has the lowest mean
absolute error on the validation split, 2.94 iRT units. On the test split the error is 2.90 iRT units, and a cubic fit of
retention time against predicted iRT in each raw file leaves an error of 0.76\,min. On PROSPECT's hold-out modifications, all of
which it reads as unknown, the error is 8.44 iRT units.

\section{On-the-fly adaptation in detail}\label{app:otf}
\noindent\textbf{Confident answers.} An answer is confident if its calibrated confidence is at least the 70th percentile of
the run's answers, it has at least six residues, and at most one of its bonds lacks a b$^+$ or y$^+$ ion (also 2+ for
precursors of charge 3 or more) within 0.02\Da\ among all peaks of the spectrum. For the CRF,
$\min(N/2,\max(200,N/5))$ of the $N$ answers are held out, and for the predictor 20\%. An answer outside the
confident set with at least six residues joins the fine-tuning examples, but not the calibrations, when one of its
siblings inside the set has the same peptide (\S\ref{sec:sibling}).

\noindent\textbf{Order of the steps.} After the first pass, its RT term and its sibling fusion, NovoCRF selects the
confident answers. It
fits the precursor and fragment calibrations on the uncalibrated spectra and applies both to all spectra, and builds the
fine-tuning examples from the calibrated spectra. On these examples it fine-tunes the CRF, and it calibrates the collision energy and then fine-tunes the
predictor at it. The two fine-tunings do not depend on each other. If any of these steps changed something, it decodes
all spectra again. After the last pass it fine-tunes the iRT predictor, refits the RT term and repeats the sibling
fusion, with the links made again from the recalibrated masses.

\noindent\textbf{Precursor calibration.} Each confident answer gives the error
(theoretical $-$ measured)/measured in ppm, and each distinct peptide keeps the median of its answers' errors. With at
least 100 peptides, the densest window of 4\,ppm among their sorted errors gives a center $c$. If at least 50 errors lie
within $c\pm4$\,ppm and their density is at least three times that of the errors within $c\pm20$\,ppm but outside
$c\pm4$\,ppm, the median of the errors within $c\pm4$\,ppm corrects every precursor of the run. Otherwise there is no
correction. A cubic polynomial in \mz{} replaces the single value when it lowers the robust spread of the errors by at
least 10\%. It is fitted by least squares to the most confident third of the confident answers, on the errors within
25\,ppm of the single value (at least 200), with \mz{} clipped to their central 98\% and two refits without the points
beyond three robust standard deviations.

\noindent\textbf{Fragment calibration.} Every b$^+$ and y$^+$ ion of every confident answer (also 2+ for charge 3 or
more) takes its nearest peak within 30\,ppm. With at least 500 such points, the error in ppm is fitted by least squares
as a polynomial of degree 4 in $x=$\,\mz/1000, with \mz{} clipped to the 0.5th and 99.5th percentiles of the points.
The fit is repeated twice without the points beyond three robust standard deviations, and the fitted error is divided
out of every peak of the run.

\noindent\textbf{Collision-energy calibration.} Thermo HCD scans carry the NCE set in the method. A timsTOF collision energy $E$ in
eV is converted to $\mathrm{NCE}=500E/(m\,c_z)$ for the precursor \mz{} $m$, with $c_z=1$, 0.9, 0.85, 0.8 and 0.75
for charges 1, 2, 3, 4 and $\ge5$. Other activation types give an unknown NCE. The first pass uses the
raw NCE. Up to 4{,}000 confident answers, evenly spaced through the fine-tuning set, are then scored by the unadapted
predictor under each candidate setting: unknown, and either offsets of $-8$ to $+8$ in steps of 1 added to the raw NCE,
if at least half of these answers have one, or a single NCE of 18 to 42 in steps of 2 for all spectra. The setting
with the highest mean spectral angle is used from then on, and a changed setting alone is enough for a second pass.

\begin{table*}[t]
\centering\footnotesize
\setlength{\tabcolsep}{4pt}
\newcommand{\fn}[1]{{\scriptsize\nolinkurl{#1}}}
\caption{Companion runs of the main test sets. For each dataset, the test run comes first and its two companion
runs follow. The last four datasets have no two runs that fit the rule.}
\label{tab:companions}
\begin{tabular}{@{}p{3.3cm}p{8.4cm}p{5.2cm}@{}}
\toprule
Dataset & Test run and companion runs & Relation of the companions to the test run\\
\midrule
\multicolumn{3}{@{}l}{\emph{Replicates}}\\
PXD076585, human & \fn{C1_1_03.raw}\newline \fn{C1_2_04.raw}\newline \fn{C2_1_07.raw} &
second injection of the same control sample, and the second biological replicate of the control\\
PXD071561, squid & \fn{ex017363.raw}\newline \fn{ex017368.raw}\newline \fn{ex017373.raw} &
biological replicates 2 and 3 of the same tissue, enzyme and fraction\\
PXD072115, chickpea & \fn{2023_04_13_IBERA_S.1.5_S3-B1_9284.d}\newline \fn{2023_04_13_IBERA_S.2.5_S3-B2_9285.d}\newline
\fn{2023_04_13_IBERA_S.3.5_S3-B3_9286.d} & samples 2 and 3 of the same batch, same method\\
PXD064980, HLA-II & \fn{S25_Donor5_HLA_rep1_S1-F1_1_13434.d}\newline \fn{S26_Donor5_HLA_rep2_S1-F2_1_13435.d}\newline
\fn{S27_Donor5_HLA_rep3_S1-F3_1_13436.d} & second and third loads of the same donor's peptides\\
PXD062962, mouse liver & \fn{20230510_C31609_048_S502309_wt-86-581_HFD-wt-86_phospho.raw}\newline
\fn{20230510_C31609_049_S502310_wt-86-574_HFD-wt-86_phospho.raw}\newline
\fn{20230510_C31609_040_S502308_wt-86-580_HFD-wt-86_phospho.raw} &
two other mice of the same diet, genotype and treatment, same enrichment\\
PXD062859, T47D & \fn{20211125_Ex_CF_JH_T47D_F7_72h_STY_01_01.raw}\newline
\fn{20211125_Ex_CF_JH_T47D_F7_72h_STY_02_01.raw}\newline \fn{20211125_Ex_CF_JH_T47D_F7_72h_STY_03_01.raw} &
replicates 2 and 3 of the same cell line, condition, time point and enrichment\\
\addlinespace[2pt]
\multicolumn{3}{@{}l}{\emph{Not replicates in the strict sense}}\\
PXD071022, HLA-I & \fn{1319540F_CRC_200323_1.raw}\newline \fn{1319540F_CRC_200323_2.raw}\newline
\fn{1136279F_CRC_200323_1.raw} & second injection of the same tumor, and the tumor of another patient who
shares four of its six HLA-I alleles\\
PXD082505, HLA-I & \fn{1502_ND.raw}\newline \fn{1502_CBZ.raw}\newline \fn{S2_1502_CBZ.raw} &
the same cell line after carbamazepine treatment, and a second injection of that sample\\
\addlinespace[2pt]
\multicolumn{3}{@{}l}{\emph{No companion runs}}\\
PXD073929, frog & \fn{Xt-GludC-HO-toe-no-Claw.raw} & the project's four runs are four different tissues or
fractions\\
PXD074750, \emph{C.\ auris} & \fn{RO_CS_481_241111_CasadevallA_SRDS_Sample_10.raw} & nine runs named by sample
number only, and no replicates stated\\
PXD055085, antibody & eight runs, one per protease & another protease gives other peptides, and the project's other
runs are not described\\
PXD054498, HLA-I & \fn{25-Aug-23_20230825_KP_UWA_PE_P6215_S1-A2_1_2505.d} & four runs, one per patient, with
HLA types that are not given\\
\bottomrule
\end{tabular}
\end{table*}

\noindent\textbf{Retention time.} The 63 sequence features of a peptide are:
\begin{itemize}
\item its 20 residue counts,
\item the SSRCalc retention coefficients~\cite{krokhin2004} of its residues, as their sum, standard deviation, maximum
  over windows of five residues and mean over the last three residues,
\item its net charge at pH 2.5 and 7 and its isoelectric point,
\item $\ln n$ and $\sqrt{n}$ for its length $n$,
\item its N-terminal residue (one-hot, 20),
\item its mean Z1 descriptor~\cite{sandberg1998},
\item its mean Kyte--Doolittle hydropathy~\cite{kyte1982} adjusted to pH 2, 2.5, 3 and 4,
\item the charge asymmetry of its two halves,
\item one count for each modified residue and N-terminal modification of the vocabulary.
\end{itemize}
The three iRT features are $u$, $u^2$ and $u^3$ for the predicted iRT standardized by the mean and standard deviation
of the predicted iRTs of the run's high-confidence peptides. Each distinct peptide gives one row, with the median retention
time of its high-confidence answers. A ridge regression ($\lambda=1$, standardized features, unpenalized intercept) is made
robust by four rounds of Huber reweighting and cross-fitted over five folds split by peptide. The spread $\sigma$ is the
median absolute deviation of the out-of-fold residuals, scaled to a standard deviation, and a model fitted on all rows
predicts every candidate. A user-defined modification adds its own count, so the regression learns its retention
shift on the run, while the iRT predictor reads it by name or as an unknown modification with its mass shift.

\noindent\textbf{Fine-tuning of the iRT predictor.} After the last pass, the predictor's second BiGRU layer and its
MLP are trained on the rows above, with one of the five folds held out. The loss is the absolute difference between
the observed retention time and a cubic of the predicted iRT. The cubic is the run's calibration of the untuned
predictor, fitted beforehand on the training folds and then fixed. Training uses Adam at a learning rate of
$3\times10^{-4}$, batches of 64 and gradient clipping at norm 1. The held-out error is checked every 50 steps,
training stops after 600 steps without improvement or at 4{,}000 steps, and the best weights are kept. The tuned copy
replaces the predictor for this run when at least 100 sequences are held out and the regression fitted on the other
folds predicts them with a spread of at most 0.95 times that of the untuned predictor. The spread $\sigma$ and the
RT term then come from the held-out fold alone.

\noindent\textbf{RT term.} The degrees of freedom $\nu$ and the scale $\omega$ of Eq.~\eqref{eq:rt} are chosen by
maximum likelihood on the held-out residuals, $\nu$ among ten values between 0.75 and 50. The term is added to the third
stage reranker's score of every candidate, including the candidates that the sibling fusion adds.

\section{Companion runs for cross-run spectra enhancement}\label{app:companions}
Table~\ref{tab:companions} lists the companion runs that the rule of \S\ref{sec:protocol} gave for the main test sets,
and why four sets have none. The same rule gave two companions for 15 of the 19 development sets, 11 of them with
replicates in the strict sense (Table~\ref{tab:ablation}).

\section{Precursor input}\label{app:precursor}
All other tools in \S\ref{sec:results} read MGF files written by PrecursorMassFinder (\S\ref{sec:rdr}), and NovoCRF
gets its precursors from the same PrecursorMassFinder inside the engine. We made this choice to keep the comparison fair, and to separate what the
sequencing model gains from what the mass correction gains: a correct precursor mass helps every tool, so every tool
was given it. Table~\ref{tab:precursor} shows what the correction changes on seven of the ten main test sets without
phosphopeptide enrichment.

\begin{table}[t]
\centering\footnotesize
\setlength{\tabcolsep}{5pt}
\caption{Precursor input on seven main test sets (126{,}262 labels with a spectrum in both MGF files). Labels fit:
share of labels (\%) whose mass is within the precursor tolerance of the MGF precursor. The other rows give each
tool's mean peptide recall (\%) over the seven sets when it reads the MGF files written by msconvert and those
written by PrecursorMassFinder.}
\label{tab:precursor}
\begin{tabular}{@{}lrrr@{}}
\toprule
 & msconvert & PrecursorMassFinder & Gain\\
\midrule
Labels fit (\%) & 88.5 & 97.9 & $+$9.4\\
\midrule
InstaNovo & 47.5 & 49.0 & $+$1.5\\
RNovA & 40.5 & 42.2 & $+$1.7\\
$\pi$-PrimeNovo & 42.6 & 44.5 & $+$1.9\\
\Cfive & 38.9 & 39.1 & $+$0.2\\
\bottomrule
\end{tabular}
\end{table}
\begin{table*}[t]
\centering\footnotesize
\setlength{\tabcolsep}{2.7pt}
\renewcommand{\arraystretch}{0.87}
\caption{Peptide recall (\%) of NovoCRF without and with sibling spectra enhancement, on labels whose peptide is among
its training peptides (S, seen) and on its other labels (U, unseen), by the number of missing bonds, for each main
test set, in the layout of Table~\ref{tab:seen}. The first row of each dataset gives the share of seen labels, and
the last row (S\,+\,U) the recall on all its labels. A dash marks fewer than 30 labels. The columns with sibling
spectra repeat those of Tables~\ref{tab:main} and~\ref{tab:seen}.}
\label{tab:nosib}
\ifdefined\numw\else\newlength{\numw}\fi\settowidth{\numw}{100.0}%
\begin{tabular}{@{}cl*{8}{w{r}{\numw}}@{\hspace{12pt}}cl*{8}{w{r}{\numw}}@{}}
\toprule
 & & \multicolumn{4}{c}{Without sibling spectra} & \multicolumn{4}{c}{With sibling spectra} & & & \multicolumn{4}{c}{Without sibling spectra} & \multicolumn{4}{c}{With sibling spectra}\\
\cmidrule(lr){3-6}\cmidrule(lr){7-10}\cmidrule(lr){13-16}\cmidrule(l){17-20}
 & Missing & \multicolumn{2}{c}{one pass} & \multicolumn{2}{c}{two passes} & \multicolumn{2}{c}{one pass} & \multicolumn{2}{c}{two passes} & & Missing & \multicolumn{2}{c}{one pass} & \multicolumn{2}{c}{two passes} & \multicolumn{2}{c}{one pass} & \multicolumn{2}{c}{two passes}\\
\cmidrule(lr){3-4}\cmidrule(lr){5-6}\cmidrule(lr){7-8}\cmidrule(lr){9-10}\cmidrule(lr){13-14}\cmidrule(lr){15-16}\cmidrule(lr){17-18}\cmidrule(l){19-20}
 & bonds & S & U & S & U & S & U & S & U & & bonds & S & U & S & U & S & U & S & U\\
\midrule
\multirow{9}{*}{\rotatebox[origin=c]{90}{\shortstack{PXD076585\\{\scriptsize human, trypsin}}}} & Seen (\%) & \multicolumn{8}{c}{98.6} & \multirow{9}{*}{\rotatebox[origin=c]{90}{\shortstack{PXD071561\\{\scriptsize squid, Lys-C}}}} & Seen (\%) & \multicolumn{8}{c}{8.9}\\
 & 0 & 94.3 & -- & 95.9 & -- & 94.5 & -- & 95.8 & -- &  & 0 & 96.5 & 95.3 & 97.4 & 96.4 & 97.4 & 95.7 & 98.2 & 96.8\\
 & 1 & 81.8 & 57.1 & 85.9 & 60.0 & 85.1 & 54.3 & 88.5 & 62.9 &  & 1 & 86.7 & 88.9 & 88.7 & 90.7 & 93.1 & 92.5 & 96.4 & 93.9\\
 & 2 & 61.7 & 47.1 & 66.3 & 55.9 & 70.0 & 47.1 & 74.4 & 58.8 &  & 2 & 76.1 & 76.7 & 81.4 & 79.7 & 90.0 & 85.4 & 93.8 & 88.0\\
 & 3 & 37.6 & 11.1 & 42.5 & 25.0 & 49.1 & 25.0 & 55.7 & 36.1 &  & 3 & 54.3 & 42.7 & 59.5 & 48.5 & 86.2 & 63.6 & 87.9 & 69.2\\
 & 4--5 & 15.6 & 16.7 & 18.3 & 9.5 & 30.3 & 19.0 & 35.0 & 16.7 &  & 4--5 & 18.9 & 14.8 & 26.3 & 17.5 & 67.4 & 43.3 & 68.4 & 50.4\\
 & $\ge$6 & 1.7 & 0.7 & 1.8 & 0.7 & 10.7 & 4.3 & 11.9 & 4.7 &  & $\ge$6 & 0.5 & 0.6 & 1.6 & 0.8 & 31.7 & 11.4 & 37.2 & 13.1\\
 & All & 40.5 & 16.2 & 43.0 & 17.5 & 48.7 & 19.5 & 51.9 & 22.2 &  & All & 76.8 & 73.2 & 79.3 & 75.1 & 87.7 & 79.8 & 90.1 & 82.0\\
 & S\,+\,U & \multicolumn{2}{c}{40.1} & \multicolumn{2}{c}{42.7} & \multicolumn{2}{c}{48.2} & \multicolumn{2}{c}{51.5} &  & S\,+\,U & \multicolumn{2}{c}{73.5} & \multicolumn{2}{c}{75.5} & \multicolumn{2}{c}{80.5} & \multicolumn{2}{c}{82.7}\\
\addlinespace[1pt]\cmidrule(l){2-10}\cmidrule(l){12-20}\addlinespace[1pt]
\multirow{9}{*}{\rotatebox[origin=c]{90}{\shortstack{PXD073929\\{\scriptsize frog, Glu-C}}}} & Seen (\%) & \multicolumn{8}{c}{9.8} & \multirow{9}{*}{\rotatebox[origin=c]{90}{\shortstack{PXD072115\\{\scriptsize chickpea}}}} & Seen (\%) & \multicolumn{8}{c}{1.1}\\
 & 0 & 97.7 & 97.2 & 97.7 & 98.6 & 95.3 & 97.7 & 100.0 & 98.6 &  & 0 & -- & 70.5 & -- & 78.3 & -- & 84.8 & -- & 89.5\\
 & 1 & 88.7 & 87.3 & 92.5 & 89.0 & 86.8 & 89.3 & 92.5 & 92.1 &  & 1 & 66.7 & 54.3 & 73.3 & 62.3 & 83.3 & 77.0 & 90.0 & 83.5\\
 & 2 & -- & 64.8 & -- & 73.8 & -- & 68.8 & -- & 75.2 &  & 2 & 48.7 & 38.6 & 69.2 & 46.6 & 76.9 & 68.2 & 79.5 & 76.7\\
 & 3 & -- & 39.2 & -- & 51.3 & -- & 46.4 & -- & 59.2 &  & 3 & -- & 25.1 & -- & 29.7 & -- & 56.0 & -- & 64.8\\
 & 4--5 & 18.2 & 14.2 & 24.2 & 17.6 & 27.3 & 22.6 & 39.4 & 27.9 &  & 4--5 & 0.0 & 9.8 & 8.2 & 11.8 & 36.7 & 39.0 & 38.8 & 46.3\\
 & $\ge$6 & 0.0 & 1.5 & 0.0 & 1.2 & 2.2 & 6.0 & 8.9 & 10.3 &  & $\ge$6 & 0.0 & 0.6 & 3.2 & 1.0 & 6.5 & 15.5 & 9.7 & 19.8\\
 & All & 52.7 & 40.1 & 55.9 & 43.9 & 53.6 & 44.8 & 59.0 & 50.1 &  & All & 37.3 & 20.6 & 47.7 & 24.2 & 59.5 & 43.5 & 62.3 & 49.9\\
 & S\,+\,U & \multicolumn{2}{c}{41.3} & \multicolumn{2}{c}{45.1} & \multicolumn{2}{c}{45.7} & \multicolumn{2}{c}{51.0} &  & S\,+\,U & \multicolumn{2}{c}{20.8} & \multicolumn{2}{c}{24.4} & \multicolumn{2}{c}{43.7} & \multicolumn{2}{c}{50.1}\\
\addlinespace[1pt]\cmidrule(l){2-10}\cmidrule(l){12-20}\addlinespace[1pt]
\multirow{9}{*}{\rotatebox[origin=c]{90}{\shortstack{PXD074750\\{\scriptsize \emph{C.\ auris}}}}} & Seen (\%) & \multicolumn{8}{c}{0.6} & \multirow{9}{*}{\rotatebox[origin=c]{90}{\shortstack{PXD055085\\{\scriptsize rabbit antibody}}}} & Seen (\%) & \multicolumn{8}{c}{10.4}\\
 & 0 & 90.3 & 85.3 & 93.5 & 88.7 & 96.8 & 89.9 & 96.8 & 93.3 &  & 0 & 80.0 & 81.2 & 79.8 & 84.2 & 80.0 & 85.2 & 80.7 & 86.7\\
 & 1 & -- & 63.6 & -- & 70.2 & -- & 74.5 & -- & 81.9 &  & 1 & 77.3 & 75.2 & 80.2 & 80.2 & 80.4 & 84.2 & 82.1 & 87.1\\
 & 2 & -- & 40.2 & -- & 48.3 & -- & 57.5 & -- & 68.3 &  & 2 & 66.7 & 56.9 & 70.7 & 63.7 & 73.3 & 72.7 & 77.3 & 78.6\\
 & 3 & -- & 17.0 & -- & 23.6 & -- & 36.4 & -- & 51.1 &  & 3 & 48.8 & 39.4 & 49.6 & 43.9 & 62.0 & 64.5 & 62.0 & 72.0\\
 & 4--5 & -- & 5.1 & -- & 7.6 & -- & 24.6 & -- & 35.8 &  & 4--5 & 17.2 & 15.8 & 21.6 & 20.7 & 43.1 & 48.4 & 45.7 & 57.6\\
 & $\ge$6 & -- & 0.4 & -- & 0.4 & -- & 9.3 & -- & 14.3 &  & $\ge$6 & 2.8 & 2.4 & 2.1 & 2.9 & 11.3 & 19.6 & 17.0 & 24.2\\
 & All & 47.2 & 31.0 & 52.0 & 35.2 & 64.0 & 44.2 & 71.2 & 52.6 &  & All & 68.3 & 55.1 & 69.9 & 59.3 & 72.5 & 68.9 & 74.3 & 73.3\\
 & S\,+\,U & \multicolumn{2}{c}{31.1} & \multicolumn{2}{c}{35.3} & \multicolumn{2}{c}{44.3} & \multicolumn{2}{c}{52.7} &  & S\,+\,U & \multicolumn{2}{c}{56.5} & \multicolumn{2}{c}{60.4} & \multicolumn{2}{c}{69.3} & \multicolumn{2}{c}{73.4}\\
\addlinespace[1pt]\cmidrule(l){2-10}\cmidrule(l){12-20}\addlinespace[1pt]
\multirow{9}{*}{\rotatebox[origin=c]{90}{\shortstack{PXD071022\\{\scriptsize HLA-I, colorectal}}}} & Seen (\%) & \multicolumn{8}{c}{82.3} & \multirow{9}{*}{\rotatebox[origin=c]{90}{\shortstack{PXD082505\\{\scriptsize HLA-I, B*15:02}}}} & Seen (\%) & \multicolumn{8}{c}{66.6}\\
 & 0 & 97.0 & 93.0 & 97.7 & 94.1 & 97.6 & 93.5 & 98.0 & 94.8 &  & 0 & 100.0 & 100.0 & 100.0 & 98.7 & 99.7 & 100.0 & 100.0 & 100.0\\
 & 1 & 89.2 & 83.5 & 90.9 & 85.4 & 93.4 & 84.9 & 94.3 & 87.5 &  & 1 & 95.8 & 95.0 & 97.8 & 96.1 & 97.5 & 96.8 & 97.8 & 97.1\\
 & 2 & 73.1 & 64.6 & 75.8 & 69.3 & 86.4 & 72.6 & 88.4 & 75.6 &  & 2 & 85.5 & 81.9 & 87.9 & 87.4 & 95.2 & 90.5 & 95.5 & 90.8\\
 & 3 & 47.4 & 40.0 & 52.1 & 43.8 & 74.0 & 57.0 & 76.9 & 61.7 &  & 3 & 62.7 & 59.1 & 67.6 & 67.8 & 85.5 & 80.3 & 88.0 & 85.6\\
 & 4--5 & 18.7 & 19.1 & 21.1 & 21.6 & 57.3 & 42.9 & 59.9 & 47.8 &  & 4--5 & 28.4 & 25.6 & 33.5 & 31.8 & 78.1 & 64.5 & 80.8 & 69.3\\
 & $\ge$6 & 6.4 & 3.4 & 6.5 & 3.4 & 41.7 & 21.5 & 44.5 & 24.9 &  & $\ge$6 & 3.8 & 2.0 & 4.2 & 2.4 & 66.9 & 31.3 & 69.6 & 34.5\\
 & All & 70.2 & 60.2 & 72.3 & 62.9 & 83.9 & 69.6 & 85.5 & 72.8 &  & All & 66.2 & 55.5 & 69.1 & 59.8 & 88.5 & 74.9 & 89.9 & 77.5\\
 & S\,+\,U & \multicolumn{2}{c}{68.5} & \multicolumn{2}{c}{70.6} & \multicolumn{2}{c}{81.4} & \multicolumn{2}{c}{83.2} &  & S\,+\,U & \multicolumn{2}{c}{62.6} & \multicolumn{2}{c}{66.0} & \multicolumn{2}{c}{84.0} & \multicolumn{2}{c}{85.7}\\
\addlinespace[1pt]\cmidrule(l){2-10}\cmidrule(l){12-20}\addlinespace[1pt]
\multirow{9}{*}{\rotatebox[origin=c]{90}{\shortstack{PXD054498\\{\scriptsize HLA-I, effusion}}}} & Seen (\%) & \multicolumn{8}{c}{63.6} & \multirow{9}{*}{\rotatebox[origin=c]{90}{\shortstack{PXD064980\\{\scriptsize HLA-II, macrophage}}}} & Seen (\%) & \multicolumn{8}{c}{33.9}\\
 & 0 & 79.1 & 75.1 & 90.2 & 84.3 & 87.3 & 79.2 & 93.7 & 86.0 &  & 0 & 70.2 & 54.3 & 81.3 & 67.1 & 84.0 & 65.8 & 90.5 & 76.5\\
 & 1 & 58.3 & 58.5 & 74.2 & 68.5 & 77.7 & 69.0 & 87.1 & 76.3 &  & 1 & 47.0 & 35.2 & 58.9 & 47.4 & 76.6 & 60.5 & 84.8 & 72.4\\
 & 2 & 33.3 & 30.6 & 47.5 & 40.1 & 65.0 & 50.6 & 76.1 & 57.3 &  & 2 & 33.2 & 24.1 & 44.7 & 32.5 & 67.6 & 47.1 & 78.4 & 63.2\\
 & 3 & 15.8 & 13.4 & 25.7 & 21.1 & 60.1 & 39.5 & 70.1 & 46.5 &  & 3 & 19.9 & 14.6 & 25.3 & 19.4 & 56.7 & 39.9 & 68.6 & 55.7\\
 & 4--5 & 7.3 & 4.3 & 10.5 & 7.1 & 50.1 & 34.2 & 59.4 & 40.9 &  & 4--5 & 8.8 & 4.5 & 11.3 & 6.2 & 47.4 & 26.5 & 60.5 & 36.9\\
 & $\ge$6 & 0.8 & 1.3 & 0.8 & 0.0 & 26.4 & 12.7 & 34.9 & 19.0 &  & $\ge$6 & 0.9 & 0.2 & 1.3 & 0.5 & 28.6 & 9.6 & 38.1 & 15.0\\
 & All & 57.6 & 50.5 & 70.4 & 59.5 & 76.9 & 63.4 & 85.5 & 70.4 &  & All & 20.4 & 8.5 & 25.9 & 11.3 & 52.4 & 24.6 & 62.9 & 33.8\\
 & S\,+\,U & \multicolumn{2}{c}{55.0} & \multicolumn{2}{c}{66.4} & \multicolumn{2}{c}{72.0} & \multicolumn{2}{c}{80.0} &  & S\,+\,U & \multicolumn{2}{c}{12.5} & \multicolumn{2}{c}{16.2} & \multicolumn{2}{c}{34.0} & \multicolumn{2}{c}{43.7}\\
\addlinespace[1pt]\cmidrule(l){2-10}\cmidrule(l){12-20}\addlinespace[1pt]
\multirow{9}{*}{\rotatebox[origin=c]{90}{\shortstack{PXD062962\\{\scriptsize mouse, phospho}}}} & Seen (\%) & \multicolumn{8}{c}{65.7} & \multirow{9}{*}{\rotatebox[origin=c]{90}{\shortstack{PXD062859\\{\scriptsize human, phospho}}}} & Seen (\%) & \multicolumn{8}{c}{91.3}\\
 & 0 & 93.7 & 87.6 & 95.7 & 90.1 & 96.1 & 91.6 & 97.0 & 94.5 &  & 0 & 91.8 & 75.6 & 94.1 & 85.5 & 95.3 & 90.1 & 95.9 & 94.2\\
 & 1 & 85.7 & 74.9 & 88.0 & 78.5 & 89.8 & 81.8 & 91.4 & 84.5 &  & 1 & 84.1 & 55.1 & 86.8 & 61.2 & 91.3 & 76.2 & 91.6 & 78.0\\
 & 2 & 64.9 & 56.6 & 68.5 & 56.3 & 78.9 & 70.1 & 80.4 & 71.6 &  & 2 & 59.4 & 33.6 & 62.3 & 41.4 & 76.1 & 55.3 & 75.9 & 62.5\\
 & 3 & 42.0 & 39.7 & 45.6 & 42.7 & 63.3 & 55.6 & 66.4 & 59.9 &  & 3 & 39.5 & 29.2 & 43.1 & 31.9 & 62.0 & 42.5 & 63.0 & 42.5\\
 & 4--5 & 23.8 & 17.5 & 25.2 & 20.1 & 54.2 & 42.3 & 55.9 & 45.7 &  & 4--5 & 16.4 & 11.9 & 16.6 & 13.6 & 42.3 & 31.2 & 46.1 & 32.4\\
 & $\ge$6 & 1.0 & 1.3 & 2.5 & 1.5 & 25.5 & 15.2 & 28.9 & 18.6 &  & $\ge$6 & 2.8 & 1.0 & 2.2 & 1.4 & 20.4 & 10.1 & 22.0 & 8.7\\
 & All & 78.0 & 60.8 & 80.3 & 63.1 & 85.2 & 70.8 & 86.6 & 73.7 &  & All & 76.3 & 34.3 & 78.6 & 39.0 & 85.1 & 50.8 & 85.7 & 52.9\\
 & S\,+\,U & \multicolumn{2}{c}{72.1} & \multicolumn{2}{c}{74.4} & \multicolumn{2}{c}{80.2} & \multicolumn{2}{c}{82.2} &  & S\,+\,U & \multicolumn{2}{c}{72.6} & \multicolumn{2}{c}{75.2} & \multicolumn{2}{c}{82.2} & \multicolumn{2}{c}{82.9}\\
\bottomrule
\end{tabular}
\end{table*}

With MGF files written by msconvert (vendor peak picking, precursor refinement), 11.5\% of the labels had a mass outside
the precursor tolerance of the MGF precursor: the instrument had reported the $^{13}$C peak (3.9\%) or a heavier one
(0.5\%), a peak 1--7 isotope spacings lighter (1.8\%), or a mass off by a non-isotopic amount (5.4\%). The labels are
fine, since PEAKS corrects the precursor before its search. PrecursorMassFinder's precursors fit 97.9\% of the labels,
and every tool gains from them, by 0.2 to 1.9 points of mean recall.
A model that reads the precursor mass as one input among others can partly work around a wrong one: with the
msconvert files, \Cfive\ still finds up to 38\% of a dataset's labels whose precursor does not fit, InstaNovo up to
30\%, $\pi$-PrimeNovo up to 17\% and RNovA up to 7\%.
NovoCRF searches only peptides that fit the mass it is given, so it depends on this correction more.

\section{NovoCRF without sibling spectra enhancement}\label{app:nosib}
To separate what NovoCRF reads from one spectrum from what the sibling spectra add, we ran it on the twelve main test
sets again with sibling spectra enhancement (\S\ref{sec:sibling}) switched off: the same program, raw files and
options as for Table~\ref{tab:main}, without the links between spectra, the fusion after either pass, the fine-tuning
examples that a sibling confirms and the siblings' part in the reported confidence. Table~\ref{tab:nosib} gives the
recall on seen and unseen labels in the layout of Table~\ref{tab:seen}, beside that of the default.

Without the enhancement, the recall of the two passes on all labels is 5.9 to 27.5 points lower, and it is the
highest of the five tools on eight of the twelve sets (Table~\ref{tab:main}). On the spectra with peptides that no tool has seen
(Table~\ref{tab:novel}), it is the highest on eleven of the twelve sets, and on all twelve among those with at
most one missing bond.

The enhancement raises the recall on seen and on unseen labels. Over all labels of a dataset (rows All), the recall
of the two passes rises on both sides on each of the twelve datasets, by 3.1 to 37.0 points on seen labels and by 4.7
to 25.7 points on unseen labels. Among the 62 cells of a dataset and a bin of missing bonds with at least 30 labels
on each side, it rises in 59 on the seen side and in 61 on the unseen side, and it falls in none.

\end{document}